\documentclass[10pt,letterpaper,compsoc,conference]{iiswc22}

\usepackage{cite}
\usepackage{amsmath,amssymb,amsfonts}
\usepackage{algorithmic}
\usepackage{graphicx}
\usepackage[dvipsnames]{xcolor}
\usepackage[final]{microtype}
\usepackage[italic]{mathastext}
\usepackage{libertine}
\usepackage[T1]{fontenc}
\usepackage{textcomp}
\usepackage[varqu,varl]{zi4}
\usepackage[all]{nowidow}
\usepackage[auth-lg,affil-it]{authblk}
\usepackage[keeplastbox]{flushend}
\usepackage{url}
\usepackage{adjustbox}
\usepackage{multirow}
\usepackage{enumitem}
\usepackage[algo2e, linesnumbered]{algorithm2e}
\usepackage{algorithm}
\usepackage{bm}
\usepackage{balance}

\newlength\mylen

\usepackage[caption=false]{subfig}

\begin{document}

\title{Characterizing the Efficiency of Graph Neural Network Frameworks with a Magnifying Glass}

\renewcommand\Authsep{\qquad}
\renewcommand\Authand{\qquad}
\renewcommand\Authands{\qquad}

\author[1]{Xin Huang}
\author[2]{Jongryool Kim}
\author[3]{Bradley Rees}
\author[1]{Chul-Ho Lee}
\affil[1]{Texas State University}
\affil[2]{SK hynix America}
\affil[3]{NVIDIA}

\maketitle

\begin{abstract}
Graph neural networks (GNNs) have received great attention due to their success in various graph-related learning tasks. Several GNN frameworks have then been developed for fast and easy implementation of GNN models. Despite their popularity, they are not well documented, and their implementations and system performance have not been well understood. In particular, unlike the traditional GNNs that are trained based on the entire graph in a \emph{full-batch} manner, recent GNNs have been developed with different graph sampling techniques for \emph{mini-batch} training of GNNs on large graphs. While they improve the scalability, their training times still depend on the implementations in the frameworks as sampling and its associated operations can introduce non-negligible overhead and computational cost. In addition, it is unknown how much the frameworks are `eco-friendly' from a green computing perspective. In this paper, we provide an in-depth study of two mainstream GNN frameworks along with three state-of-the-art GNNs to analyze their performance in terms of runtime and power/energy consumption. We conduct extensive benchmark experiments at several different levels and present detailed analysis results and observations, which could be helpful for further improvement and optimization.
\end{abstract}

\section{Introduction}

Graphs are everywhere, from social networks to transportation networks to biological networks. It is of vital importance to mining graph-structured data~\cite{li2021efficient,li2021estimating} and learning on graphs~\cite{hamilton2020graph,ma2021deep} since they contain rich underlying information and can be used for a wide range of applications. In particular, graph neural networks (GNNs) have attracted a lot of attention in recent years. Unlike the conventional machine learning (ML) algorithms, which assume data samples are independent and identically distributed, GNNs take graph-structured data as input for downstream tasks and capture the correlation between data samples (nodes in the graph) according to their connections (edges in the graph). GNNs have been shown to be effective for many tasks, such as representation learning, node classification, and link prediction.

Early GNN studies mainly reply on general deep learning (DL) frameworks, such as TensorFlow~\cite{tensorflow2015-whitepaper} and PyTorch~\cite{NEURIPS2019_9015}. It is, however, non-trivial to implement a GNN model using the DL frameworks. While they are designed and optimized for regular yet often dense data, real-world graphs often exhibit irregularity and sparsity, thereby making them inefficient for GNNs. Thus motivated, several GNN frameworks have been developed to speed up the computation and to simplify GNN implementation. The examples include Graph Nets~\cite{battaglia2018relational}, Deep Graph Library (DGL)~\cite{wang2019deep}, PyTorch Geometric (PyG)~\cite{FeyLenssen2019}, StellarGraph~\cite{StellarGraph}, Spektral~\cite{grattarola2021graph}, TF-Geometric~\cite{hu2021efficient}, and CogDL~\cite{cen2021cogdl}.

DGL and PyG are two most popular ones among them, thanks to their user-friendly designs, rich functionalities, and easy-to-follow tutorials. Inspired by NetworkX~\cite{hagberg2008exploring}, DGL uses a graph-centric programming abstraction, making it easy for NetworkX users to use. It defines a `DGLGraph' object as its key data structure for computations with graph-structured data and GNN operations. DGL also realizes the message passing operations of GNNs with generalized sparse-dense matrix multiplication (g-SpMM) and generalized sampled dense-dense matrix multiplication (g-SDDMM). Furthermore, it develops highly tuned CPU and GPU kernels for GNN operations and supports a wide range of applications for general-purpose graph learning. In addition, PyG is an extension library of PyTorch for deep learning on graph-structured data. It provides a simple `MessagePassing' interface for the message passing operations based on a gather-and-scatter paradigm, which is built on top of its own PyTorch Scatter\footnote{\url{https://github.com/rusty1s/pytorch_scatter}.} and PyTorch Sparse\footnote{\url{https://github.com/rusty1s/pytorch_sparse}.} that provide dedicated kernels for relevant computations. It also provides a large number of off-the-shelf examples along with a lot of commonly used benchmark datasets for users to easily use and test. Both DGL and PyG have been updated and optimized significantly compared with their initial versions. However, their current implementations and system performance are not well understood.

On the other hand, `\emph{sustainability}' becomes an important factor in both industry and academia due to climate change. Energy and power consumption ought to be critical metrics in ML/DL since training advanced models are often energy and resource hungry. Early studies in ML/DL have, however, mainly focused on improving their model accuracy to achieve state-of-the-art performance. Schwartz \textit{et al.}~\cite{schwartz2020green} recently urge researchers to provide not only the accuracy, but also the efficiency in terms of carbon emission, energy consumption, runtime, to name a few. Strubell \textit{et al.}~\cite{strubell2019energy} bring power and energy concerns in ML/DL research by estimating the financial and environmental costs of building well-trained state-of-the-art natural language processing models. There is then a movement, albeit slowly, in recent studies that take power and energy consumption into consideration~\cite{schwartz2020green}. Nonetheless, there is no prior work to quantify the power and energy consumption of GNN models and frameworks.

In this paper, we study the two mainstream GNN frameworks -- DGL and PyG, by evaluating their efficiency in terms of runtime (not only at the level of each key function but also at the level of the entire model), and power and energy consumption.\footnote{Here we do not report the accuracy results of GNN models as they mainly depend on their underlying GNN methods, not the software frameworks. It has also been shown that there is no clear difference between two frameworks when it comes to the accuracy of each GNN model~\cite{wang2019deep, NEURIPS2019_9015, wu2021performance}.} We benchmark their performance via functional testing on each main component of GNNs and three state-of-the-art GNNs, namely GraphSAGE, ClusterGCN, and GraphSAINT, which adopt graph sampling for mini-batch training. We further provide case studies on different implementation strategies, GPU-based sampling, and full-batch training. We provide detailed and comprehensive analysis to fully understand their performance and find opportunities for further improvement and optimization. We summarize our contributions as follows:
\vspace{0mm}
\begin{itemize}[itemsep=2pt,leftmargin=1.1em]
\item First, we present the results of functional testing on each key component of building a GNN model in DGL and PyG, including data loader, sampler, and graph convolutional layers. We find that DGL is more efficient for sampling and GNN operations, especially when it comes to large graphs.

\item Second, we evaluate the efficiency of sampling and GNN operations on different hardware devices (CPU vs. GPU).

\item Third, we provide runtime breakdown of three state-of-the-art GNNs in both frameworks. Our results indicate that there is still a room for further improvement, especially for sampling and data movement.

\item Finally, we quantify the power and energy consumption of GNN models and frameworks. To the best of our knowledge, we are the first to analyze the GNN performance from such a green computing perspective.
\end{itemize}

\section{Background and Related Work}

\subsection{Graph Neural Networks}

GNNs have emerged as an effective means for learning on graph-structured data. They commonly rely on a `feature aggregation' mechanism, which can be written as
\begin{equation*}
\setlength{\abovedisplayskip}{5pt}
\setlength{\belowdisplayskip}{5pt}
\mathbf{H}^{(l+1)} = f(\mathbf{G}\mathbf{H}^{(l)}\mathbf{W}^{(l)}),
\end{equation*}
where $f(\cdot)$ is a non-linear activation function that is applied element-wise, $\mathbf{G}$ is a graph matrix representing the graph structure, e.g., the adjacency or (normalized) Laplacian marix of the input graph, and $\mathbf{H}^{(l)}$ and $\mathbf{W}^{(l)}$ are the node feature/embedding matrix and the weight matrix of neural networks at $l$-th layer, respectively. In other words, it is the neighborhood aggregation or message passing as each node in the graph updates its current feature vector by aggregating the feature vectors (messages) from its neighbors.

\begin{figure}[t!]
	\captionsetup[subfloat]{captionskip=1pt}
	\centering
	\subfloat[Full-batch training]{%
		\includegraphics[width=0.85\linewidth, trim=0cm 0cm 0cm 0cm, clip]{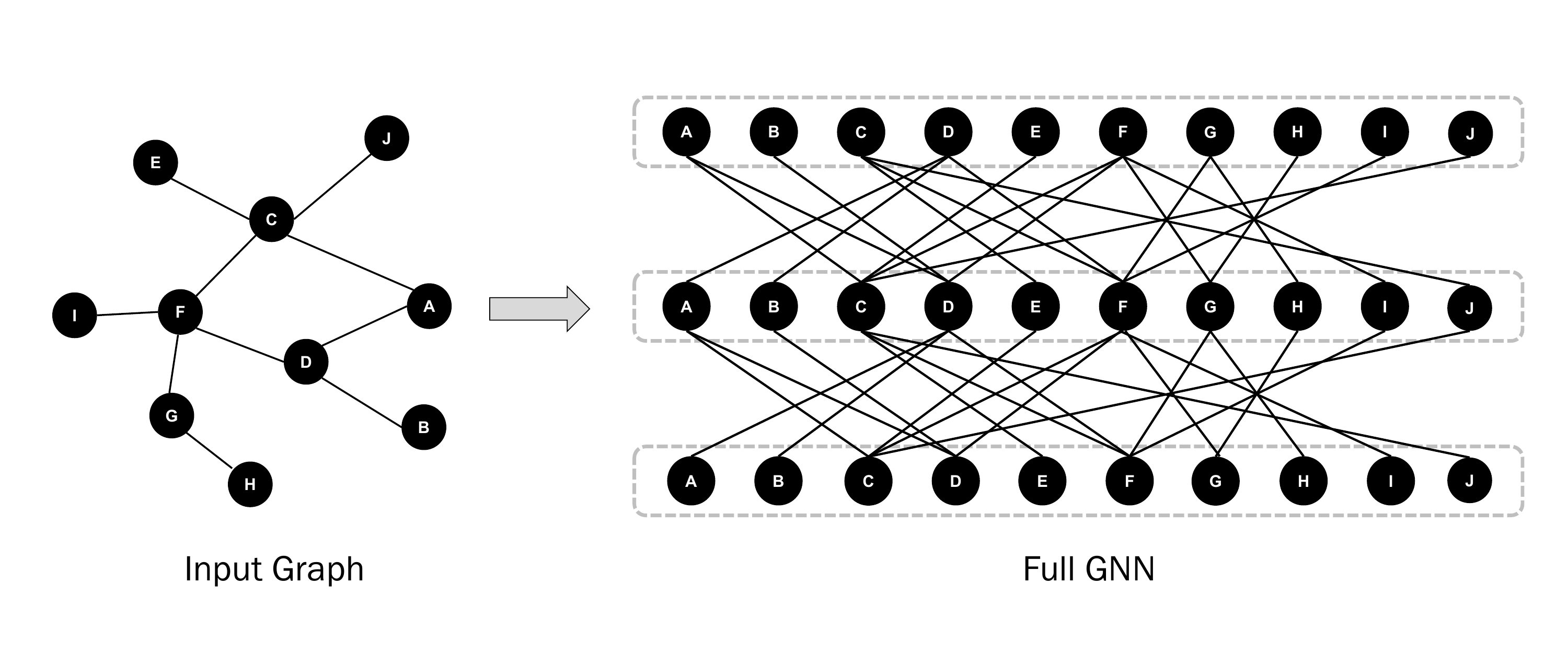}
	}
	\vspace{-3mm}
	\subfloat[Mini-batch training]{%
		\includegraphics[width=0.85\linewidth, trim=0cm 0cm 0cm 0cm, clip]{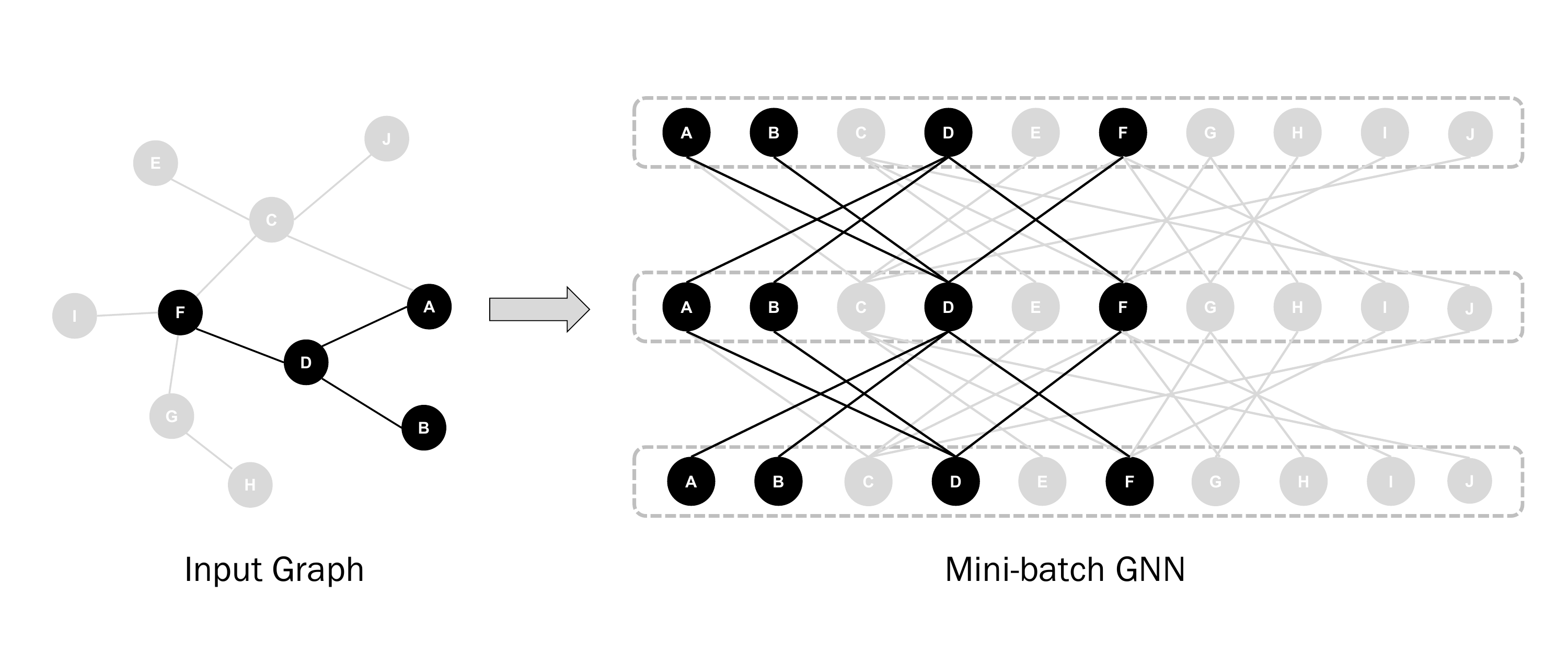}
	}
	\hspace{0mm}
	\caption{Two training methods for GNNs.}
	\label{fig:gnn}
	\vspace{-3mm}
\end{figure}

Due to the feature aggregation mechanism or the interdependence of the nodes (samples), traditional GNNs such as GCN~\cite{kipf2016semi} and GAT~\cite{velivckovic2017graph} were trained using the \emph{full-batch} gradient descent, as shown in Figure~\ref{fig:gnn}(a). In other words, they require the entire graph and node features to be maintained in memory, leading to a \emph{scalability} issue with large graphs. To cope with the scalability issue, recent GNNs have then adopted `sampling' techniques to construct \emph{mini-batches} based on the graph structure to train GNNs on large graphs, as mini-batch gradient descent is used for deep neural networks. See Figure~\ref{fig:gnn}(b) for illustration.

Hamilton \textit{et al.}~\cite{hamilton2017inductive} proposed GraphSAGE, which is the first work that introduces the use of sampling in training GNNs to improve the scalability. It combines neighborhood sampling, which samples $k$-hop neighbors with a fixed sampling size for feature aggregation, with mini-batch training. However, its resulting computation graph can be still explosive and thus cause an out-of-memory issue for large graphs. To alleviate this issue, Chen \textit{et al.}~\cite{chen2018fastgcn} developed FastGCN, which samples a fixed number of nodes in each GNN layer independently based on a pre-computed probability distribution. Nonetheless, it can generate isolated nodes, thereby leading to an accuracy drop. Zou \textit{et al.}~\cite{zou2019layer} proposed a layer-dependent importance sampling algorithm called LADIES to resolve the sparsity issue in FastGCN, while it introduces additional computational cost and non-negligible overhead in the sampling process.

\begin{table*}[t]
	\renewcommand{\arraystretch}{1.2}
	\caption{Dataset statistics}
	\label{table:dataset}
	\centering
	\scriptsize
	\begin{adjustbox}{width=1.8\columnwidth,center}
		\begin{tabular}{|c|c|c|c|c|c|c|}
			\hline
			Dataset & Description & \# Nodes & \# Edges & \# Features & \# Classes & Train / Val / Test \\
			\hline
			\hline
			PPI & Protein-Protein Interactions & 14,755 & 225,270 & 50 & 121 & 0.66 / 0.12 / 0.22 \\
			\hline
			Flickr & Images Sharing Common Properties & 89,250 & 899,756 & 500 & 7 & 0.50 / 0.25 / 0.25 \\
			\hline
			ogbn-Arxiv & Citation Network of arXiv CS papers & 169,343 & 1,166,243 & 128 & 40 & 0.54 / 0.29 / 0.17 \\
			\hline
			Reddit & Online Communities & 232,965 & 114,615,892 & 602 & 41 & 0.66 / 0.10 / 0.24\\	
			\hline
			Yelp & Businesses and Reviews & 716,847 & 13,954,819 & 300 & 100 & 0.75 / 0.10 / 0.15 \\
			\hline
			ogbn-Products & Amazon Product Co-purchasing Network & 2,449,029 & 61,859,140 & 100 & 47 & 0.08 / 0.02 / 0.90 \\
			\hline
		\end{tabular}
	\end{adjustbox}
\end{table*}

In addition, Chiang \textit{et al.}~\cite{chiang2019cluster} proposed ClusterGCN, which partitions the input graph into many small clusters, some of which are then randomly selected to form a subgraph -- or, more precisely, mini-batch, during training. It highly improves the scalability of GNNs, although it can lead to data imbalance and information loss issues. Zeng \textit{et al.}~\cite{zeng2019graphsaint} proposed GraphSAINT, which constructs training batches by sampling subgraphs of the input graph. They leveraged graph sampling techniques, such as node sampling, edge sampling, and random walk-based sampling, to obtain subgraphs.

\subsection{Related Work}

There are a few GNN benchmark studies in the literature. Dwivedi \textit{et al.}~\cite{dwivedi2020benchmarking} introduced a benchmark framework along with a set of medium-scale graph datasets for a large collection of GNN models. They developed the benchmark framework on top of PyG and DGL and presented the results in accuracy and training time, \emph{yet} without any detailed component analysis. Duan \textit{et al.}~\cite{duan2018benchmarking} used a greedy \emph{hyperparameter} search method to tune up the performance of several GNN models and reported the resulting accuracy of each model and its corresponding time and space complexity. Zhang \textit{et al.}~\cite{zhang2020architectural} provided a detailed workload analysis on the \emph{inference} of GNNs. Lin \textit{et al.}~\cite{lin2022characterizing} focused on the \emph{distributed training} benchmark of three GNN models implemented in PyG. The studies in \cite{baruah2021gnnmark,mernyei2020wiki,hu2020open} presented new datasets for GNN benchmarking.

The work by Wu \textit{et al.}~\cite{wu2021performance} is most relevant to our work as it is also concerned about the performance analysis of DGL and PyG. It was, however, based only on five datasets of \emph{small-size} graphs with six GNN models, which are mostly traditional ones. Three datasets are for `graph' classification as a downstream task. The smallest one has 600 graphs, each with about 30 nodes and 60 edges on average, while the largest one has 80K graphs, each with about 70 nodes and 500 edges on average. The other datasets are two small graphs (the larger one has about 20K nodes and 40K edges) for `node' classification. For this downstream task, they focused on the \emph{full-batch} training, not to mention lack of any detailed component analysis.

We can summarize the \emph{differences} between our work and the GNN benchmark literature as follows. First, we provide a detailed and comprehensive analysis of DGL and PyG not only at the level of the efficiency (total training time) but also at the level of the runtime of each key component of GNN models. Second, our benchmark of the frameworks is done based on a wide range of graphs, having the largest one with about 2.4M nodes and 61M edges, and three representative GNNs that support mini-batch training for scalability. Finally, we present the energy and power efficiency of GNN models and frameworks.

\section{Methodology}

\subsection{GNN Models}

To evaluate the performance of two popular GNN frameworks -- DGL~\cite{wang2019deep} and PyG~\cite{FeyLenssen2019}, we first consider several convolutional layers, which are key components of GNNs. We then consider three representative sampling-based GNNs, namely GraphSAGE~\cite{hamilton2017inductive}, ClusterGCN~\cite{chiang2019cluster}, and GraphSAINT~\cite{zeng2019graphsaint}, implemented in DGL and PyG.

\subsection{Datasets}
We focus on supervised node classification tasks in this work. To this end, we consider six popular real-world graph datsets, each of whose description and statistics are provided in Table~\ref{table:dataset}. See ~\cite{zeng2019graphsaint} and~\cite{hu2020open} for more details on the datasets. As for how to split each dataset for training, validation, and testing, we follow the common way in the GNN benchmark literature, which is to use `ﬁxed partitions' given by the original authors. The details are reported in the `Train/Val/Test' column of Table~\ref{table:dataset}.

\subsection{Hardware and Software Configuration}

For hardware, all experiments are conducted on a Linux server equipped with Dual Intel Xeon Silver 4114 CPUs @ 2.2GHz with 64GB RAM, and an NVIDIA Quadro RTX 8000 GPU with 48GB memory.

For software, we use Python 3.8, PyTorch v1.11.0, DGL v0.8.2, and PyG v2.0.4. All GNN models were implemented based on the official examples provided by DGL with PyTorch backend and PyG. To match the implementations in both frameworks for a fair comparison, we set the same values of the hyperparameters of samplers, convolutional layers, and other components of GNN models as long as both frameworks provide the same functional APIs. We use the default settings as provided by the frameworks otherwise. Our code is available on GitHub.\footnote{\url{https://github.com/xhuang2016/GNN-Benchmark}.}

Our main focus in this work is to evaluate the efficiency of the GNN frameworks in runtime and energy/power consumption. Note that we here do not consider the accuracy of each GNN model as there is no clear difference between the frameworks~\cite{wang2019deep, FeyLenssen2019, wu2021performance}. We use \texttt{pyinstrument}\footnote{\url{https://github.com/joerick/pyinstrument}.} to measure the runtime of each key function of GNNs and that of each GNN model along with its breakdown results. In addition, we use \texttt{CodeCarbon}\footnote{\url{https://github.com/mlco2/codecarbon}.} to measure power and energy consumption, which is a Python package for tracking carbon emissions produced by algorithms and programs.\footnote{This profiling tool is a Python wrapper of Intel running average power limit (RAPL) interface and NVIDIA ‘pynvml’ library. For CPUs, it measures energy consumption by reading the Intel RAPL files and computes the power by dividing energy by time duration. For GPUs, it reads the instant power recorded by pynvml and computes the energy by multiplying power by time duration between two consecutive measurements.} We use the sampling interval of 0.1 seconds in this tool instead of its default setting, which is 15 seconds. Considering the fact that it is a software tool, we admit possible discrepancies between measured values and actual ones. Nonetheless, we emphasize that \emph{relative} comparisons remain meaningful and informative regardless, not to mention that we are the first work to evaluate the power and energy consumption of GNNs and GNN frameworks.

\section{Results and Discussion}

In this section, we provide and discuss the detailed benchmark results on the efficiency of DGL and PyG.

\begin{figure}[t]
	\vspace{0mm}
	\centering
	\includegraphics[width=0.95\linewidth, trim=2mm 1mm 2mm 1mm, clip]{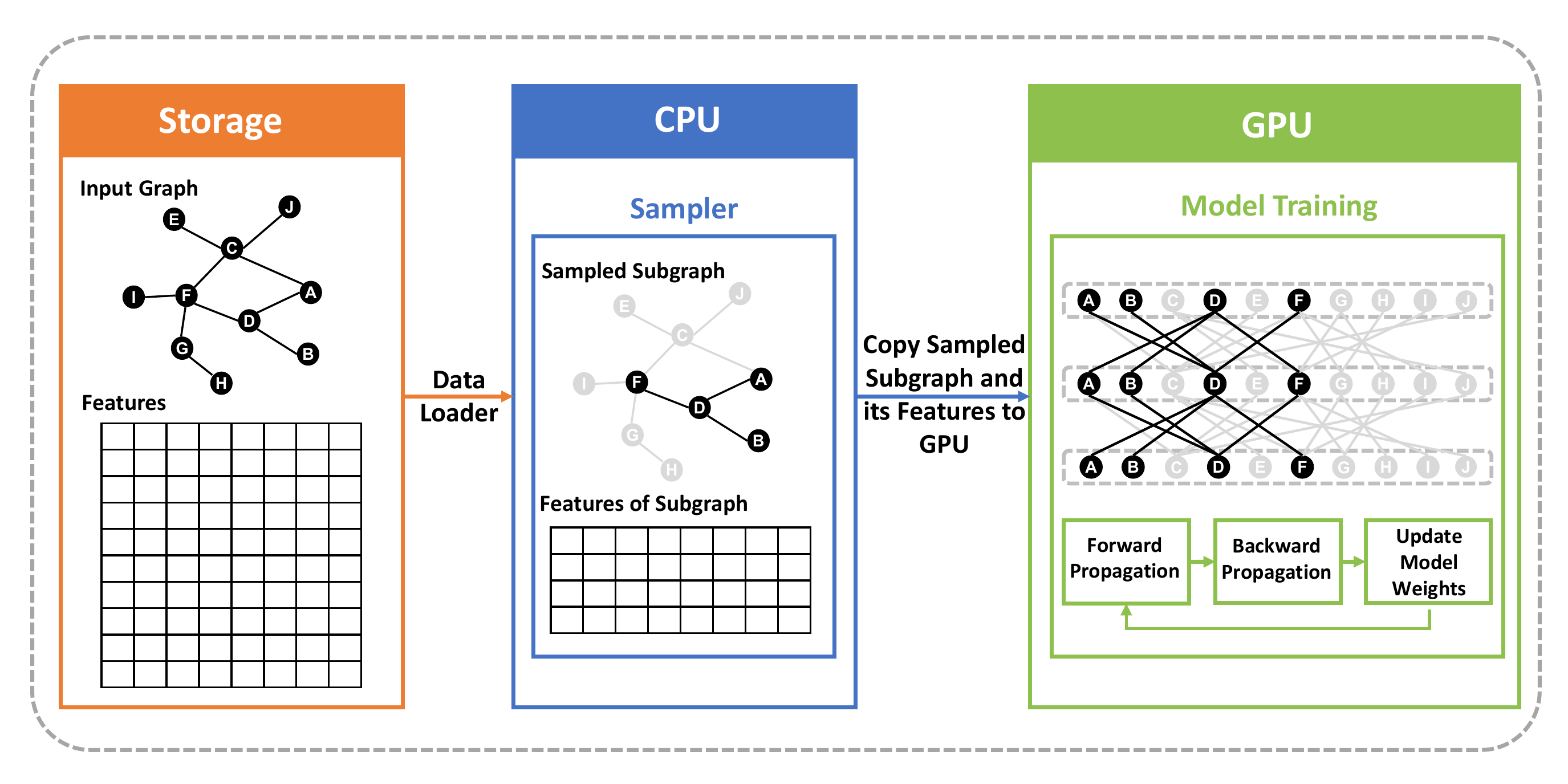}
	\vspace{0mm}
	\caption{Workflow of sampling-based GNN training.}
	\vspace{0mm}
	\label{fig:workflow}
\end{figure}

\subsection{Functional Testing}

Figure~\ref{fig:workflow} illustrates the end-to-end workflow of training a sampling-based GNN with mini-batch training. It can be divided into the following three main processes: data loading, graph sampling, and model training. We thus conduct `functional testing' on each main process to evaluate the performance of DGL and PyG. Note that for the entire training process, data loading is a one-time operation while the other two process, i.e., graph sampling and model training, are performed repeatedly and periodically for each training batch. Note also that we do not consider the inference of each model in this paper. We repeat the experiments for each functional test for ten times and report the average values. In addition, for the functional tests, we do not include the power/energy consumption results since the runtime of some functions are too small, e.g., a few milliseconds, which can lead to incorrect power/energy measurement.

\begin{figure}[t]
	\vspace{0mm}
	\centering
	\includegraphics[width=0.7\linewidth, trim=2mm 1mm 2mm 1mm, clip]{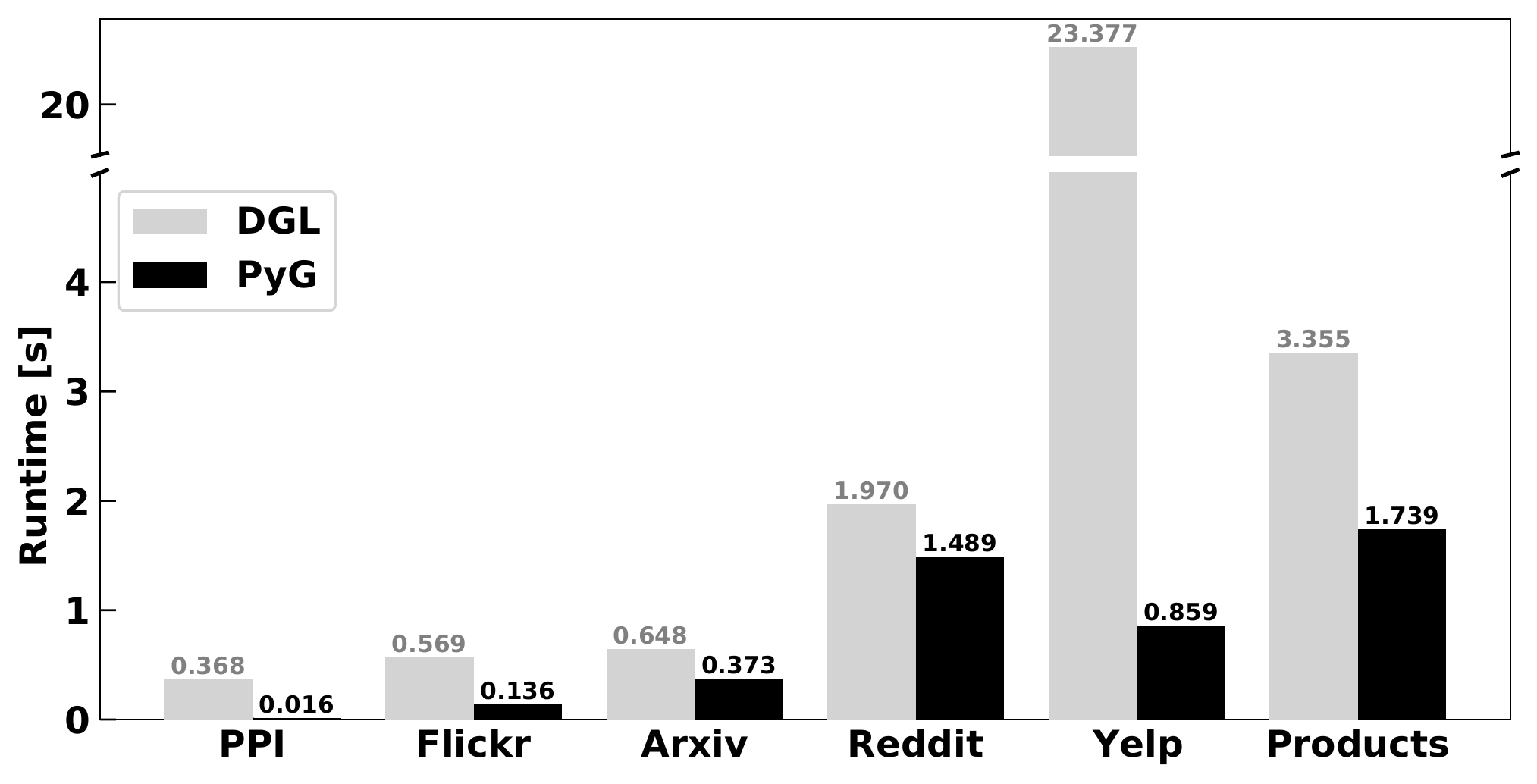}
	\vspace{0mm}
	\caption{Runtime of data loader.}
	\vspace{0mm}
	\label{fig:loader}
	\vspace{-3mm}
\end{figure}

\vspace{1mm}
\noindent \textbf{Data loader.} We first compare the data loader of DGL and PyG, which is used to load the input graph and its associated node features from storage and to create a library-specific graph object for the next process of graph sampling and model training. We present the runtime results in Figure~\ref{fig:loader}.

\vspace{1mm}

\noindent \textbf{Observation 1:} \textit{PyG's data loader is more efficient and user-friendly than DGL’s data loader.}

\vspace{1mm}

There are two main reasons. First, while both frameworks provide an easy-to-use interface to create and process the datasets, PyG integrates more datasets (around 80) into its library as compared with DGL (around 40). Specifically, five out of six datasets used in this work can be directly accessed from PyG's `dataset' module while three datasets are already included in DGL. Note that, for the datasets that are not included in the libraries, we follow the official instructions to process the raw datasets and to create their corresponding graph objects. Second, DGL uses a graph-centric programming abstraction, which makes rich information of the input graph accessible and enables full control of manipulating the input graph. As a consequence, the workload of creating a `DGLGraph' object is relatively higher than its counterpart in PyG.

\begin{figure}[t]
	\captionsetup[subfloat]{captionskip=1pt}
	\centering
	\subfloat[Neighborhood sampler]{%
		\includegraphics[width=0.47\linewidth, trim=0cm 0cm 0cm 0cm, clip]{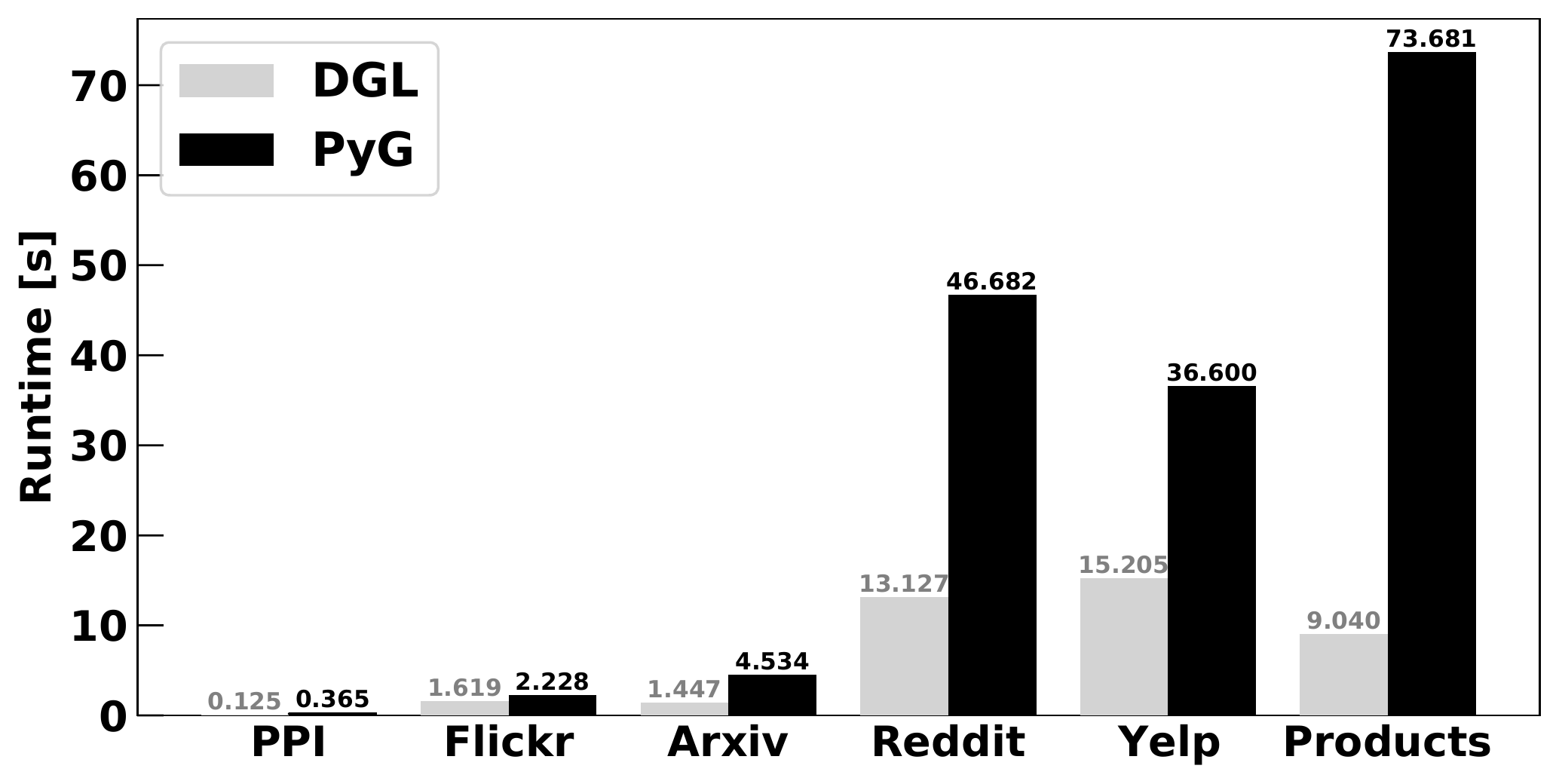}
	}
	\hspace{0mm}
	\subfloat[GraphSAINT sampler]{%
		\includegraphics[width=0.47\linewidth, trim=0cm 0cm 0cm 0cm, clip]{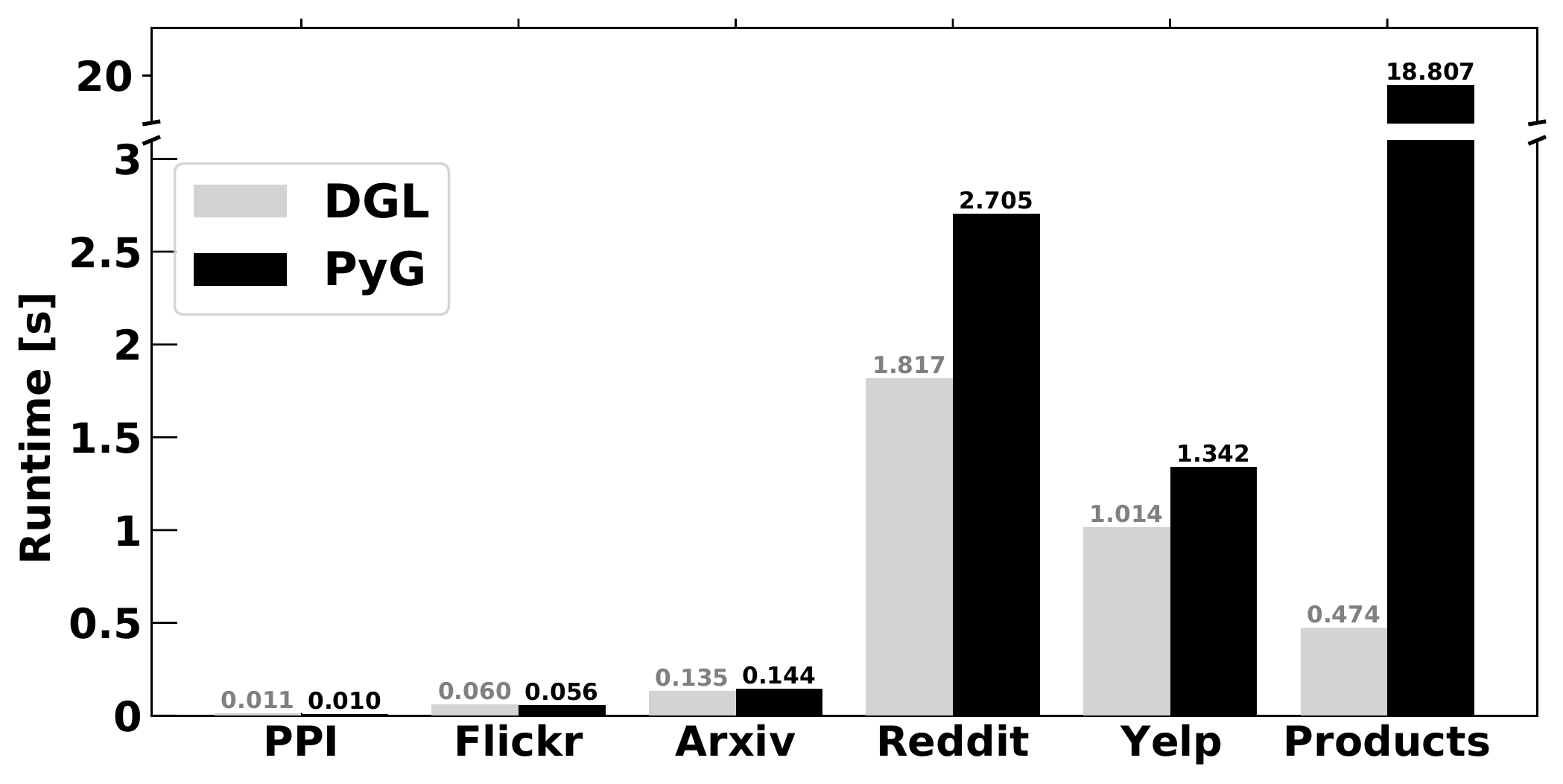}
	}
	\vspace{0mm}
		\subfloat[ClusterGCN sampler: METIS]{%
		\includegraphics[width=0.47\linewidth, trim=0cm 0cm 0cm 0cm, clip]{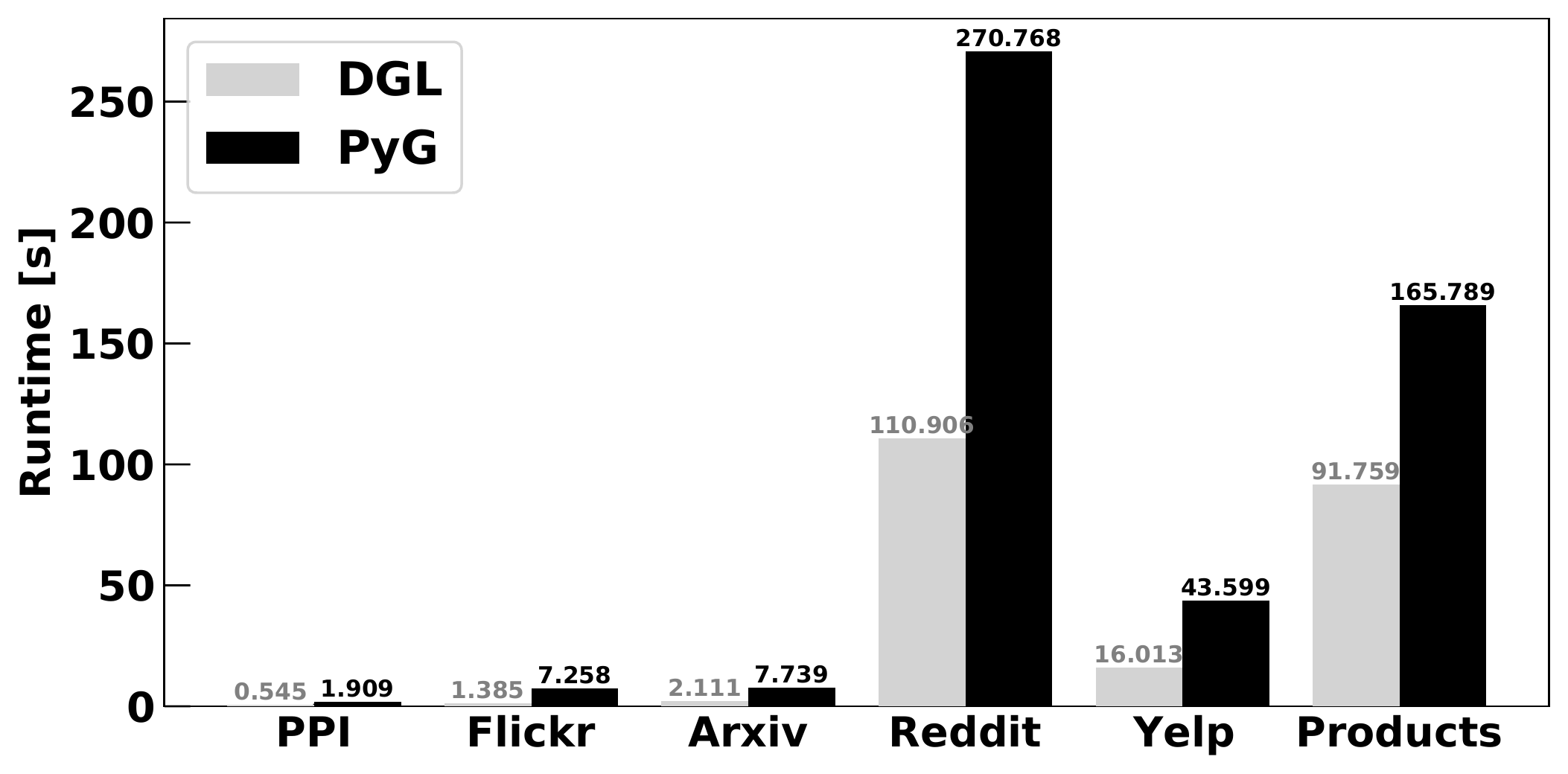}
	}
	\hspace{0mm}
	\subfloat[ClusterGCN sampler: Combining]{%
		\includegraphics[width=0.47\linewidth, trim=0cm 0cm 0cm 0cm, clip]{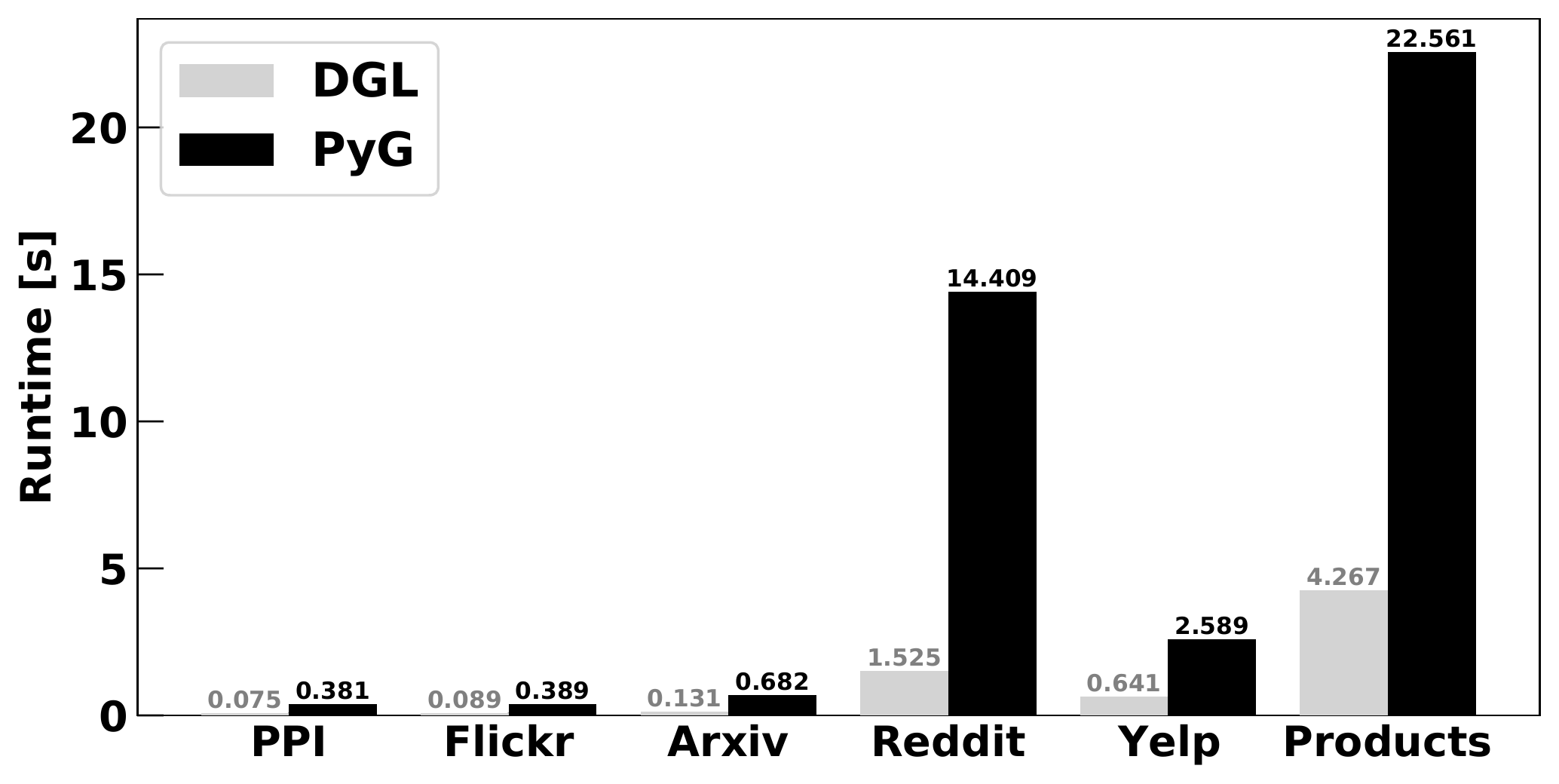}
	}
	\vspace{0mm}
	\caption{Runtime comparison of graph samplers. Note that the range of y-axis is different across different figures.}
	\label{fig:sampler}
	\vspace{-3mm}
\end{figure}

\begin{figure*}[t]
	\vspace{0mm}
	\captionsetup[subfloat]{captionskip=1pt}
	\centering
	\subfloat[GCNConv-CPU]{%
		\includegraphics[width=0.23\linewidth, trim=0cm 0cm 0cm 0cm, clip]{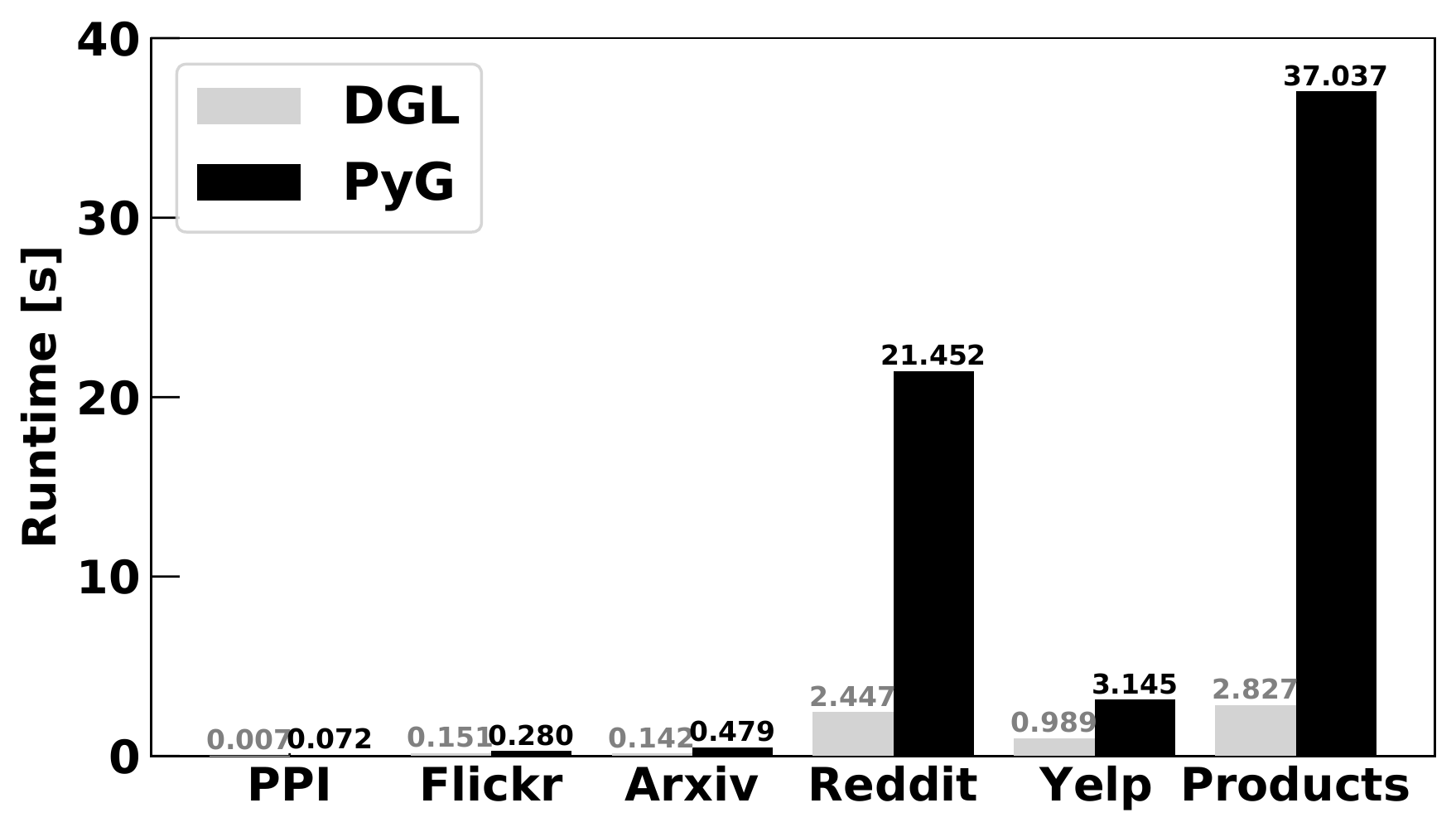}
	}
	\hspace{0mm}
	\subfloat[GCNConv-GPU]{%
		\includegraphics[width=0.23\linewidth, trim=0cm 0cm 0cm 0cm, clip]{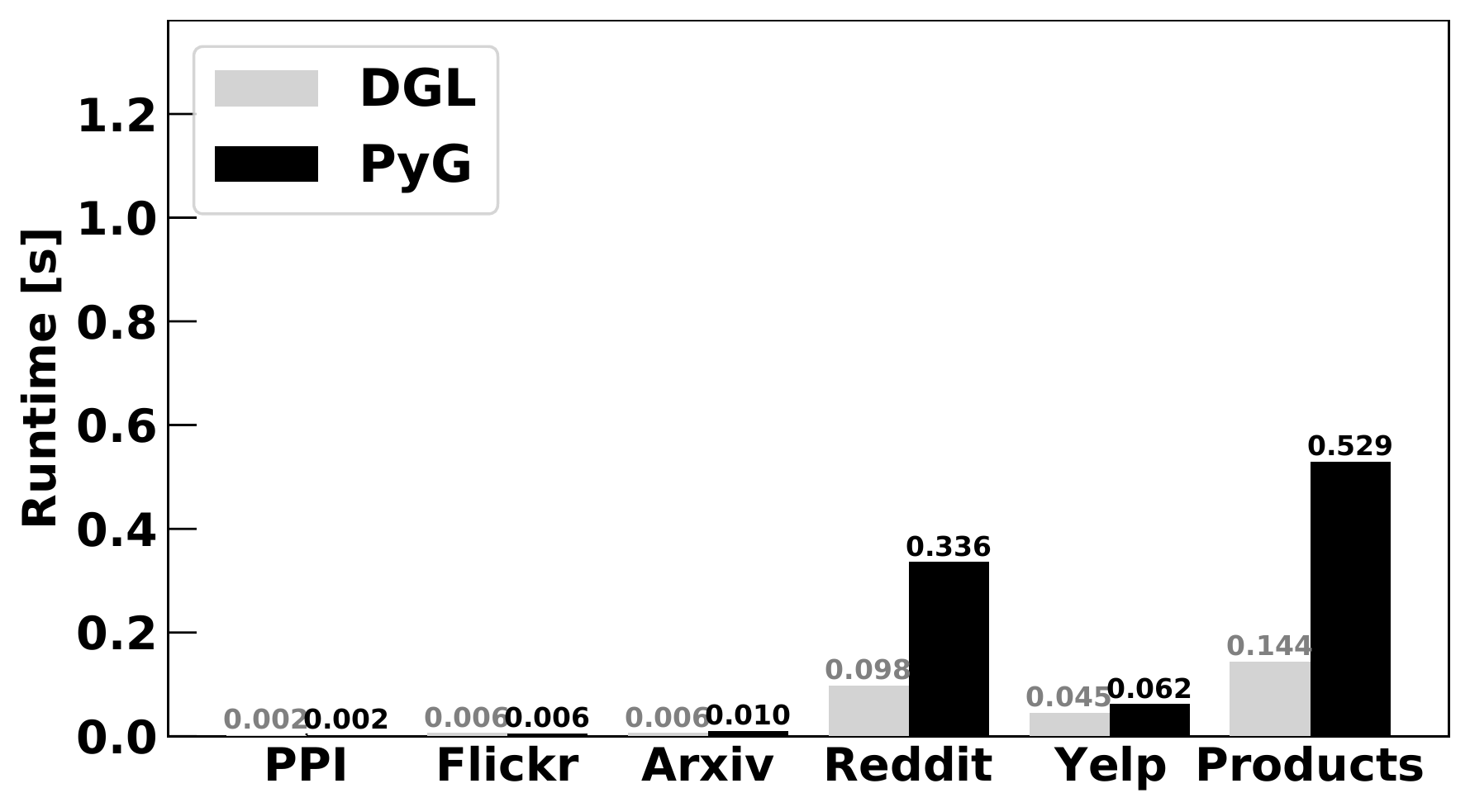}
	}
	\hspace{0mm}
	\subfloat[GCN2Conv-CPU]{%
		\includegraphics[width=0.23\linewidth, trim=0cm 0cm 0cm 0cm, clip]{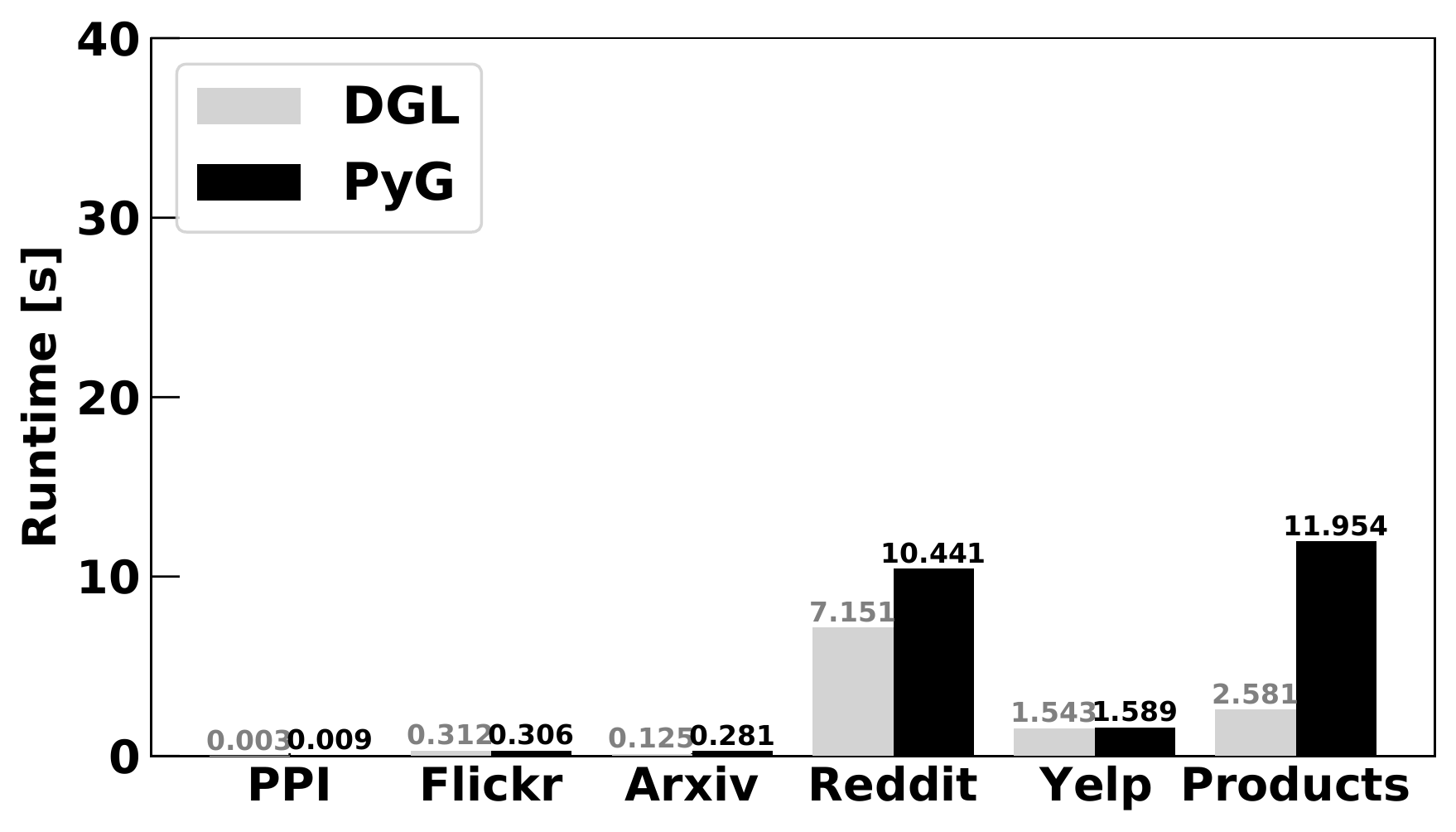}
	}
	\hspace{0mm}
	\subfloat[GCN2Conv-GPU]{%
		\includegraphics[width=0.23\linewidth, trim=0cm 0cm 0cm 0cm, clip]{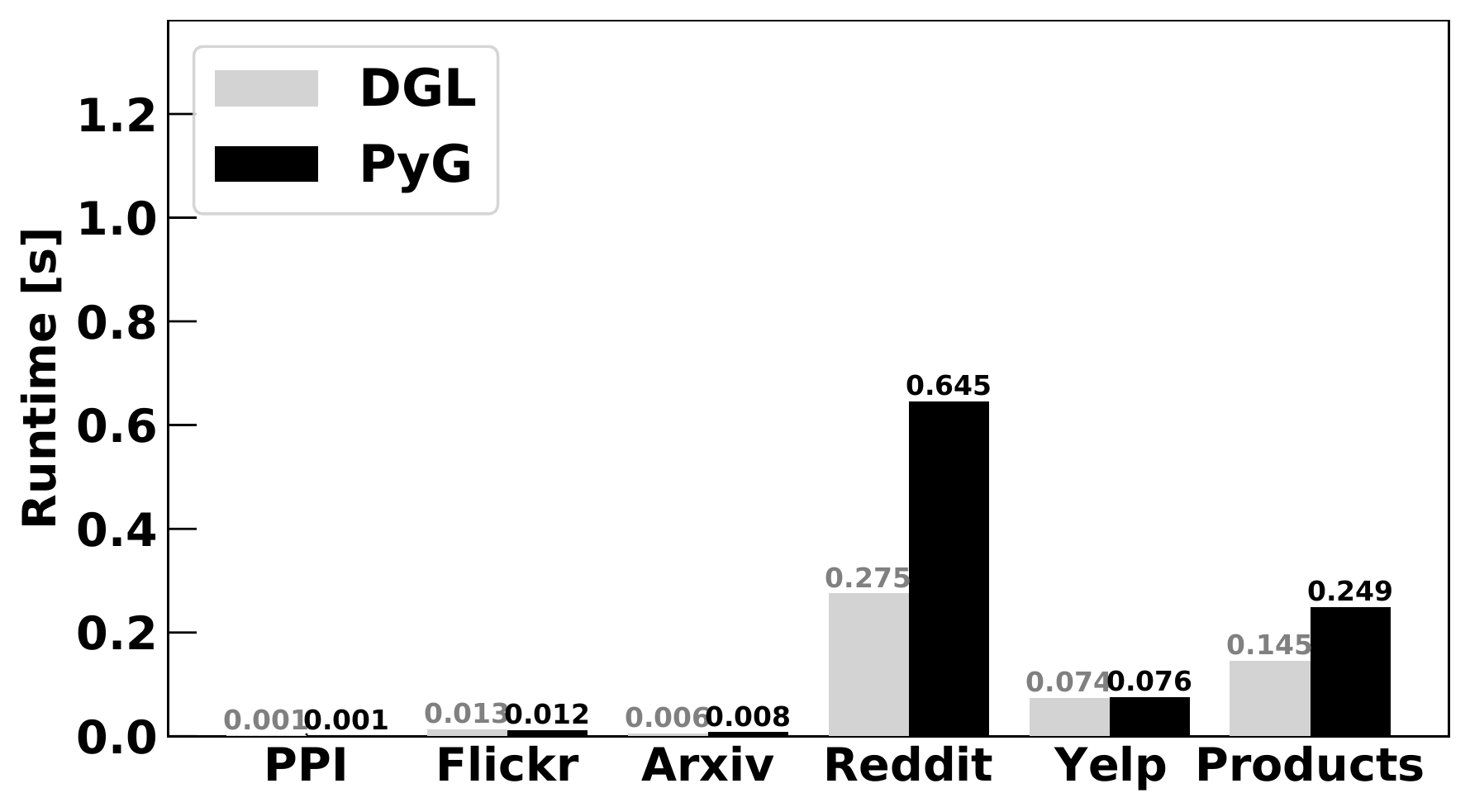}
	}
	\vspace{-3mm}
	\subfloat[GATConv-CPU]{%
		\includegraphics[width=0.23\linewidth, trim=0cm 0cm 0cm 0cm, clip]{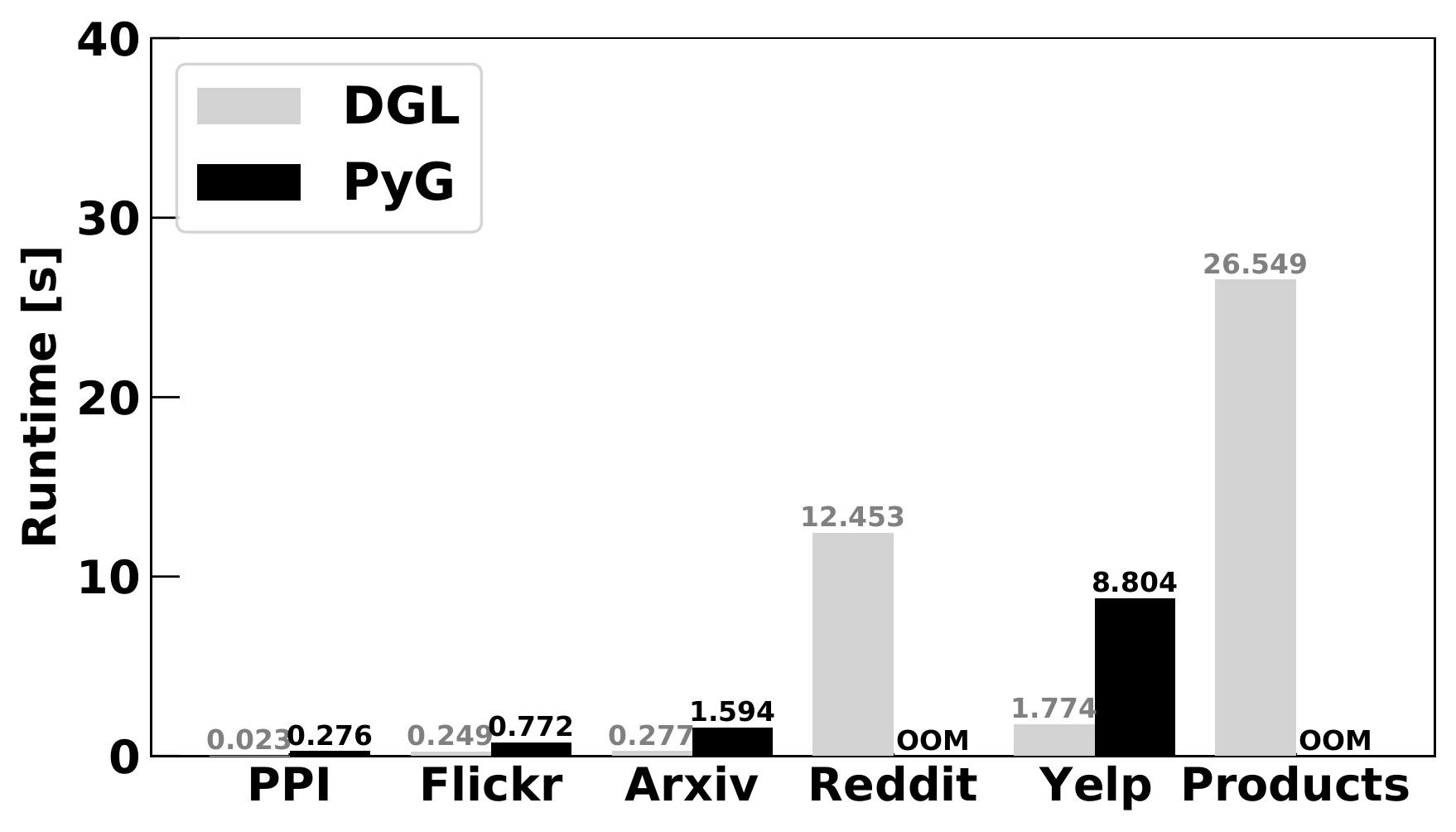}
	}
	\hspace{0mm}
	\subfloat[GATConv-GPU]{%
		\includegraphics[width=0.23\linewidth, trim=0cm 0cm 0cm 0cm, clip]{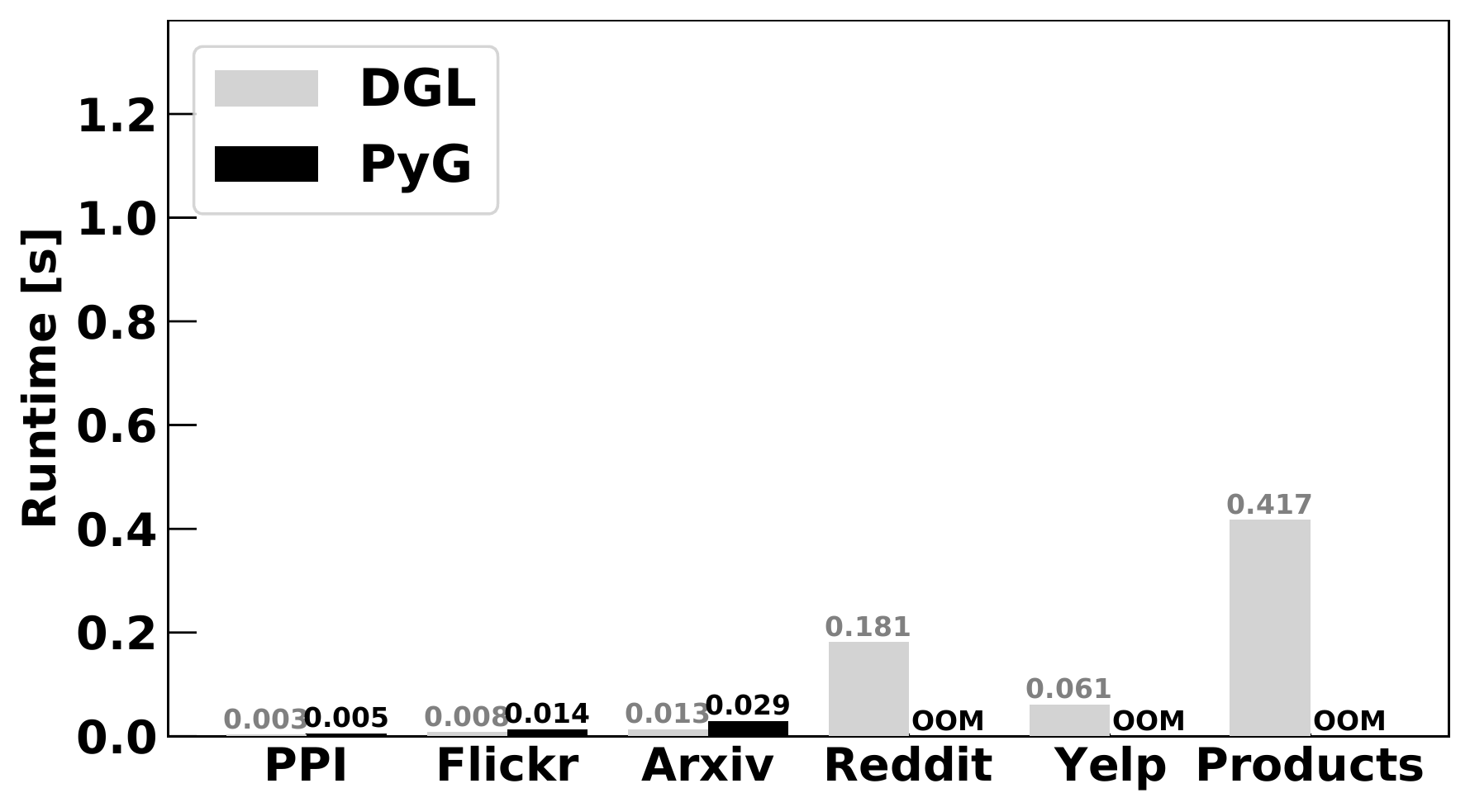}
	}
	\hspace{0mm}
	\subfloat[GATv2Conv-CPU]{%
		\includegraphics[width=0.23\linewidth, trim=0cm 0cm 0cm 0cm, clip]{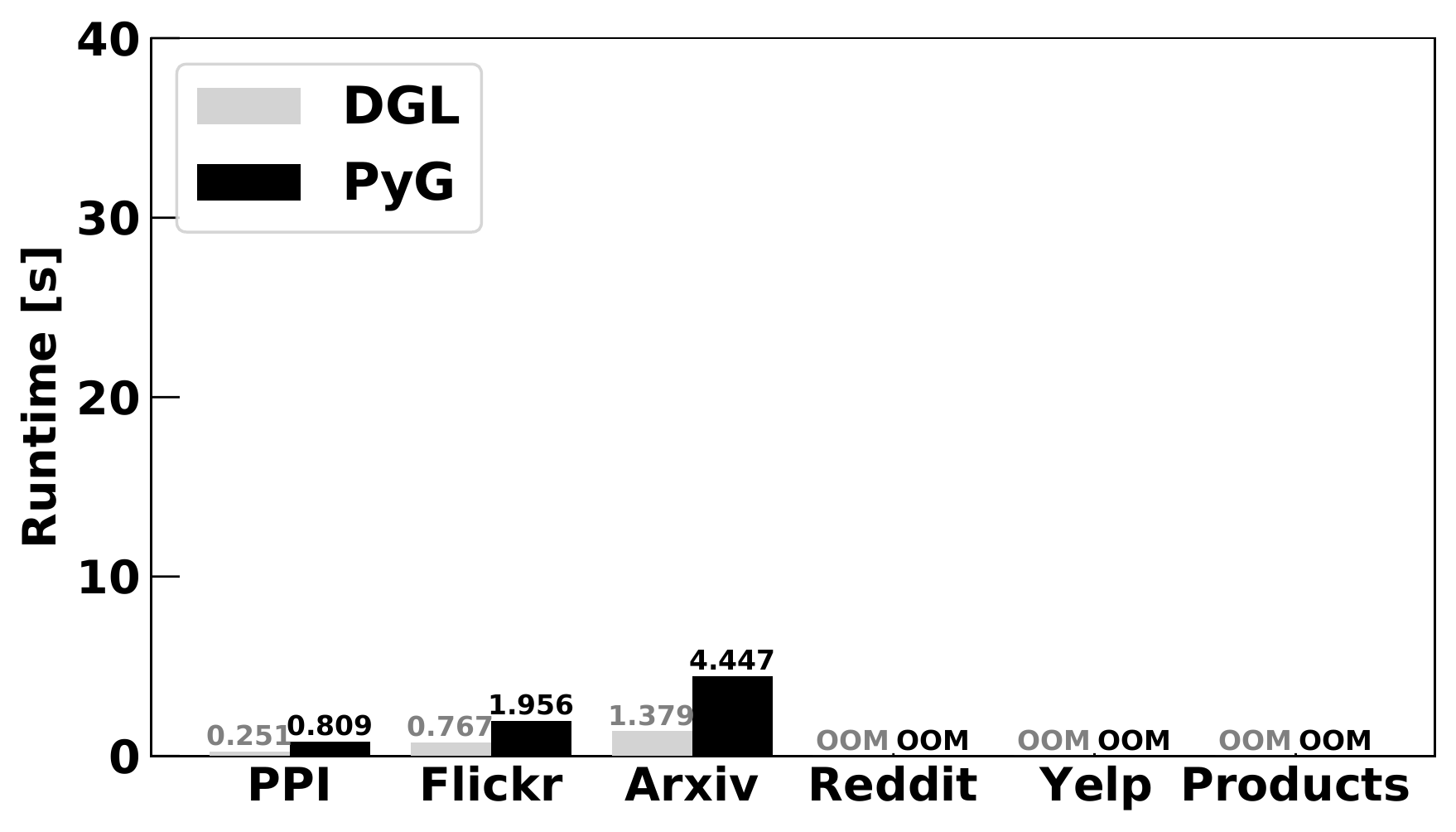}
	}
	\hspace{0mm}
	\subfloat[GATv2Conv-GPU]{%
		\includegraphics[width=0.23\linewidth, trim=0cm 0cm 0cm 0cm, clip]{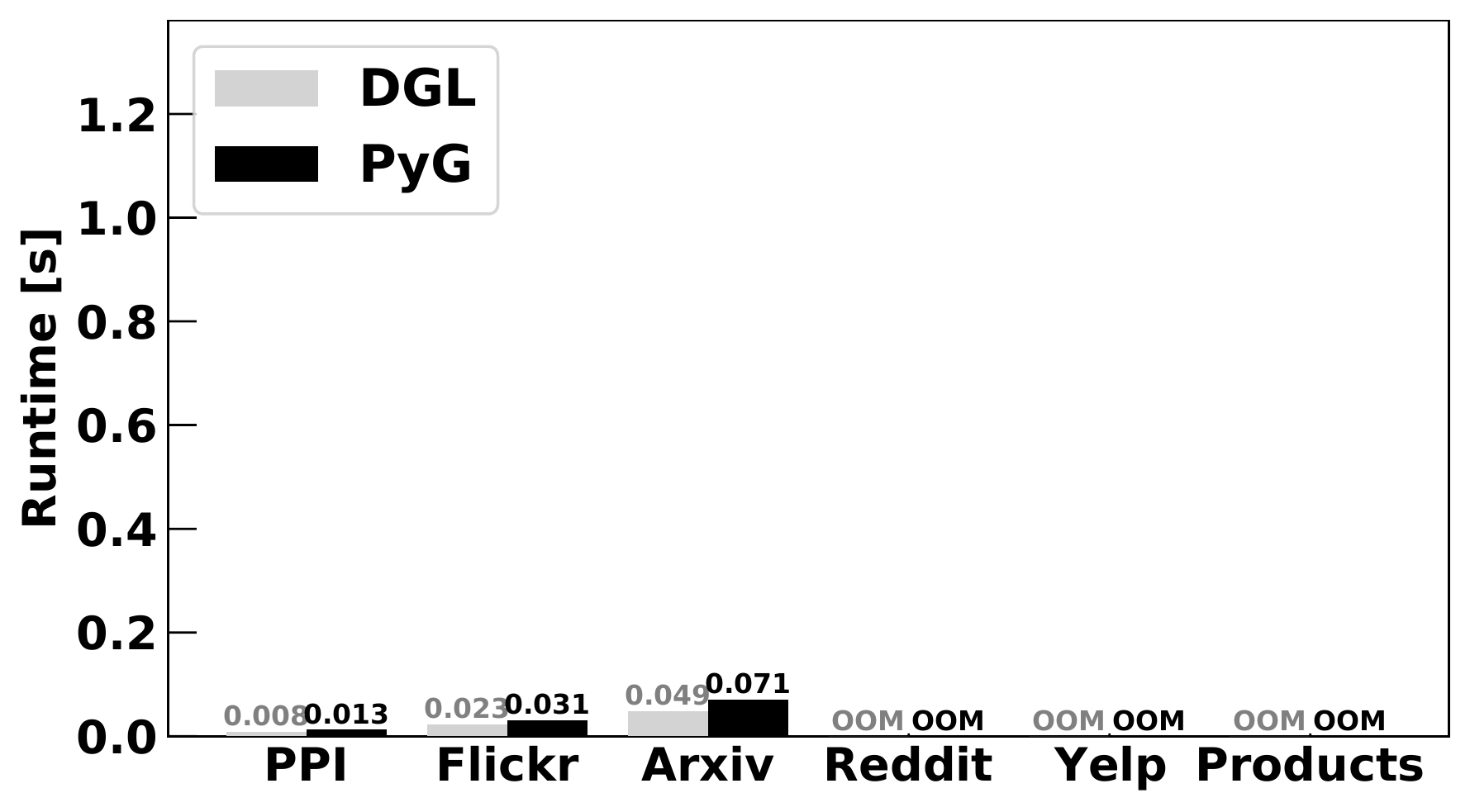}
	}
	\vspace{-3mm}
	\subfloat[SAGEConv-CPU]{%
		\includegraphics[width=0.23\linewidth, trim=0cm 0cm 0cm 0cm, clip]{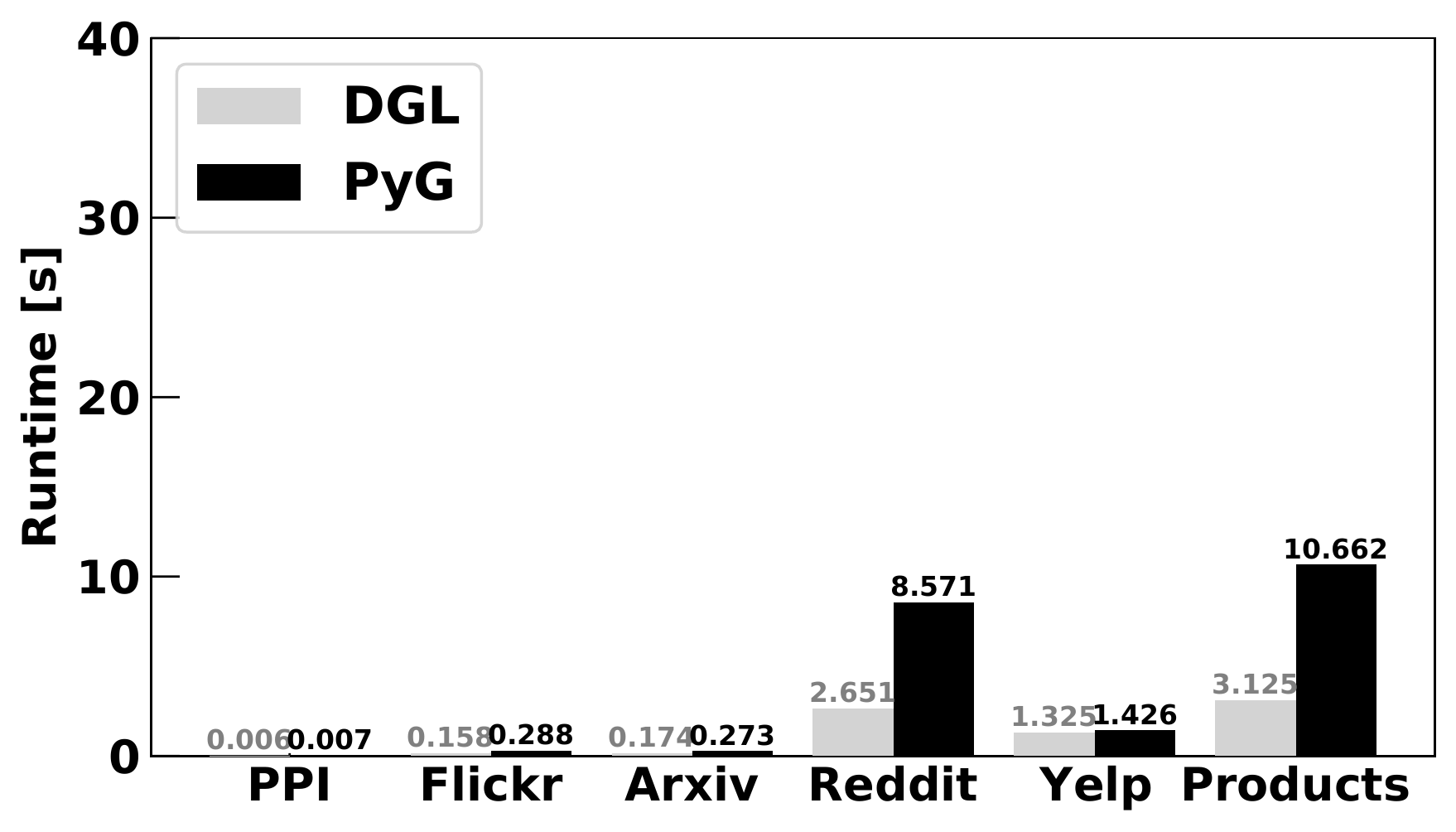}
	}
	\hspace{0mm}
	\subfloat[SAGEConv-GPU]{%
		\includegraphics[width=0.23\linewidth, trim=0cm 0cm 0cm 0cm, clip]{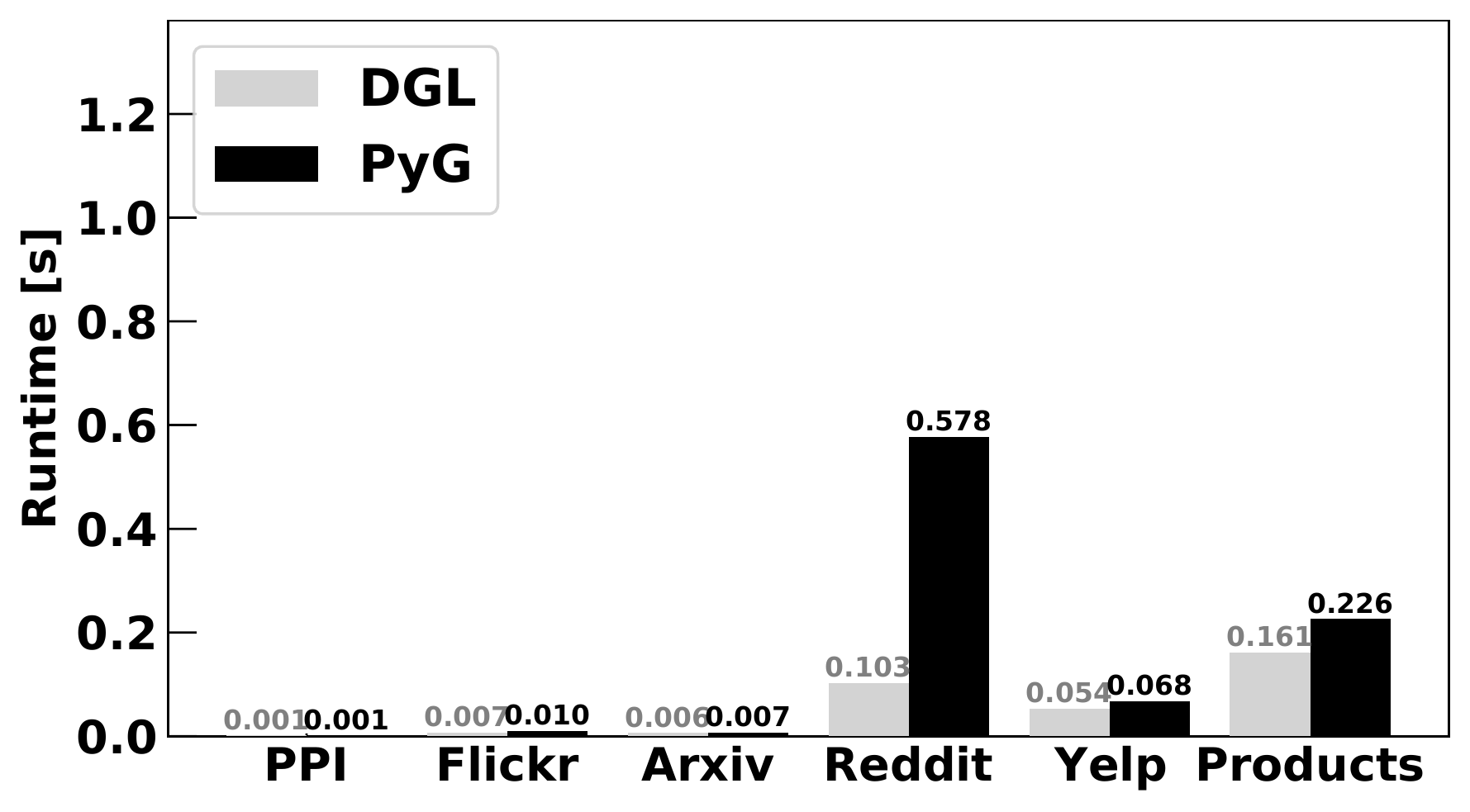}
	}
	\hspace{0mm}
	\centering
	\subfloat[ChebConv-CPU]{%
		\includegraphics[width=0.23\linewidth, trim=0cm 0cm 0cm 0cm, clip]{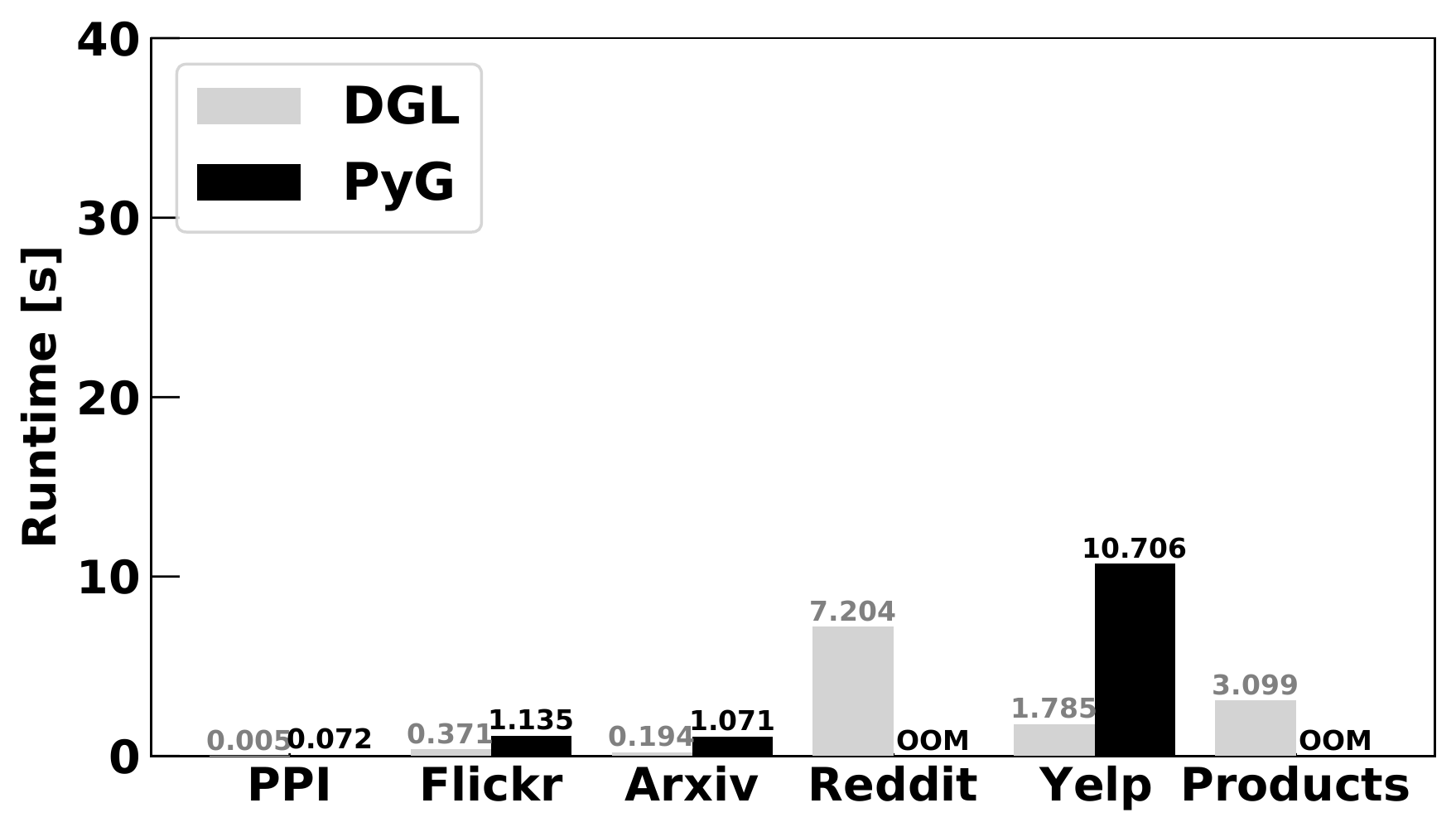}
	}
	\hspace{0mm}
	\subfloat[ChebConv-GPU]{%
		\includegraphics[width=0.23\linewidth, trim=0cm 0cm 0cm 0cm, clip]{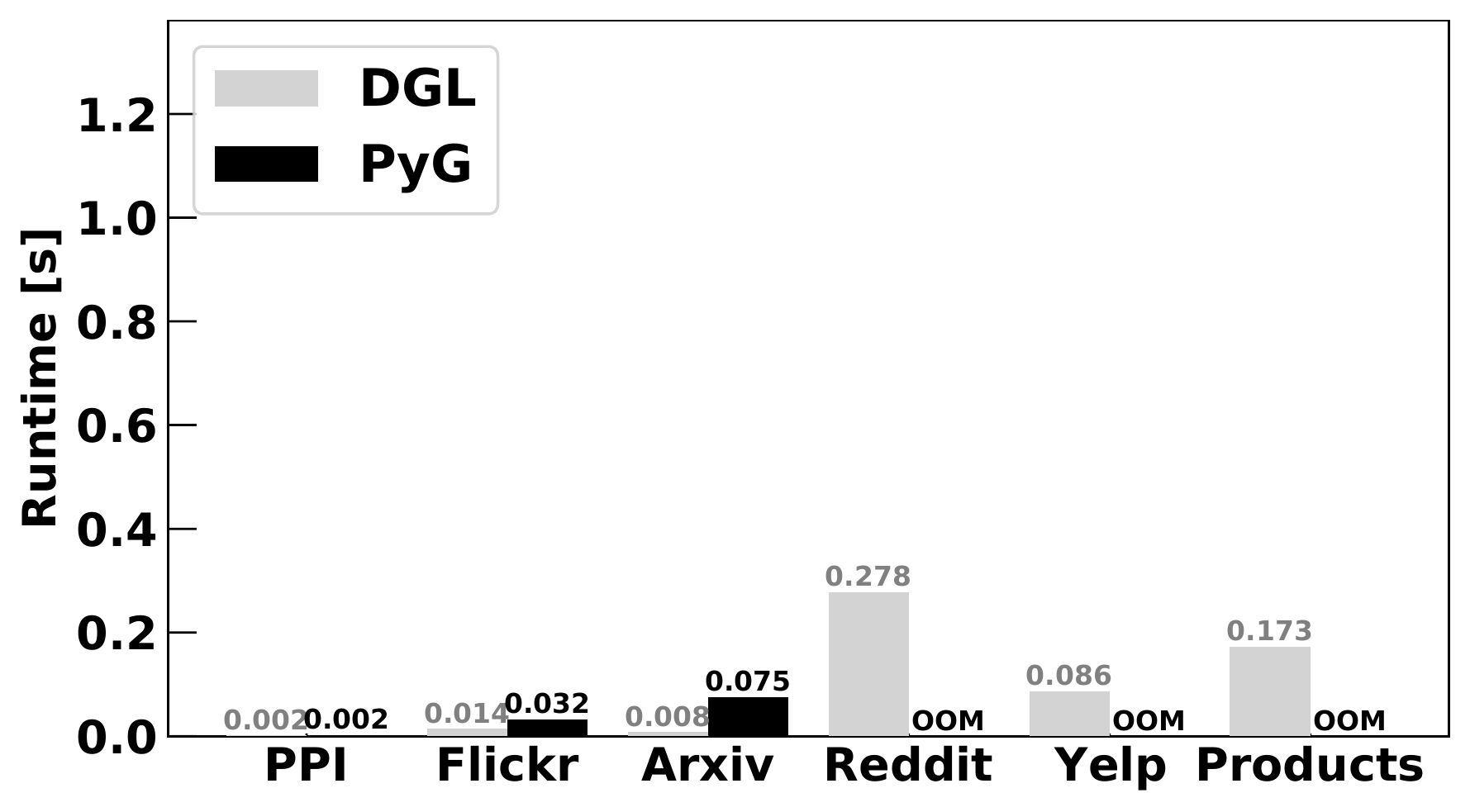}
	}
	\vspace{-3mm}
	\centering
	\subfloat[TAGConv-CPU]{%
		\includegraphics[width=0.23\linewidth, trim=0cm 0cm 0cm 0cm, clip]{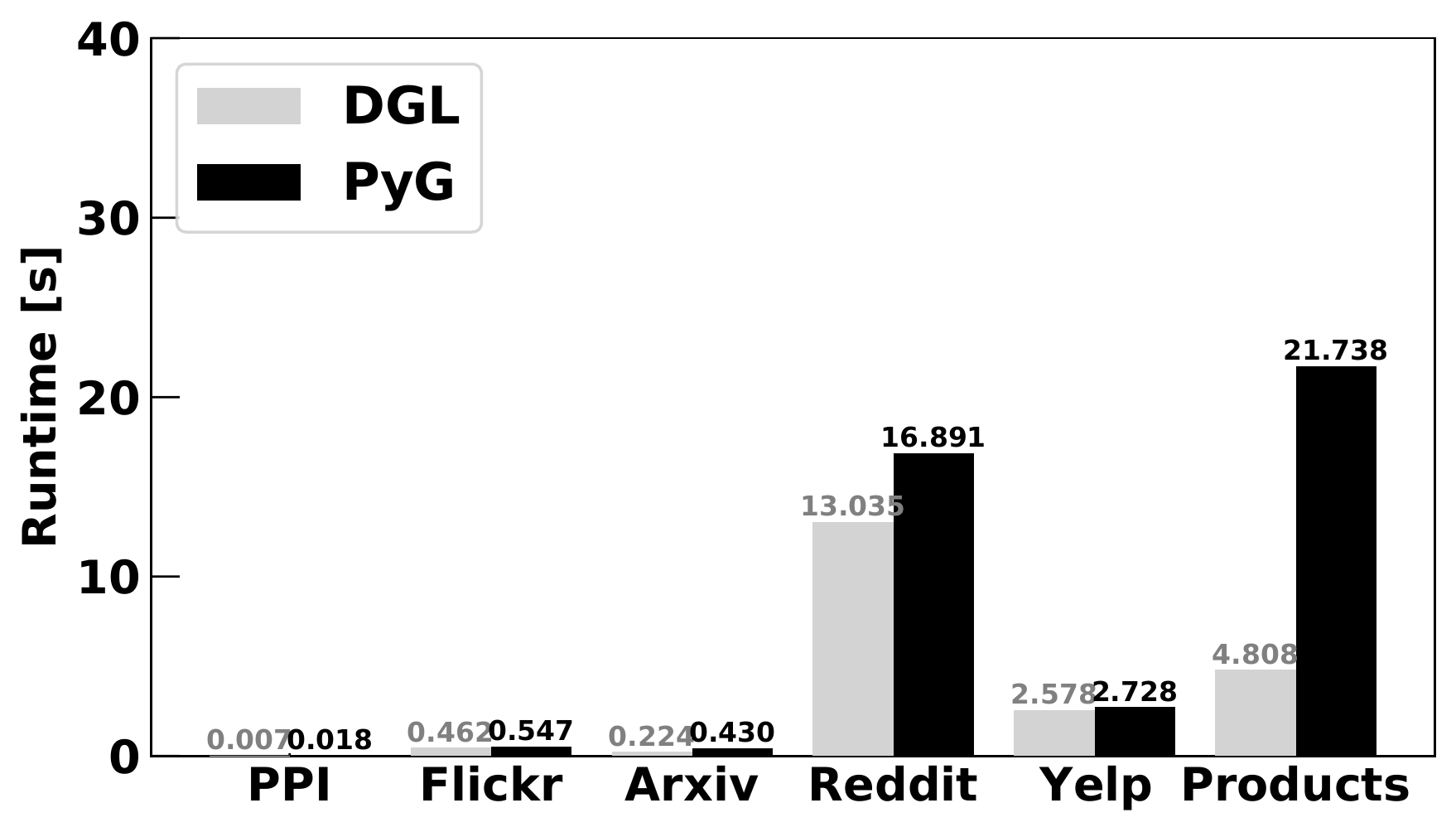}
	}
	\hspace{0mm}
	\subfloat[TAGConv-GPU]{%
		\includegraphics[width=0.23\linewidth, trim=0cm 0cm 0cm 0cm, clip]{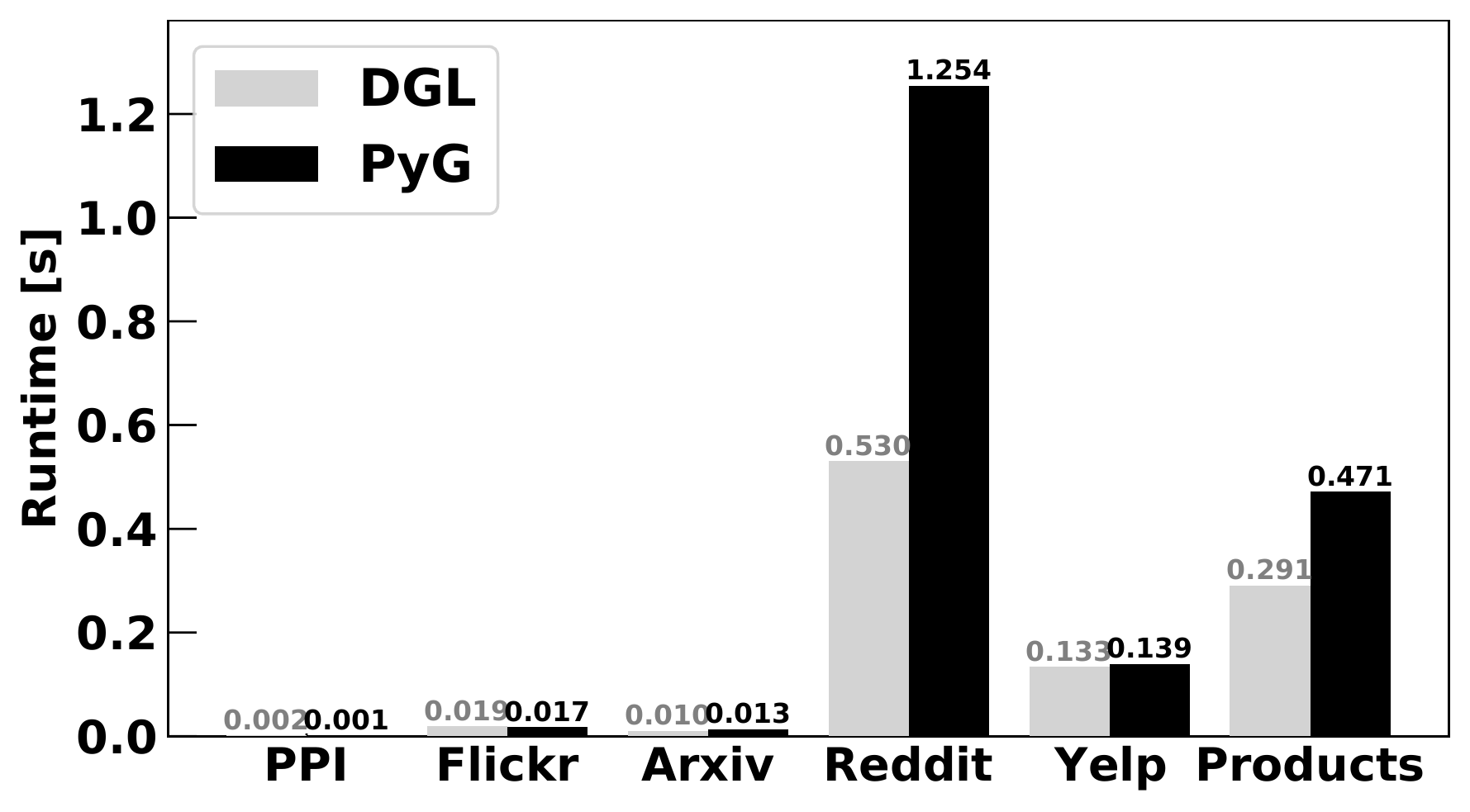}
	}
	\hspace{0mm}
	\subfloat[SGConv-CPU]{%
		\includegraphics[width=0.23\linewidth, trim=0cm 0cm 0cm 0cm, clip]{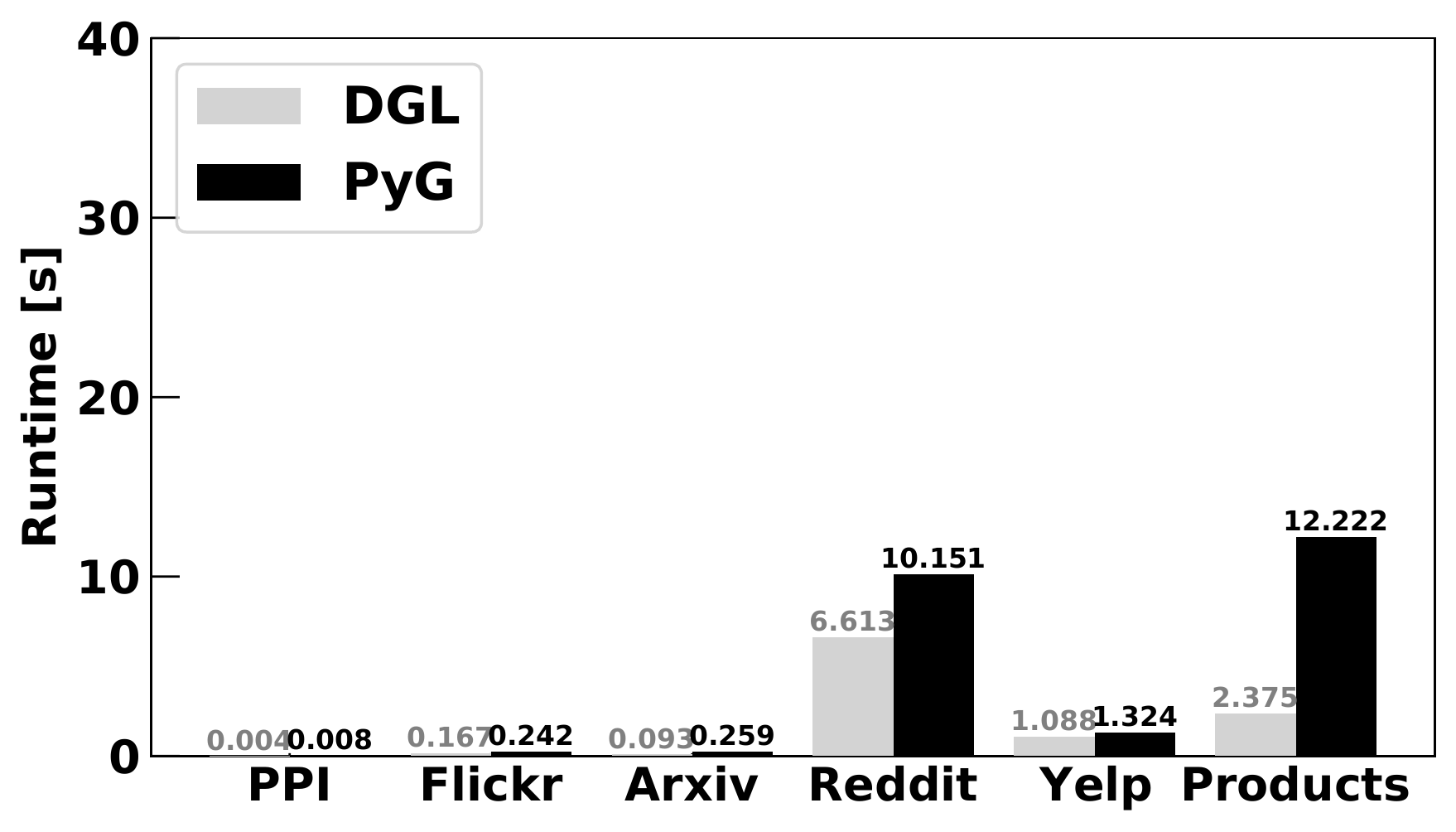}
	}
	\hspace{0mm}
	\subfloat[SGConv-GPU]{%
		\includegraphics[width=0.23\linewidth, trim=0cm 0cm 0cm 0cm, clip]{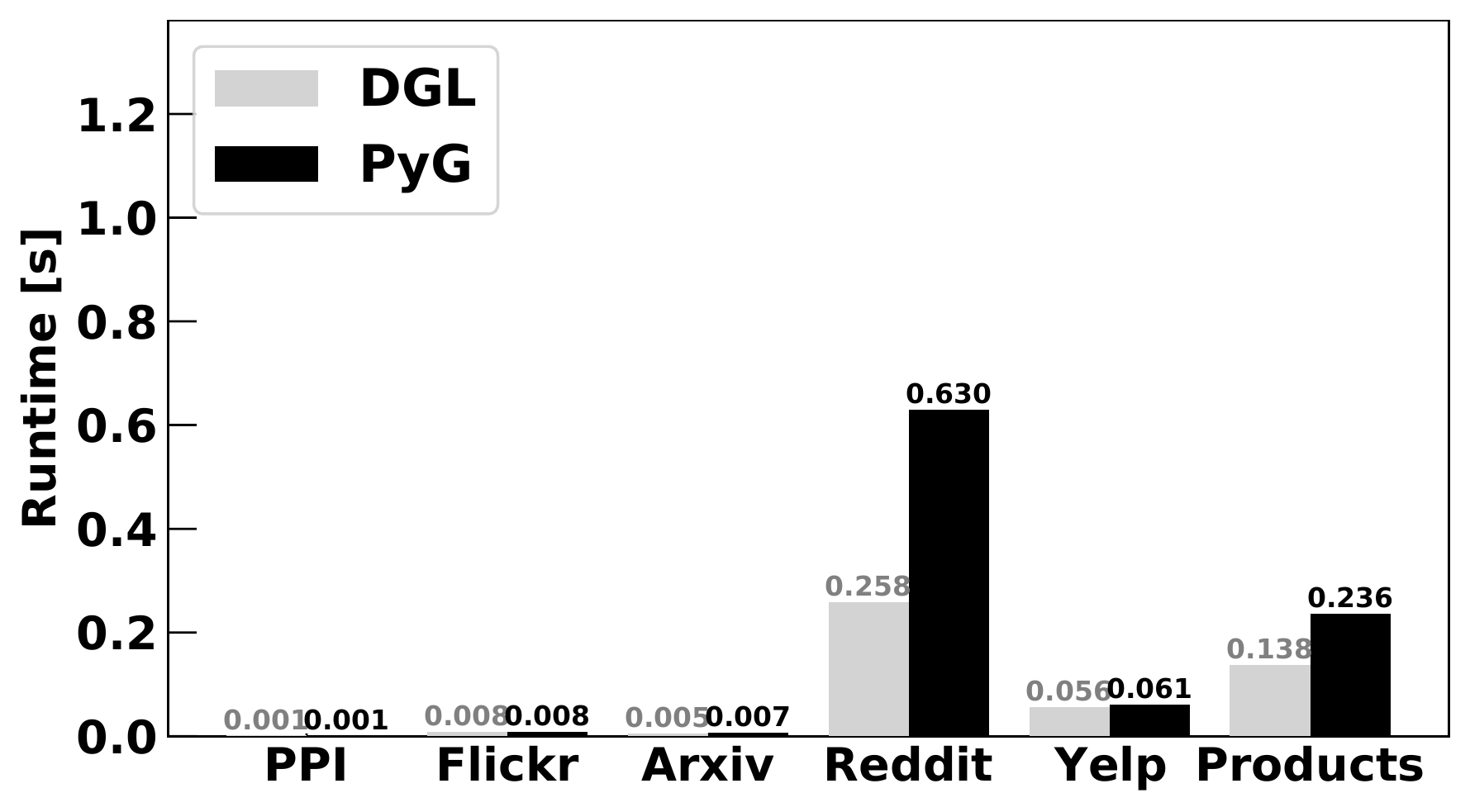}
	}
	\vspace{0mm}
	\caption{Runtime of eight Conv layers. Note that the range of y-axis is different for CPU and GPU cases.}
	\label{fig:Conv}
	\vspace{-3mm}
\end{figure*}

\vspace{1mm}
\noindent \textbf{Sampler.} We then compare the performance of three different graph samplers provided by DGL and PyG, namely neighborhood sampler in GraphSAGE~\cite{hamilton2017inductive}, graph clustering-based sampler in ClusterGCN~\cite{chiang2019cluster}, and random walk-based sampler in GraphSAINT~\cite{zeng2019graphsaint}. 

For GraphSAGE sampler, we follow the settings in~\cite{hamilton2017inductive}, which sample 25 and 10 neighbors per node in its first-hop and second-hop neighborhoods, respectively, with a batch size of 512. Note that each mini-batch is composed of 512 subgraphs. For ClusterCGN sampler, there are two steps, which are (1) graph partitioning with METIS algorithm and (2) cluster aggregation. The former partitions the input graph into a given number of small clusters with METIS algorithm, while the latter is to randomly select a few of them to form a subgraph for a training batch. Note that the former is done only once, but the latter is repeated to obtain different mini-batches. In this experiment, we partition the input graph into 2000 clusters and combine 50 of them for each mini-batch. For GraphSAINT sampler, we use the random walk sampling method with 3000 roots and a walk length of two steps to construct subgraphs from the input graph for mini-batch training. While there are two other sampling methods, namely node sampling and edge sampling, in GraphSAINT, we here do not consider them as they are shown to be inferior to the random walk sampling~\cite{zeng2019graphsaint}. We measure the runtime of each sampler for one training epoch, i.e., one pass over the entire graph, and report the results in Figure~\ref{fig:sampler}.

\vspace{1mm}

\noindent \textbf{Observation 2:} \textit{All three samplers provided by DGL are more efficient than the ones in PyG. The performance gap is relatively small for GraphSAINT sampler since it is computationally cheaper than the other two samplers.}

\vspace{1mm}

We observe that DGL implements its samplers in C++ with OpenMP, thus leading to superior performance to the ones of PyG, which are developed in Python. In addition, although the choices of hyperparameters can affect the sampling performance, GraphSAINT sampler is generally faster than GraphSAGE's neighborhood sampler and ClusterGCN sampler. It is also worth noting that the neighborhood sampler can lead to a very large computational graph for each node, while the ClusterGCN sampler can lead to information loss and data imbalance. Thus, we expect that the GraphSAINT sampler is a preferable choice in practice. Furthermore, we observe that PyG requires data format conversion to the compressed sparse column (CSC) format, e.g., if it was in the compressed sparse row (CSR) format, which turns out to be quite slow on large datasets. Finally, while all three samplers in both DGL and PyG run on CPU, DGL also provides GPU support and CUDA-Unified Virtual Addressing (UVA) support for GraphSAGE, but not for other GNN models. We shall discuss them in Section~\ref{sec:case-study}.

\vspace{0mm}
\noindent \textbf{Graph convolutional layer.} A convolutional (Conv) layer is a key and dominant component of GNNs, and its runtime performance can often reflect the overall performance. We thus conduct functional testing on a collection of Conv layers available in DGL and PyG. Both frameworks provide an `nn' module that contains the implementations of popular Conv layers.  We notice that PyG covers more than 50 Conv layers and DGL has about 30 of them. We here select eight commonly used Conv layers for functional testing. They are GCNConv~\cite{kipf2016semi}, GCN2Conv~\cite{chen2020simple}, ChebConv~\cite{defferrard2016convolutional}, SAGEConv~\cite{hamilton2017inductive}, GATConv~\cite{velivckovic2017graph}, GATv2Conv~\cite{brody2021attentive}, TAGConv~\cite{du2017topology}, and SGConv~\cite{wu2019simplifying}.

\begin{figure}[t]
	\captionsetup[subfloat]{captionskip=1pt}
	\centering
	\subfloat[DGL-CPU]{%
		\includegraphics[width=0.47\linewidth, trim=0cm 0cm 0cm 0cm, clip]{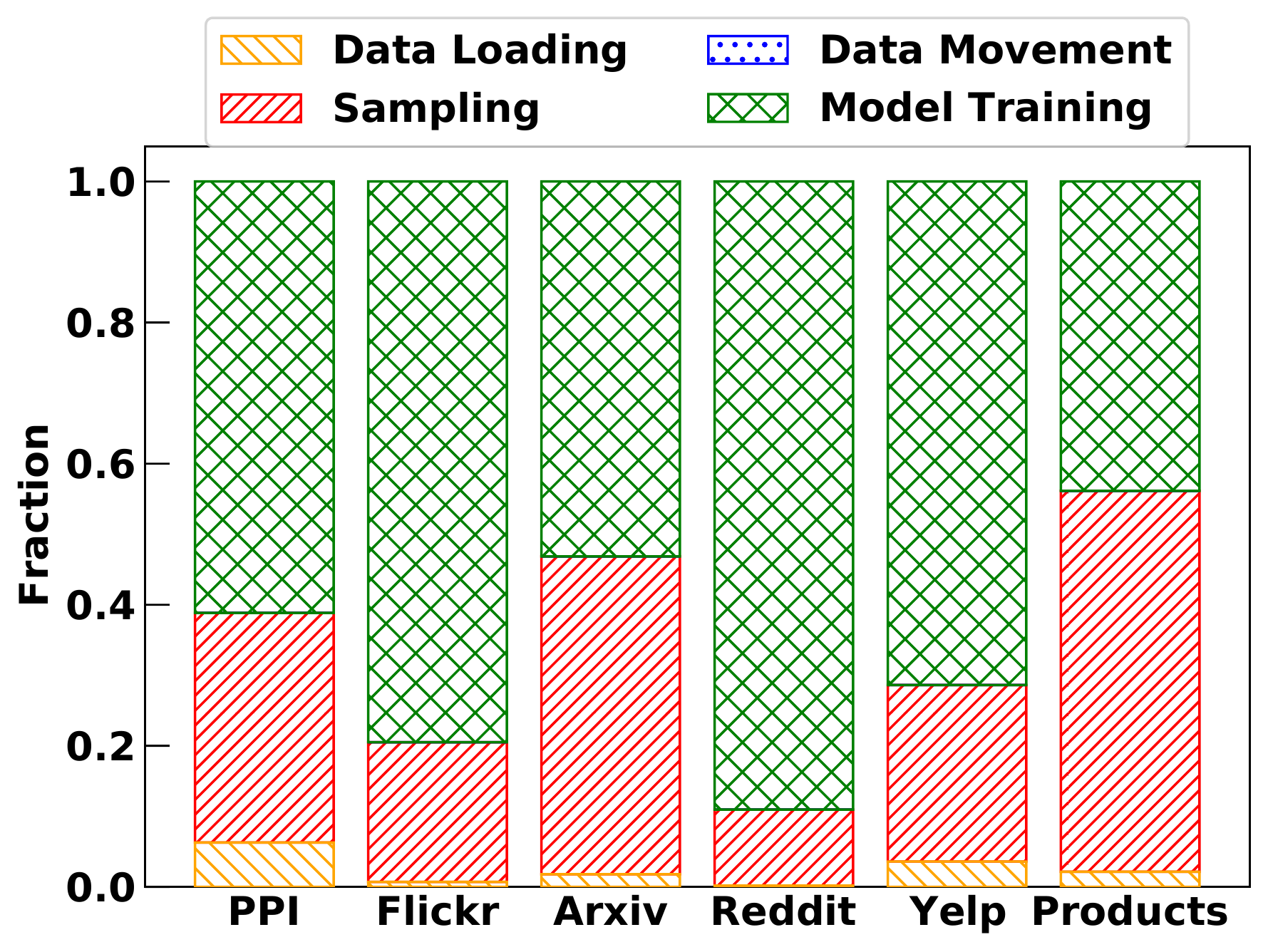}
	}
	\vspace{0mm}
	\subfloat[DGL-CPUGPU]{%
		\includegraphics[width=0.47\linewidth, trim=0cm 0cm 0cm 0cm, clip]{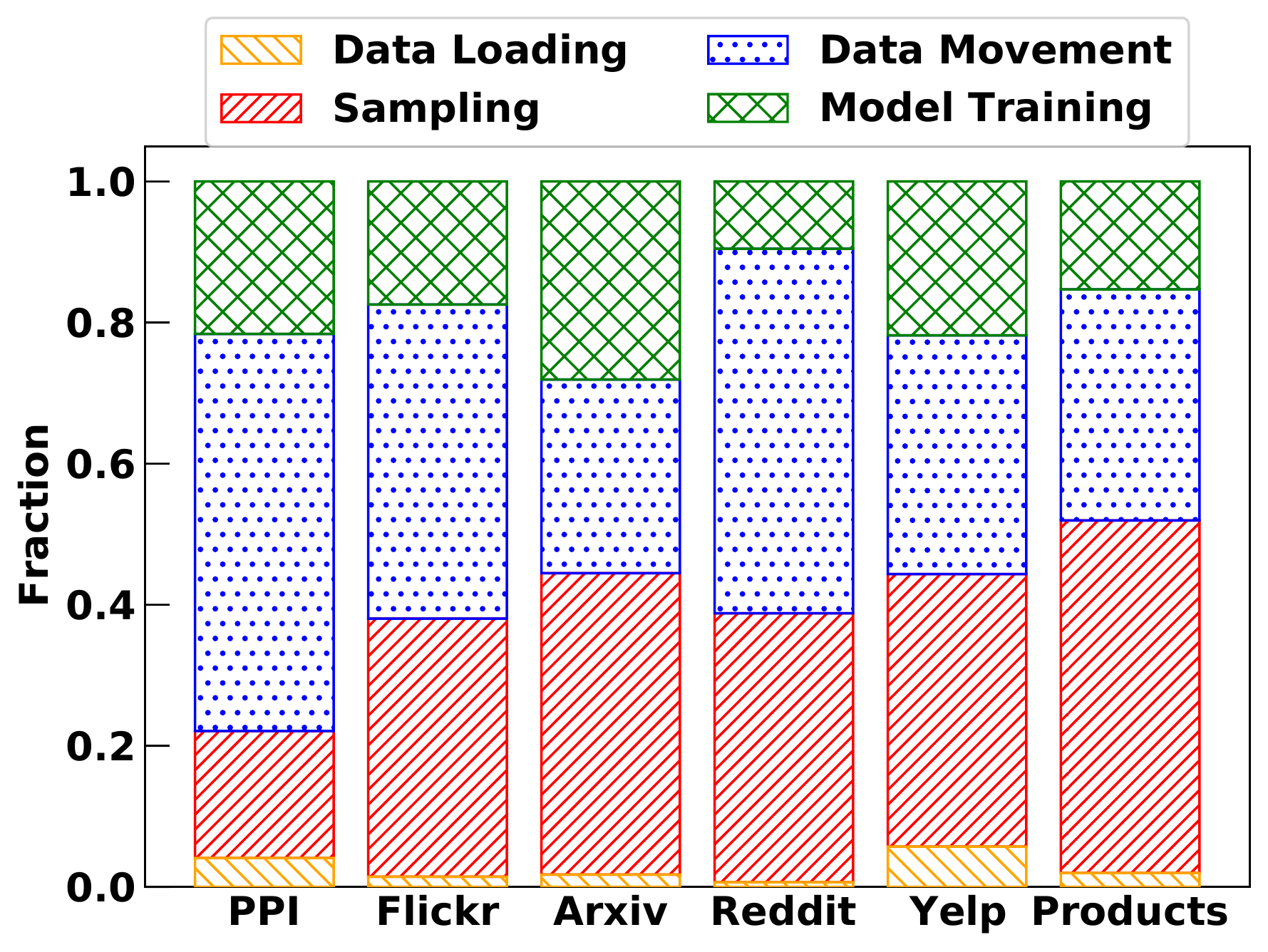}
	}
	\hspace{1mm}
	\subfloat[PyG-CPU]{%
		\includegraphics[width=0.47\linewidth, trim=0cm 0cm 0cm 0cm, clip]{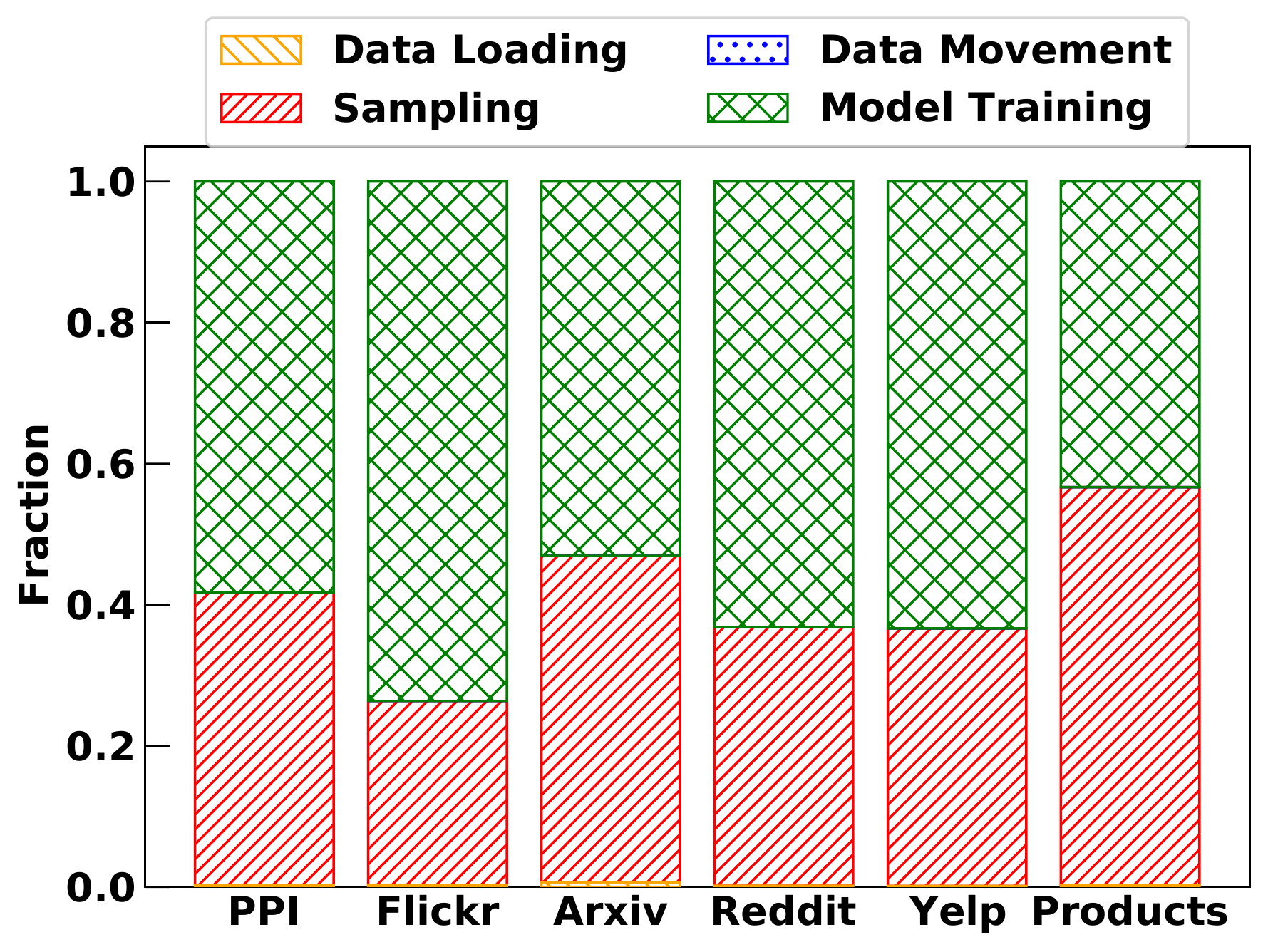}
	}
	\vspace{0mm}
	\subfloat[PyG-CPUGPU]{%
		\includegraphics[width=0.47\linewidth, trim=0cm 0cm 0cm 0cm, clip]{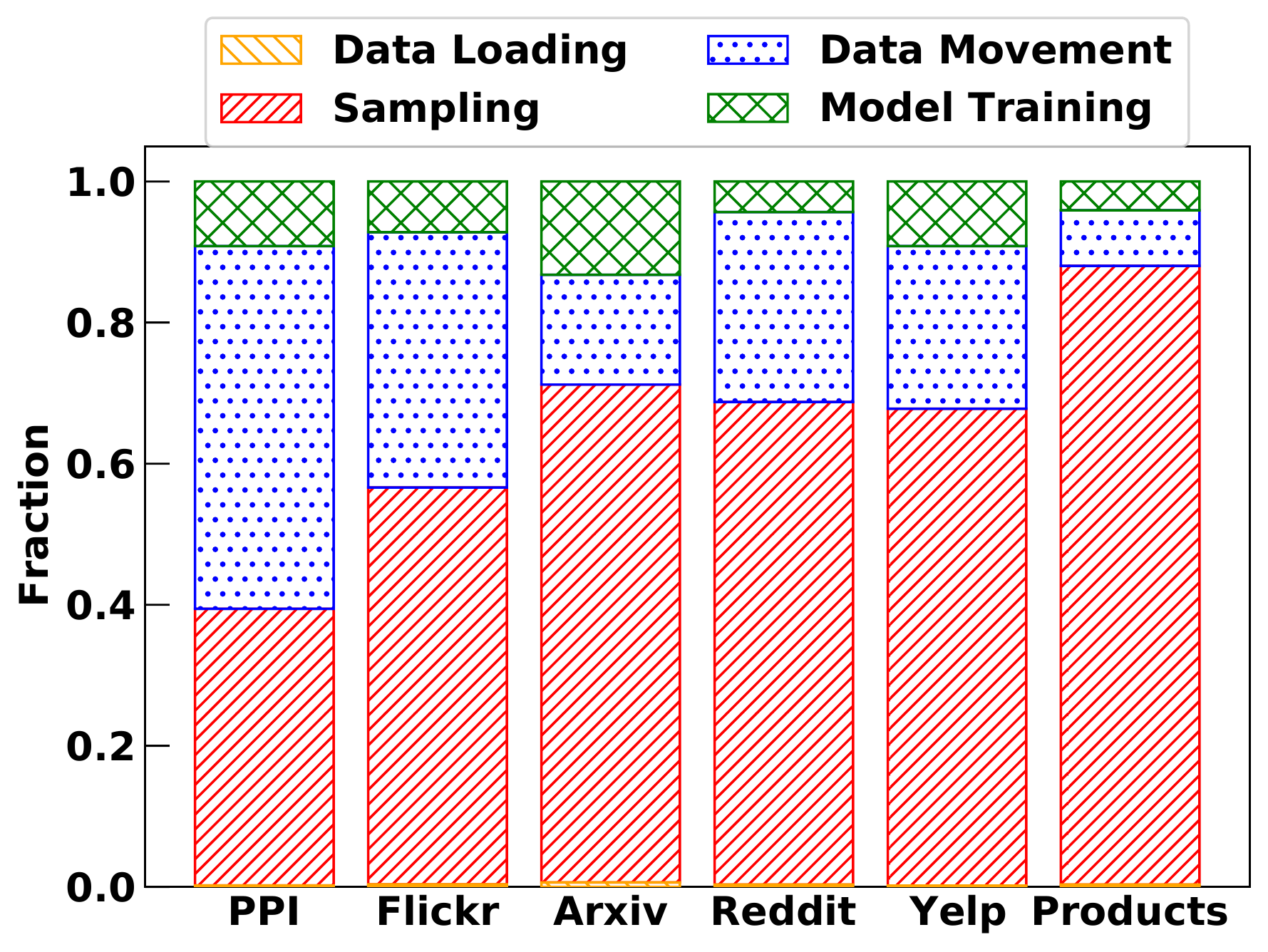}
	}
	\vspace{0mm}
	\caption{Runtime breakdown of GraphSAGE.}
	\label{fig:sage-breakdown}
	\vspace{-2mm}
\end{figure}

We measure the runtime of executing each Conv layer on CPU and GPU. In other words, the reported runtime is equivalent to the time of running \emph{one forward propagation} over a single Conv layer with the entire input graph. We manually set the hyperparameters to be the same across the frameworks for each Conv layer. The output dimension is fixed to be 256 for all test cases. The results are presented in Figure~\ref{fig:Conv}.

\vspace{1mm}

\noindent \textbf{Observation 3:} \textit{All eight Conv layers in DGL run faster than the ones of PyG on CPU. The ones in DGL also run faster than their PyG counterparts on GPU in most cases, while PyG only outperforms DGL for few cases with small graphs. Furthermore, graph convolutional operations on GPU show up to 70x speedup over them on CPU.}

\vspace{1mm}

The main reason for the performance on CPU is that DGL adopts an improved CPU message passing kernel developed by~\cite{md2021distgnn} to boost the performance, while PyG relies on the CPU kernels included in its own PyTorch Sparse and PyTorch Scatter, where some `scatter' operations are not well optimized on CPU. As for the performance on GPU, it is worth noting that our observation does not conflict but match with the observation in \cite{wu2021performance}, which shows that PyG is more efficient than DGL, yet for small graphs. Our observation also confirms the claim in~\cite{wang2019deep}. Although DGL is a bit slower on small graphs due to its framework overhead, it is generally more efficient than PyG, especially on large graphs, thanks to its highly tuned kernels. We also find that SAGEConv is relatively computationally cheaper than the other Conv layers, due to its simple aggregation operation.

In addition, we observe that both frameworks provide \emph{fused} kernels to improve their efficiency and scalability, where two separate message-passing and aggregation operations are merged as a single message aggregation operation. DGL uses `g.update\_all()' function to invoke its g-SpMM and g-SDDMM kernels, while PyG simply calls `matmul()' function in PyTorch Sparse. It is worth noting that PyG does not provide such fused kernel support for ChebConv, GATConv, and GATv2Conv layers. As a result, all three layers of PyG suffer from an out-of-memory issue on large graphs.

\begin{figure}[t]
	\captionsetup[subfloat]{captionskip=1pt}
	\centering
	\subfloat{%
		\includegraphics[width=0.47\linewidth, trim=0cm 0cm 0cm 0cm, clip]{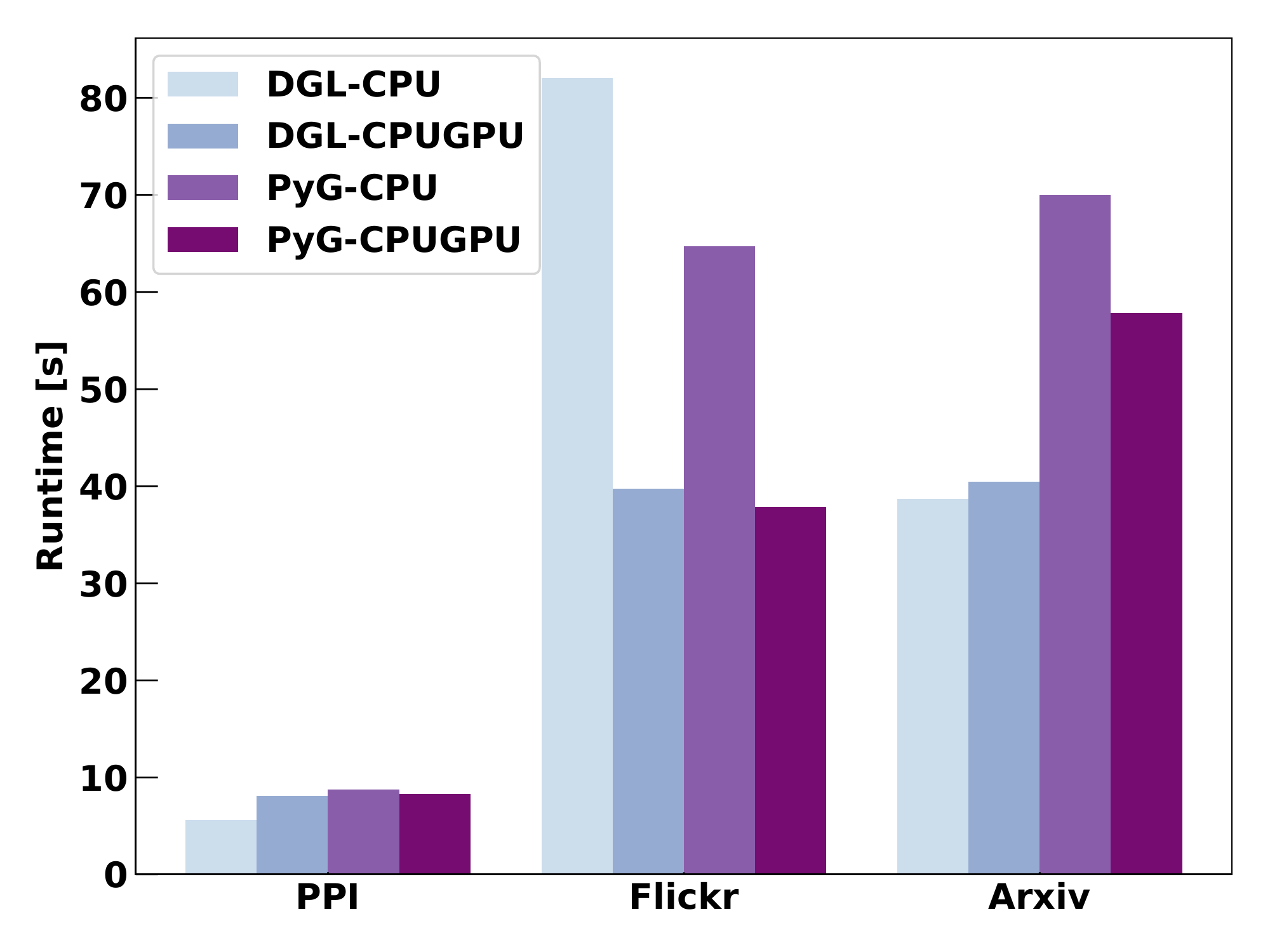}
	}
	\hspace{0mm}
	\subfloat{%
		\includegraphics[width=0.47\linewidth, trim=0cm 0cm 0cm 0cm, clip]{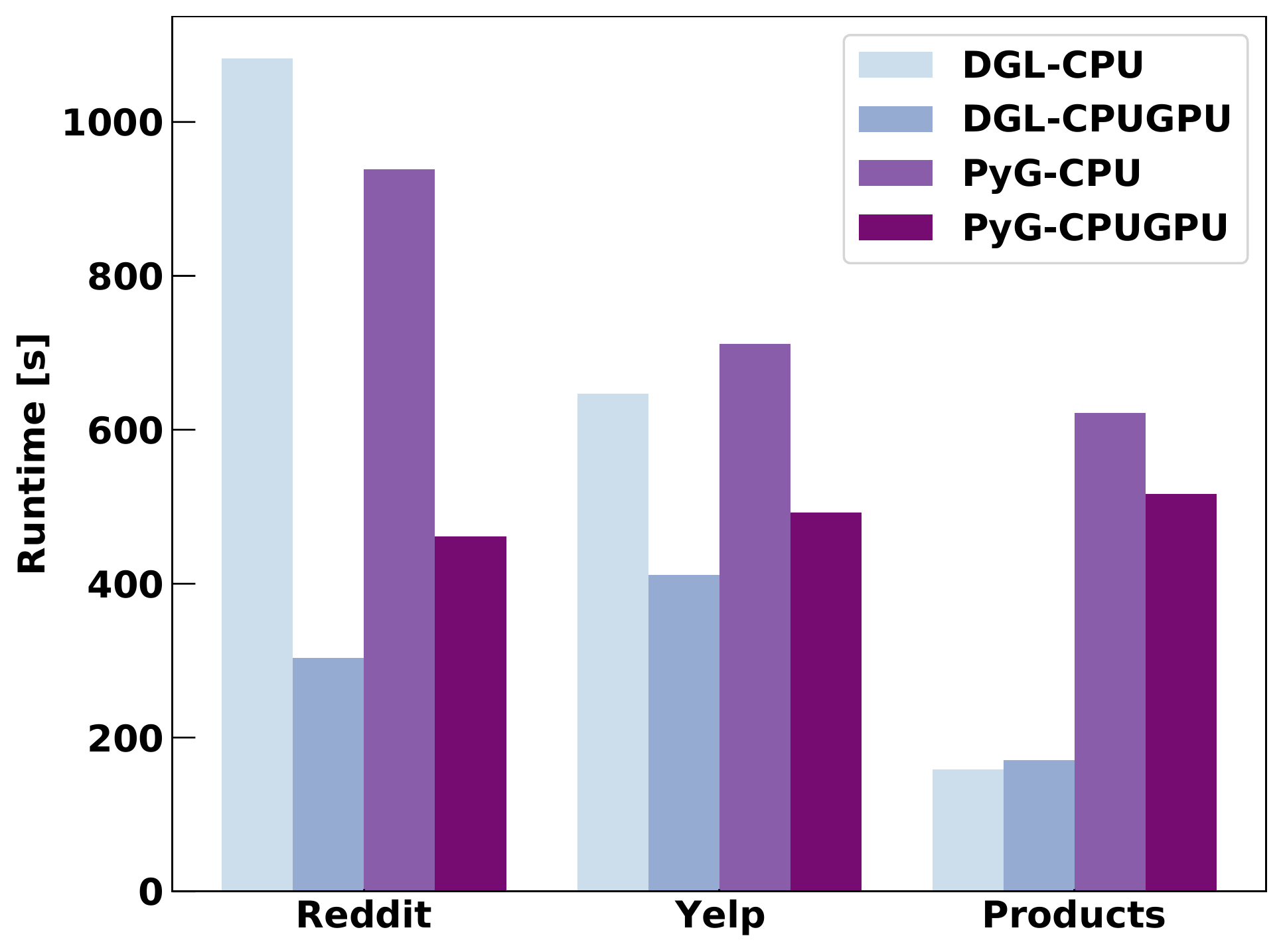}
	}
	\vspace{-2mm}
	\caption{Total runtime of GraphSAGE.}
	\label{fig:sage-runtime}
	\vspace{-3mm}
\end{figure}

\begin{figure}[t!]
	\captionsetup[subfloat]{captionskip=1pt}
	\centering
	\subfloat{%
		\includegraphics[width=0.47\linewidth, trim=0cm 0cm 0cm 0cm, clip]{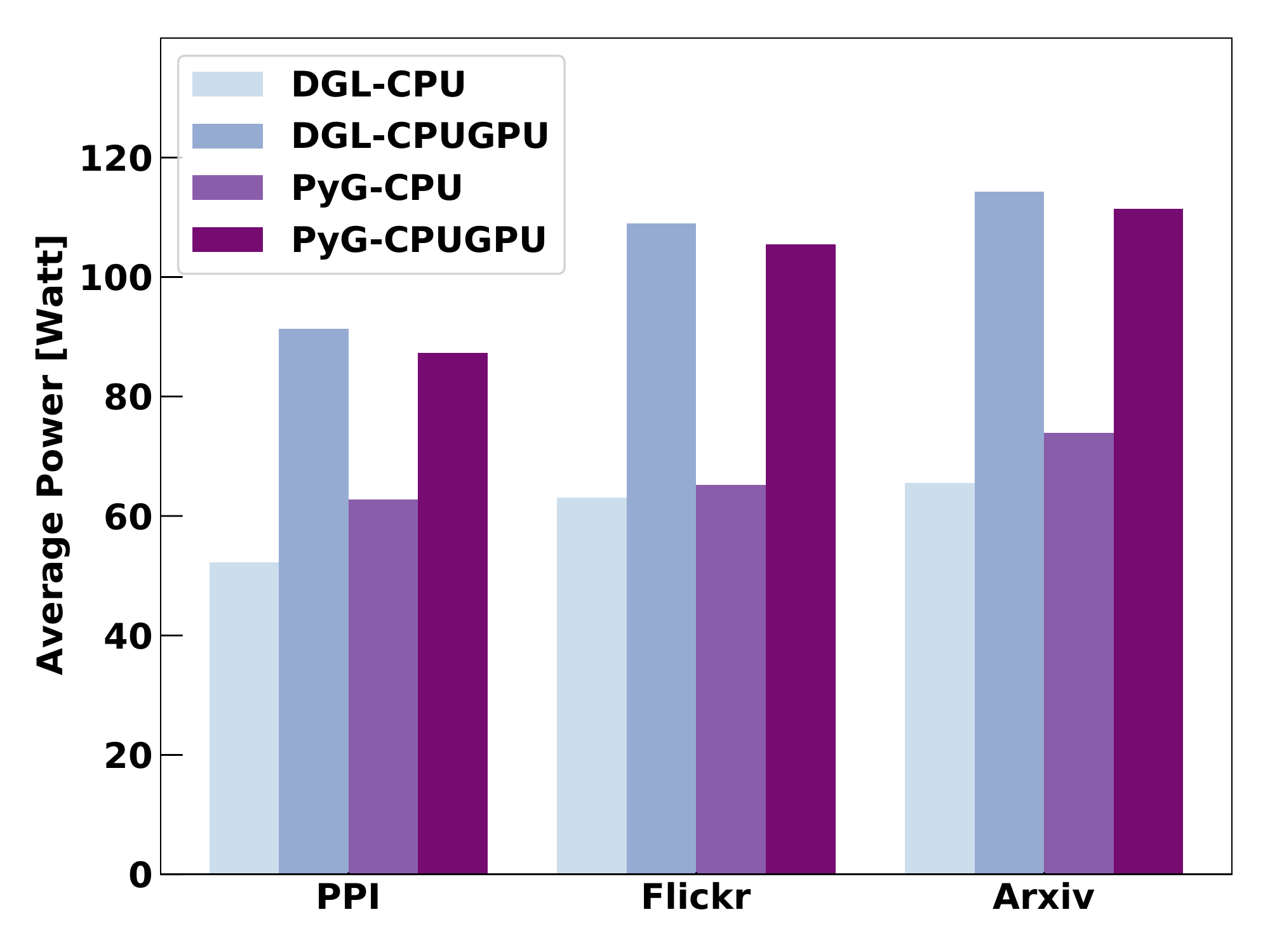}
	}
	\vspace{0mm}
	\subfloat{%
		\includegraphics[width=0.47\linewidth, trim=0cm 0cm 0cm 0cm, clip]{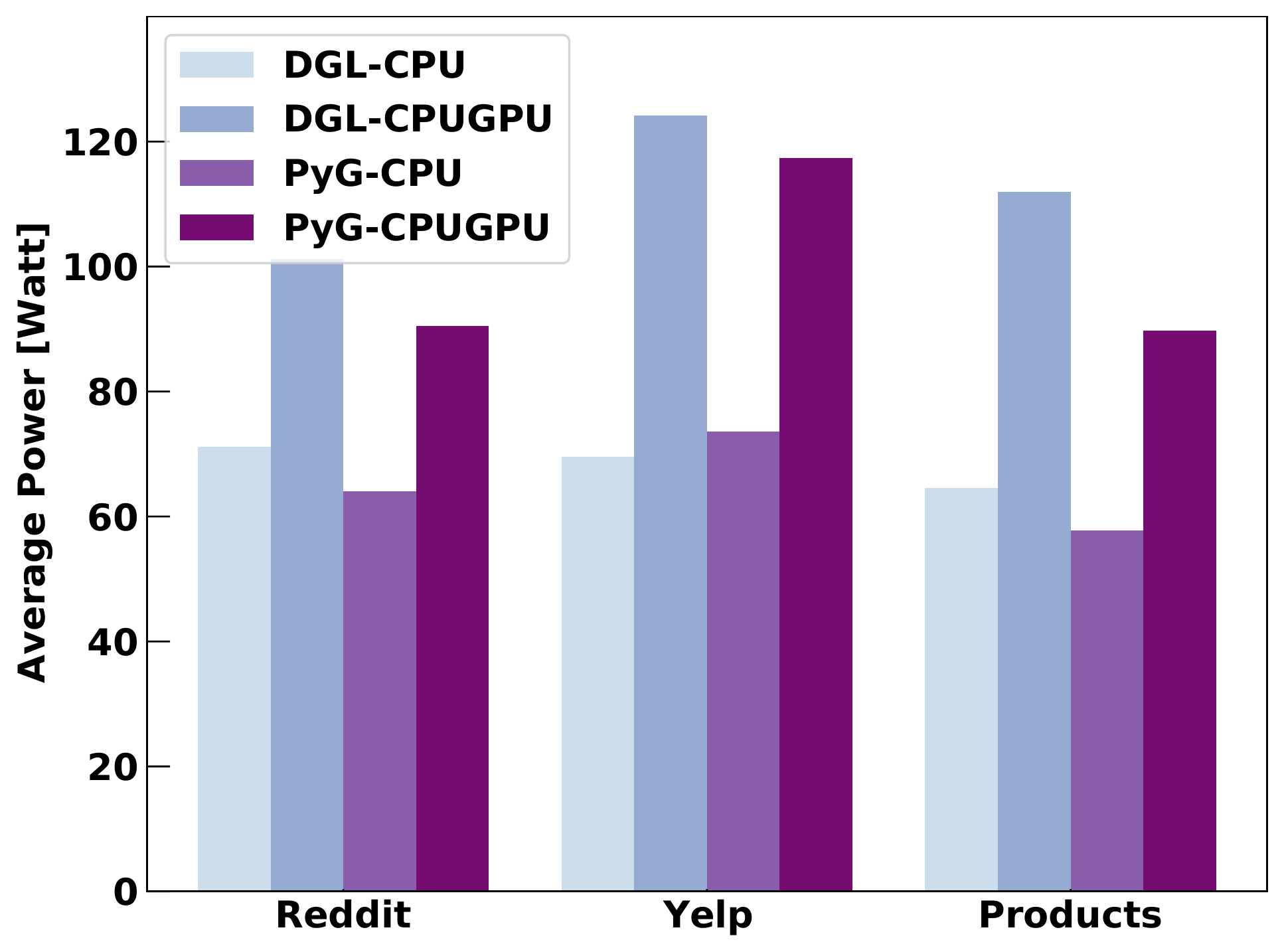}
	}
	\vspace{-2mm}
	\caption{Average power consumption of GraphSAGE.}
	\label{fig:sage-power}
	\vspace{-2mm}
\end{figure}

\begin{figure}[t!]
	\captionsetup[subfloat]{captionskip=1pt}
	\centering
	\subfloat{%
		\includegraphics[width=0.47\linewidth, trim=0cm 0cm 0cm 0cm, clip]{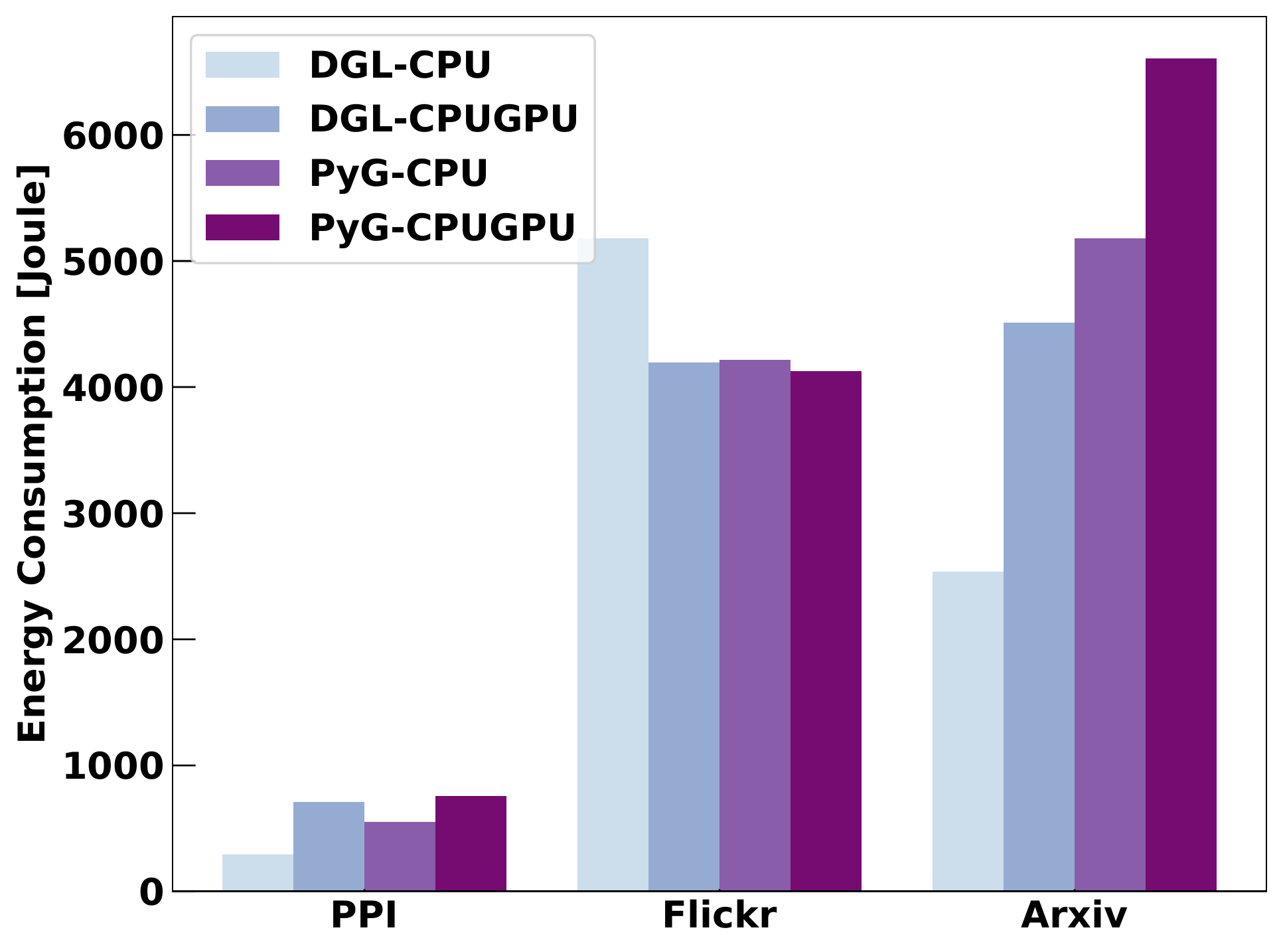}
	}
	\vspace{0mm}
	\subfloat{%
		\includegraphics[width=0.47\linewidth, trim=0cm 0cm 0cm 0cm, clip]{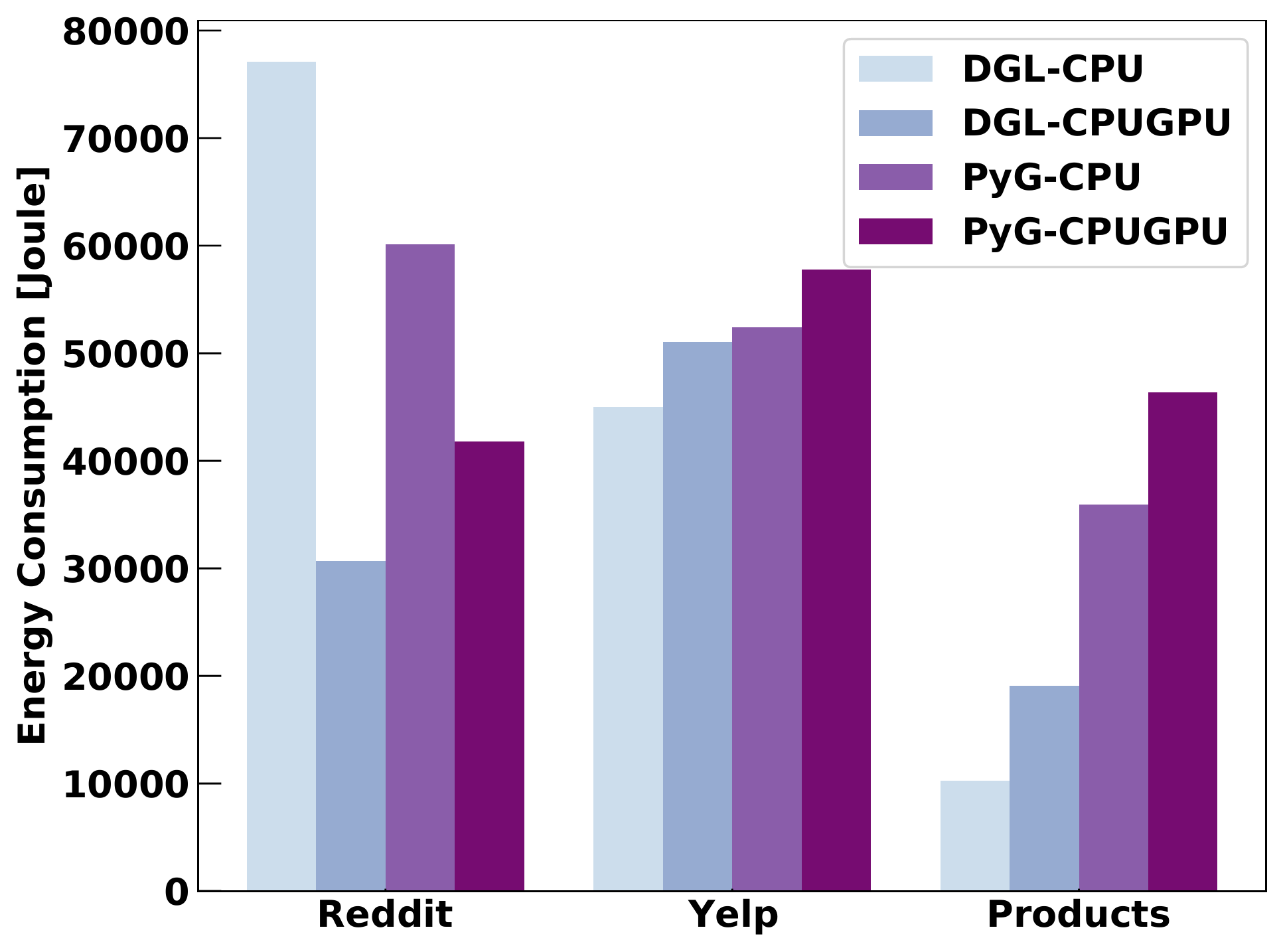}
	}
	\vspace{-2mm}
	\caption{Energy consumption of GraphSAGE.}
	\label{fig:sage-energy}
	\vspace{-3mm}
\end{figure}

\subsection{Performance Evaluation of GNNs}\label{sec:performance}

\begin{figure}[t!]
	\captionsetup[subfloat]{captionskip=1pt}
	\centering
	\subfloat[DGL-CPU]{%
		\includegraphics[width=0.47\linewidth, trim=0cm 0cm 0cm 0cm, clip]{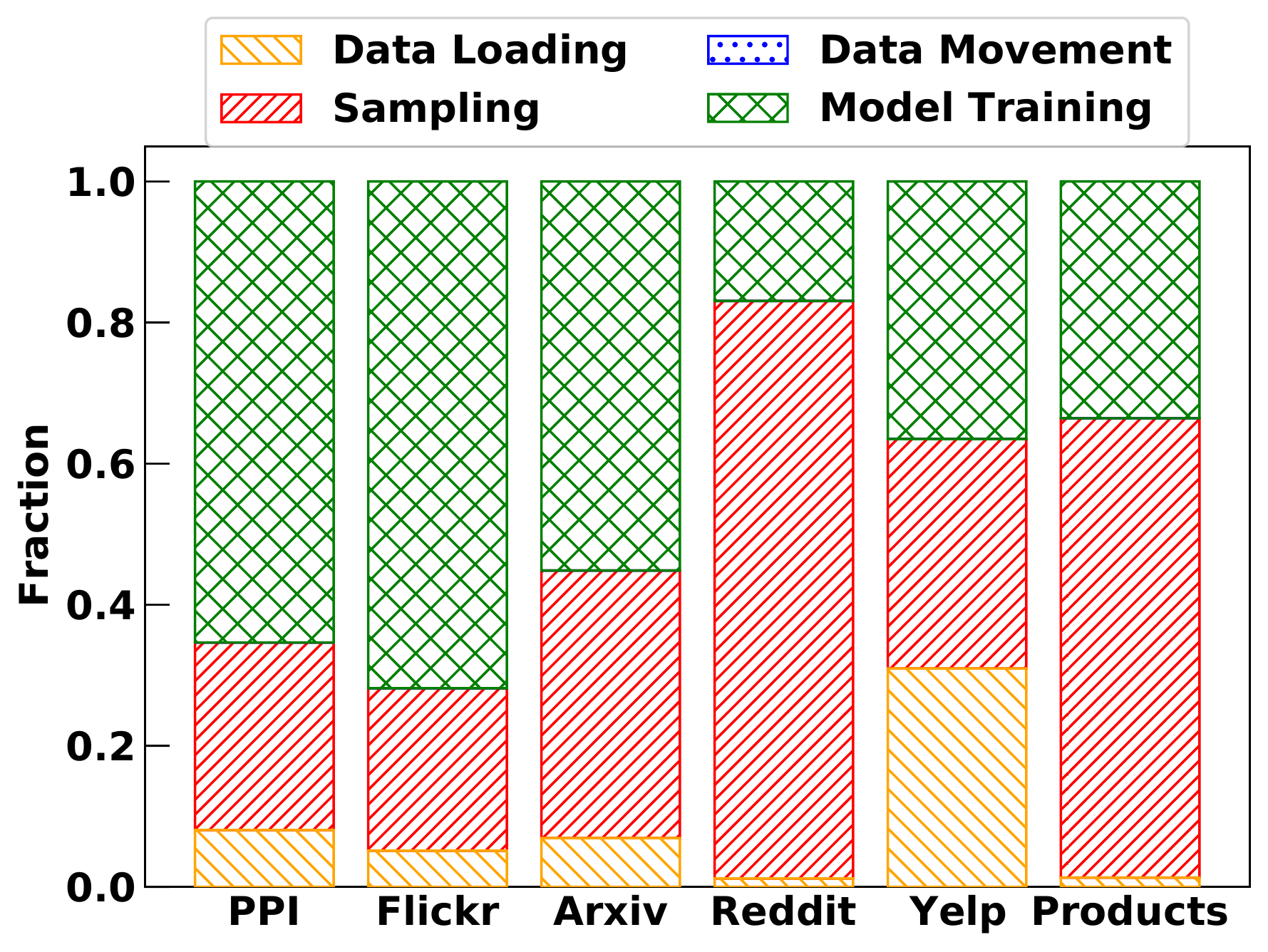}
	}
	\hspace{0mm}
	\subfloat[DGL-CPUGPU]{%
		\includegraphics[width=0.47\linewidth, trim=0cm 0cm 0cm 0cm, clip]{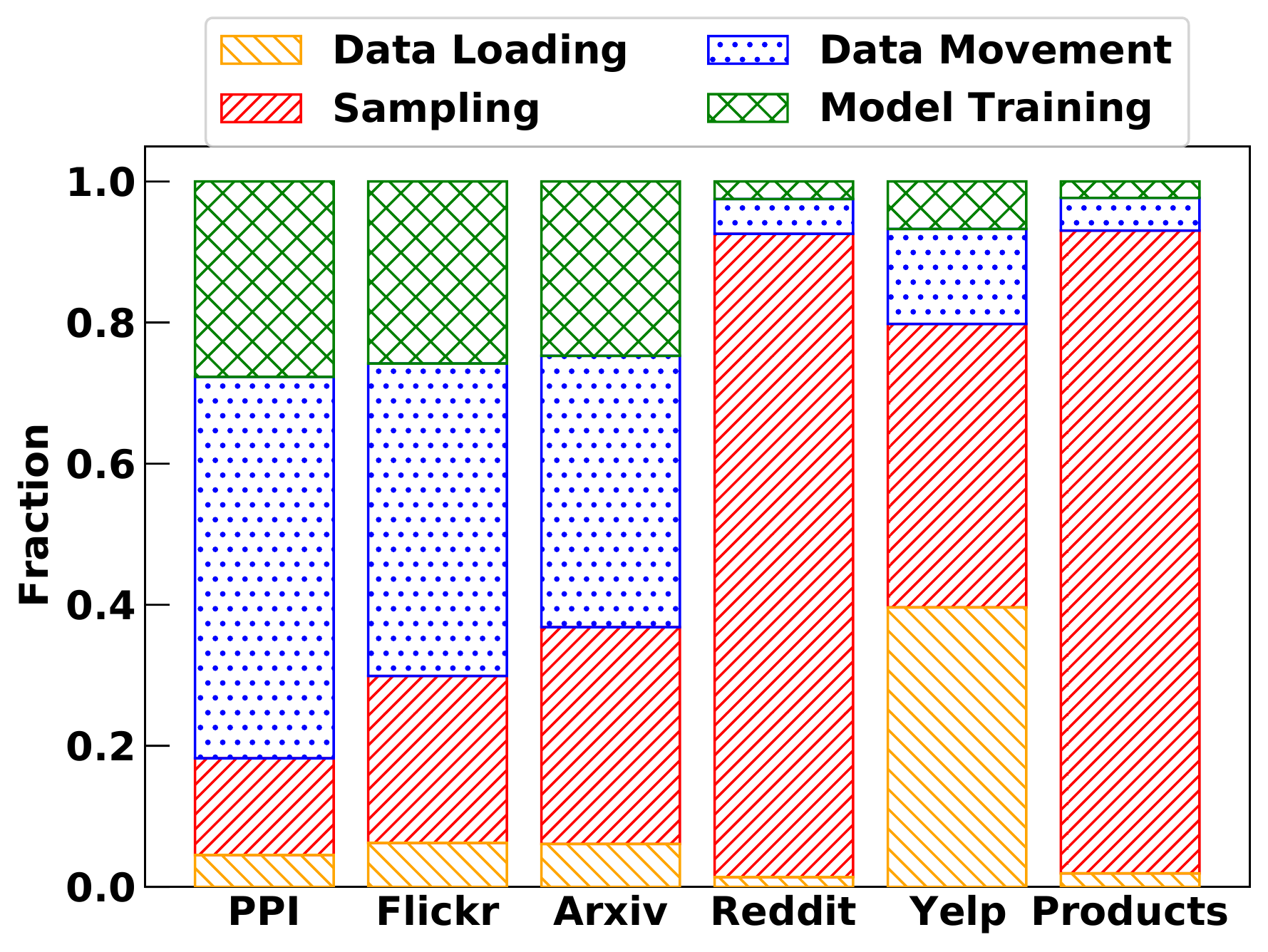}
	}
	\vspace{-2mm}
	\subfloat[PyG-CPU]{%
		\includegraphics[width=0.47\linewidth, trim=0cm 0cm 0cm 0cm, clip]{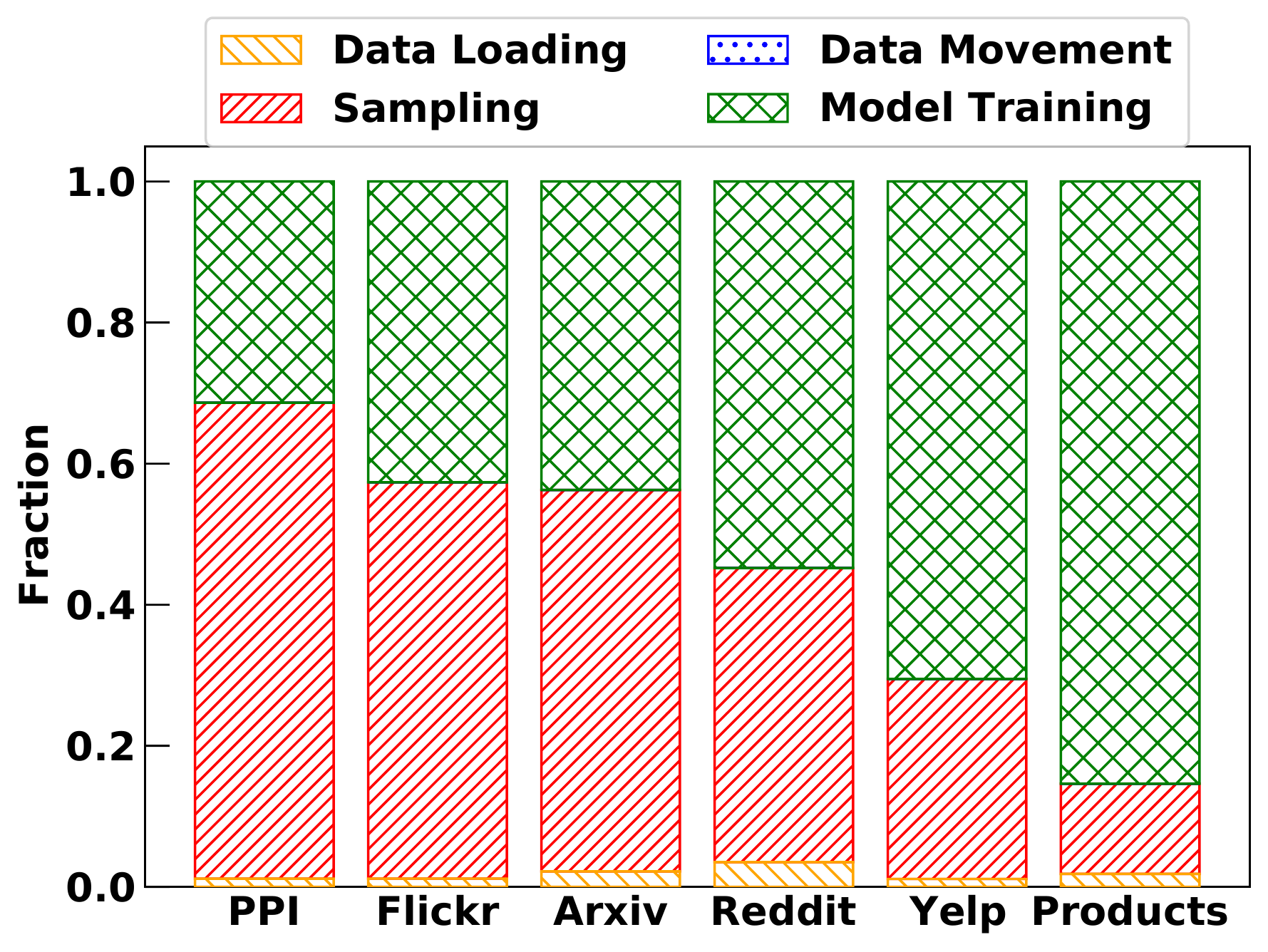}
	}
	\vspace{0mm}
	\subfloat[PyG-CPUGPU]{%
		\includegraphics[width=0.47\linewidth, trim=0cm 0cm 0cm 0cm, clip]{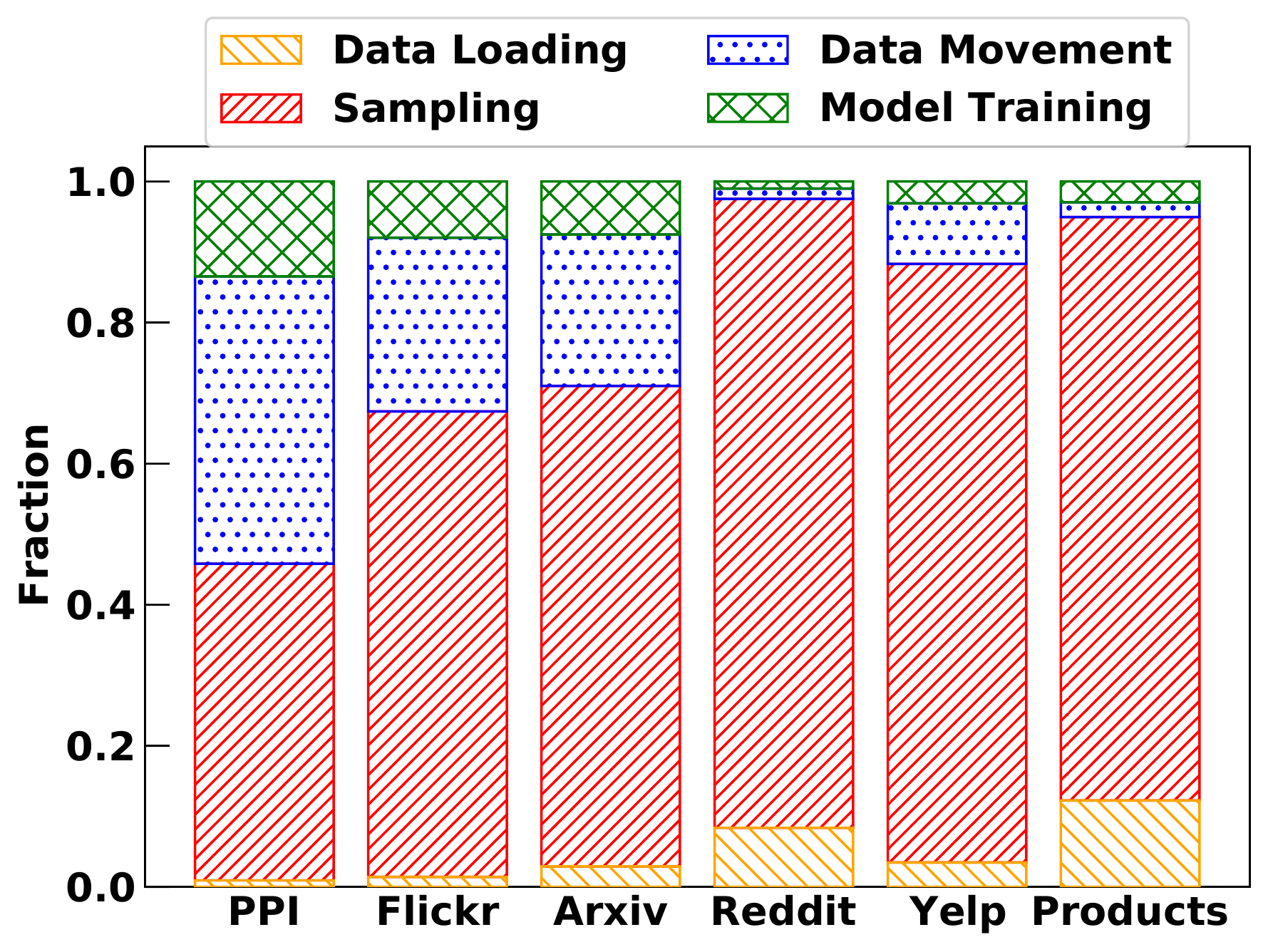}
	}
	\vspace{0mm}
	\caption{Runtime breakdown of ClusterGCN.}
	\label{fig:cluster-breakdown}
	\vspace{-4mm}
\end{figure}

\begin{figure}[t!]
	\captionsetup[subfloat]{captionskip=1pt}
	\centering
	\subfloat{%
		\includegraphics[width=0.47\linewidth, trim=0cm 0cm 0cm 0cm, clip]{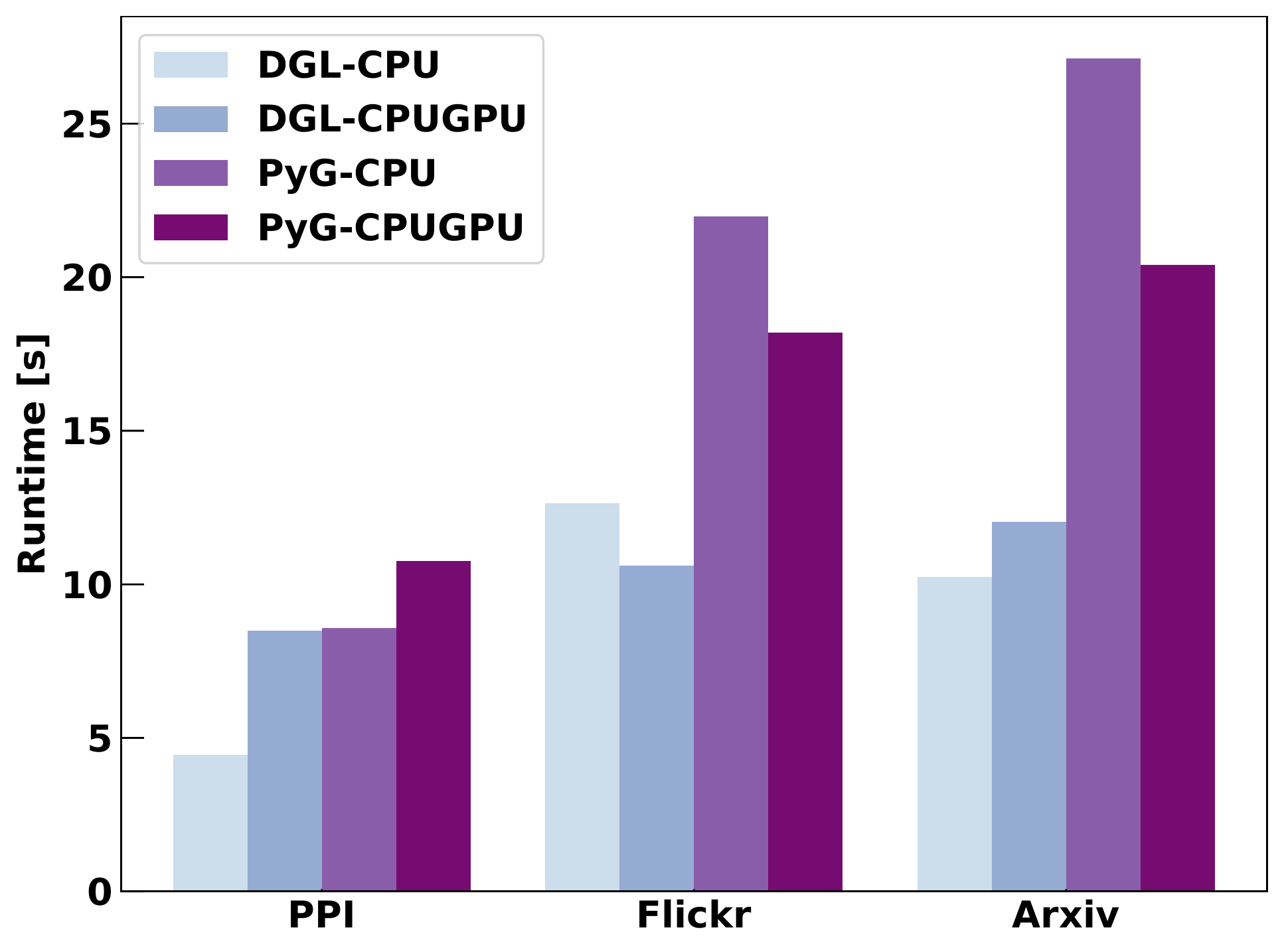}
	}
	\hspace{0mm}
	\subfloat{%
		\includegraphics[width=0.47\linewidth, trim=0cm 0cm 0cm 0cm, clip]{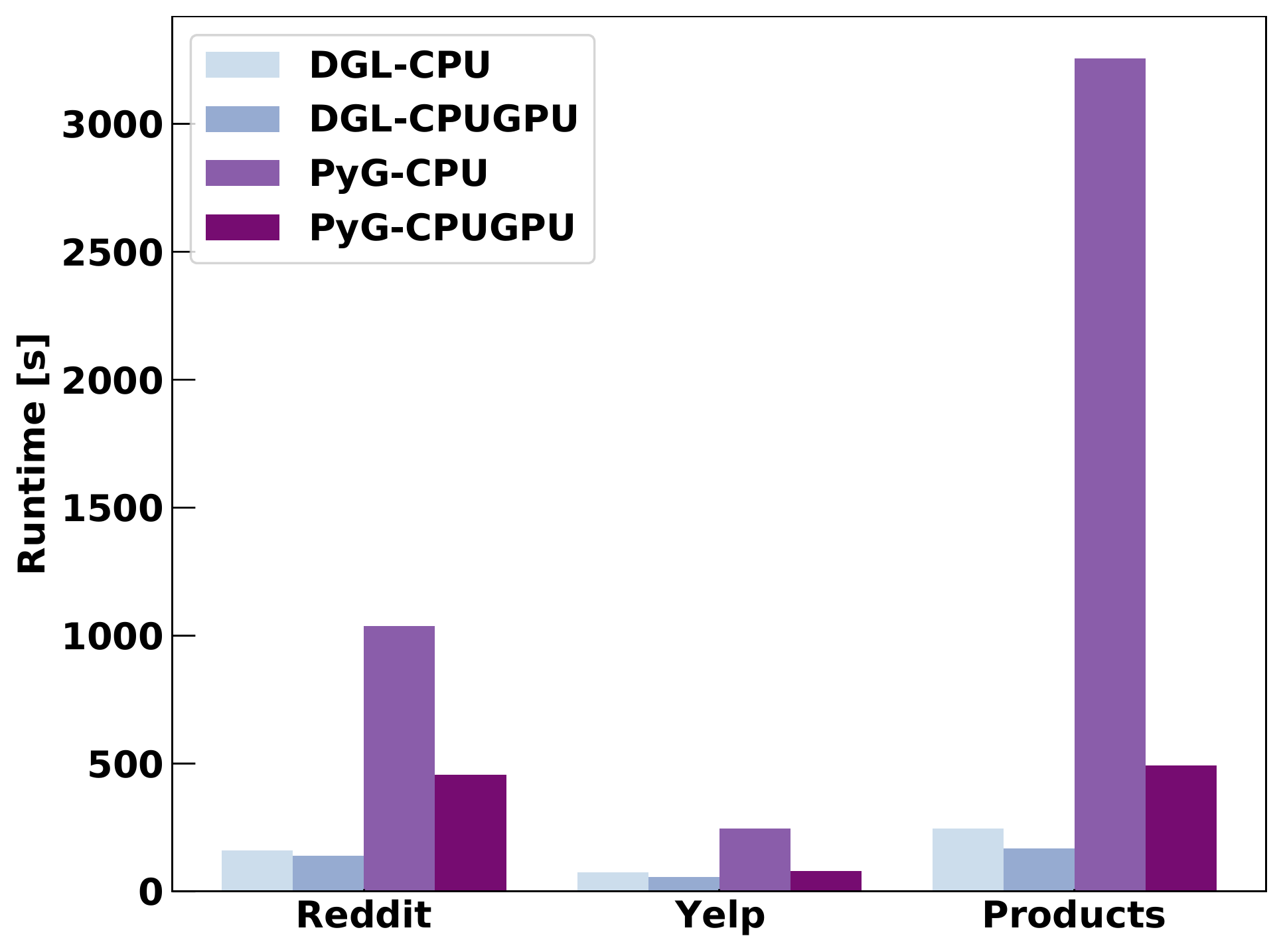}
	}
	\vspace{0mm}
	\caption{Total runtime of ClusterGCN.}
	\label{fig:clustergcn-runtime}
	\vspace{-4mm}
\end{figure}

\begin{figure}[t!]
	\captionsetup[subfloat]{captionskip=1pt}
	\centering
	\subfloat{%
		\includegraphics[width=0.47\linewidth, trim=0cm 0cm 0cm 0cm, clip]{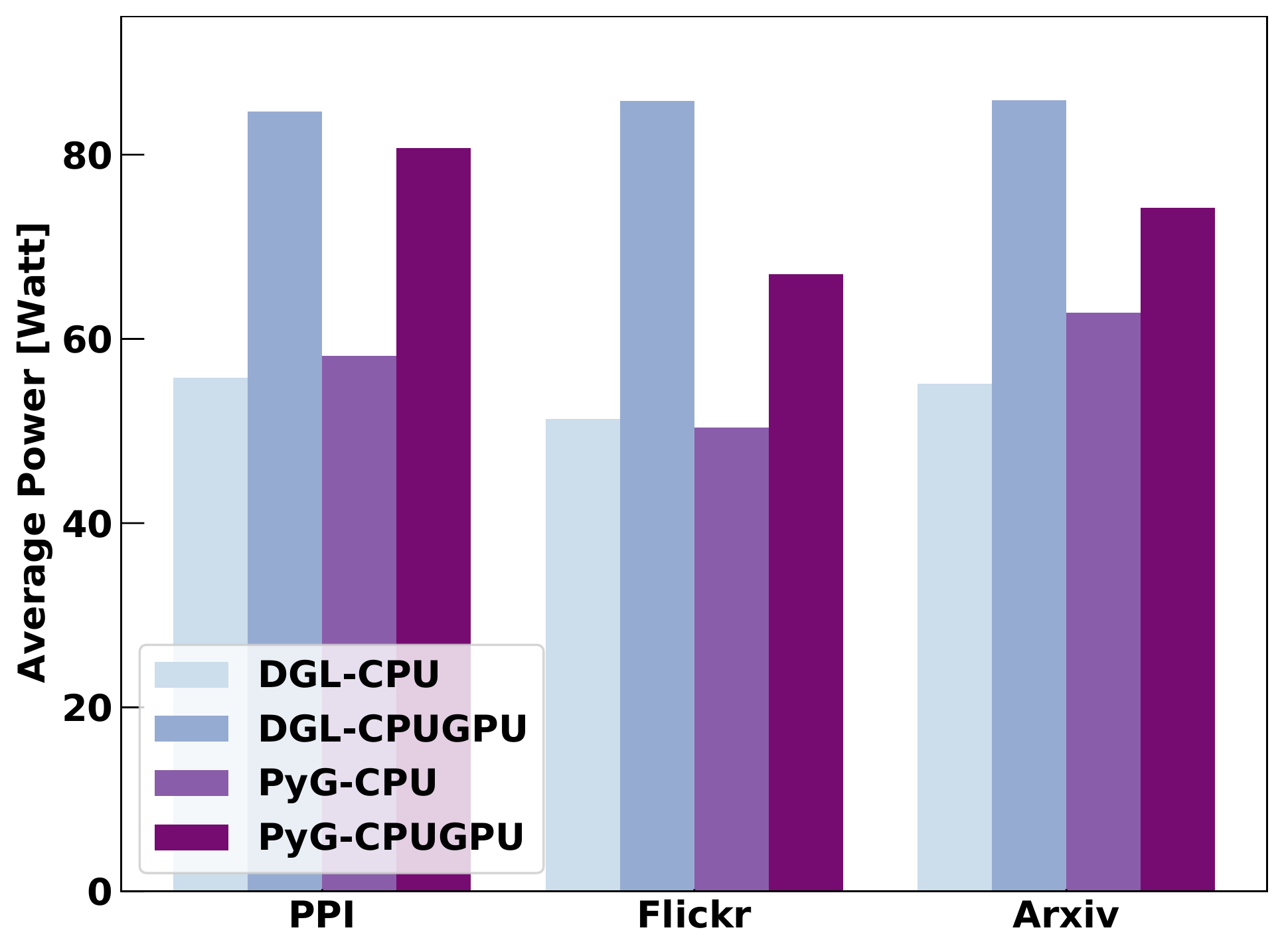}
	}
	\vspace{0mm}
	\subfloat{%
		\includegraphics[width=0.47\linewidth, trim=0cm 0cm 0cm 0cm, clip]{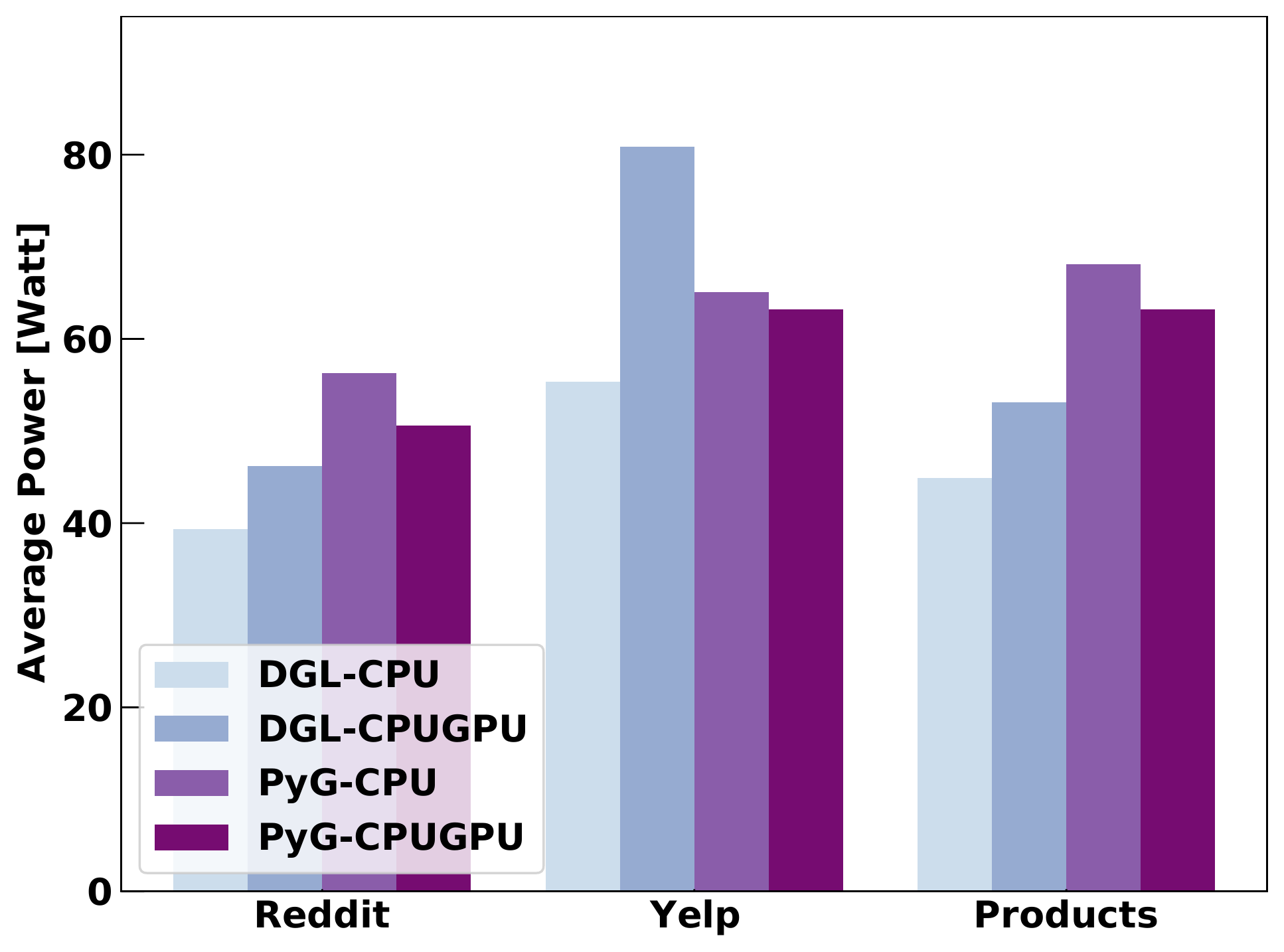}
	}
	\vspace{0mm}
	\caption{Average power consumption of ClusterGCN.}
	\label{fig:cluster-power}
	\vspace{-4mm}
\end{figure}

\begin{figure}[t!]
	\captionsetup[subfloat]{captionskip=1pt}
	\centering
	\subfloat{%
		\includegraphics[width=0.47\linewidth, trim=0cm 0cm 0cm 0cm, clip]{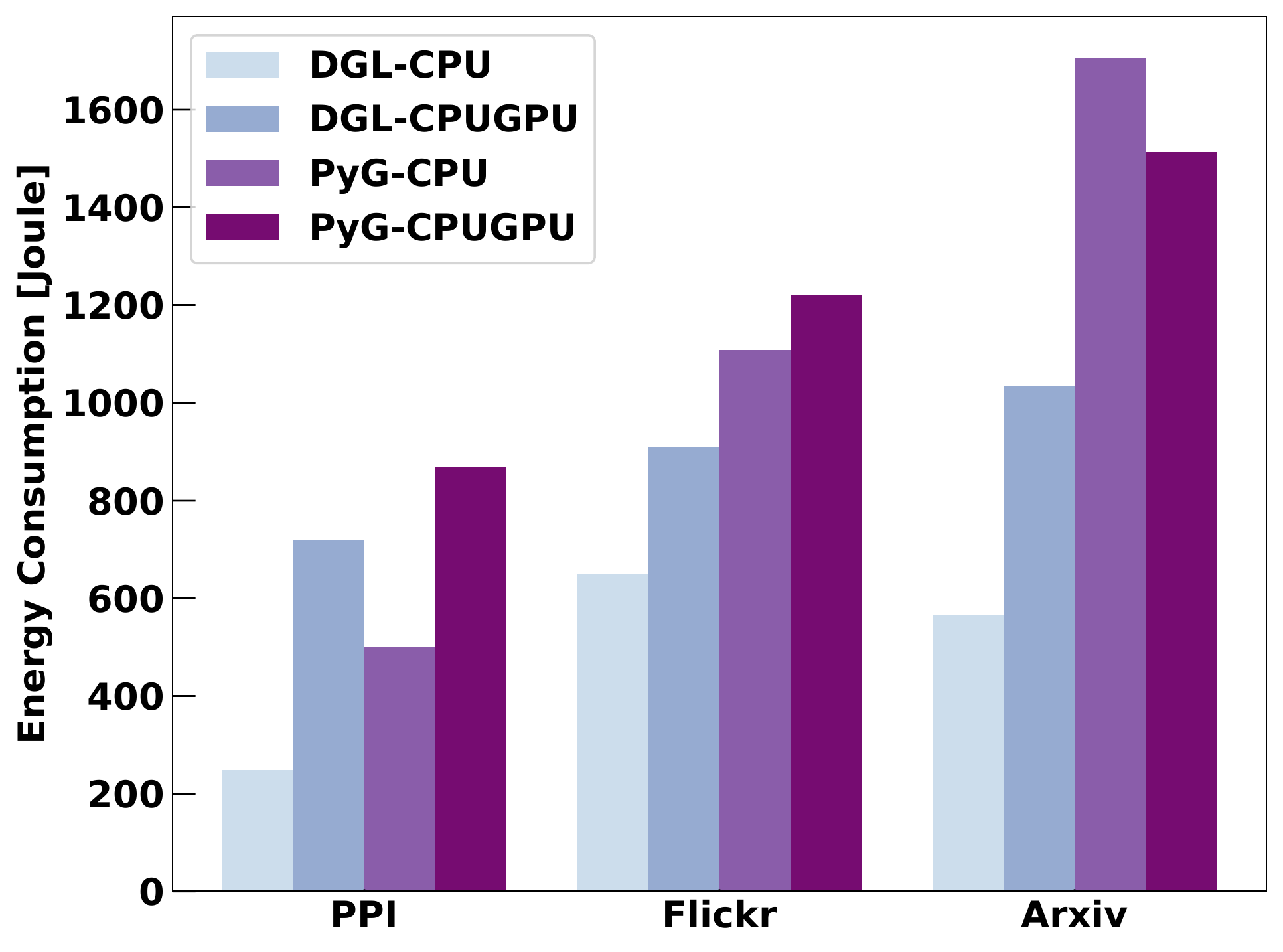}
	}
	\vspace{0mm}
	\subfloat{%
		\includegraphics[width=0.47\linewidth, trim=0cm 0cm 0cm 0cm, clip]{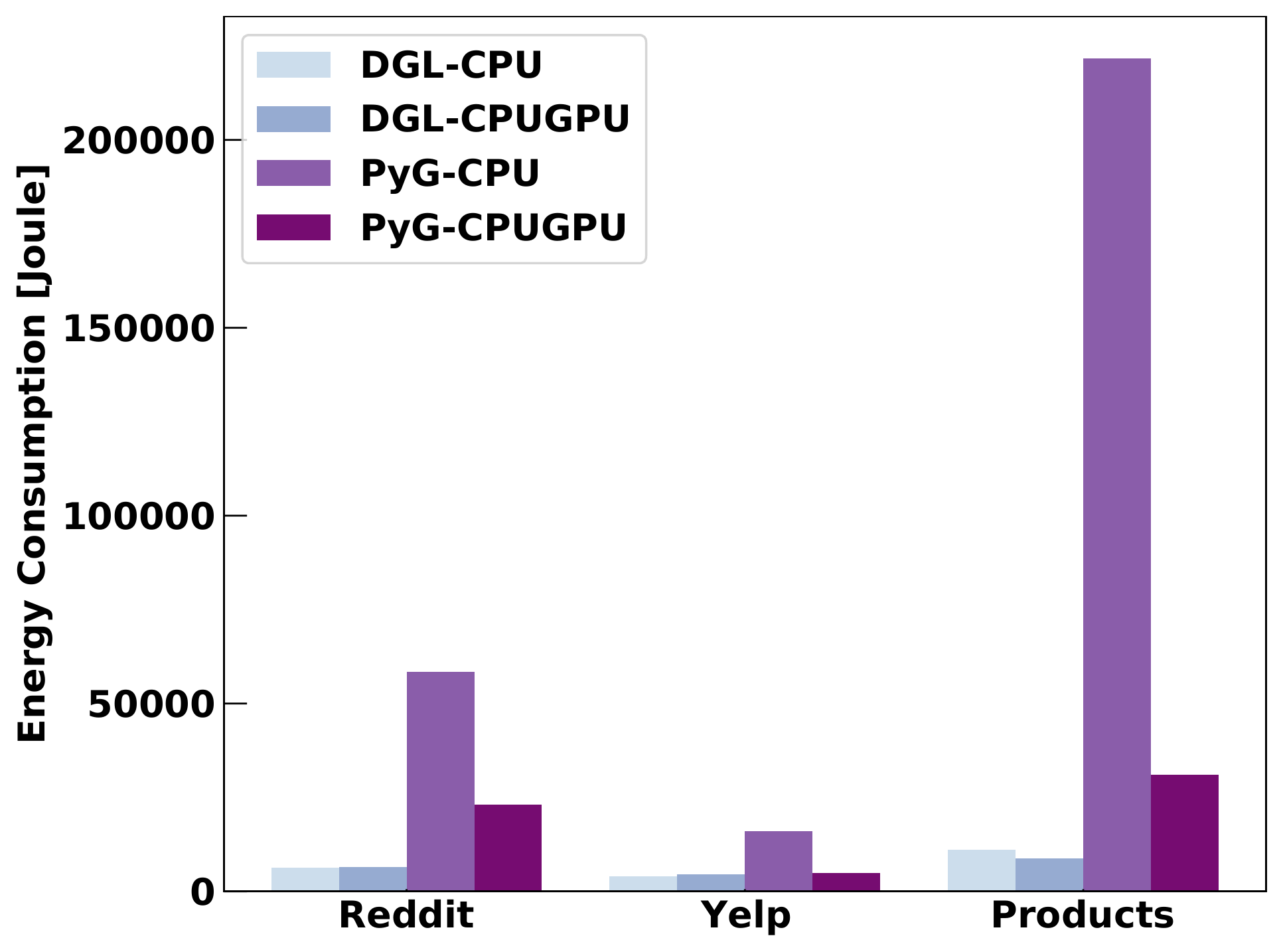}
	}
	\vspace{0mm}
	\caption{Energy consumption of ClusterGCN.}
	\label{fig:cluster-energy}
	\vspace{-4mm}
\end{figure}

\begin{figure}[t!]
	\captionsetup[subfloat]{captionskip=1pt}
	\centering
	\subfloat[DGL-CPU]{%
		\includegraphics[width=0.47\linewidth, trim=0cm 0cm 0cm 0cm, clip]{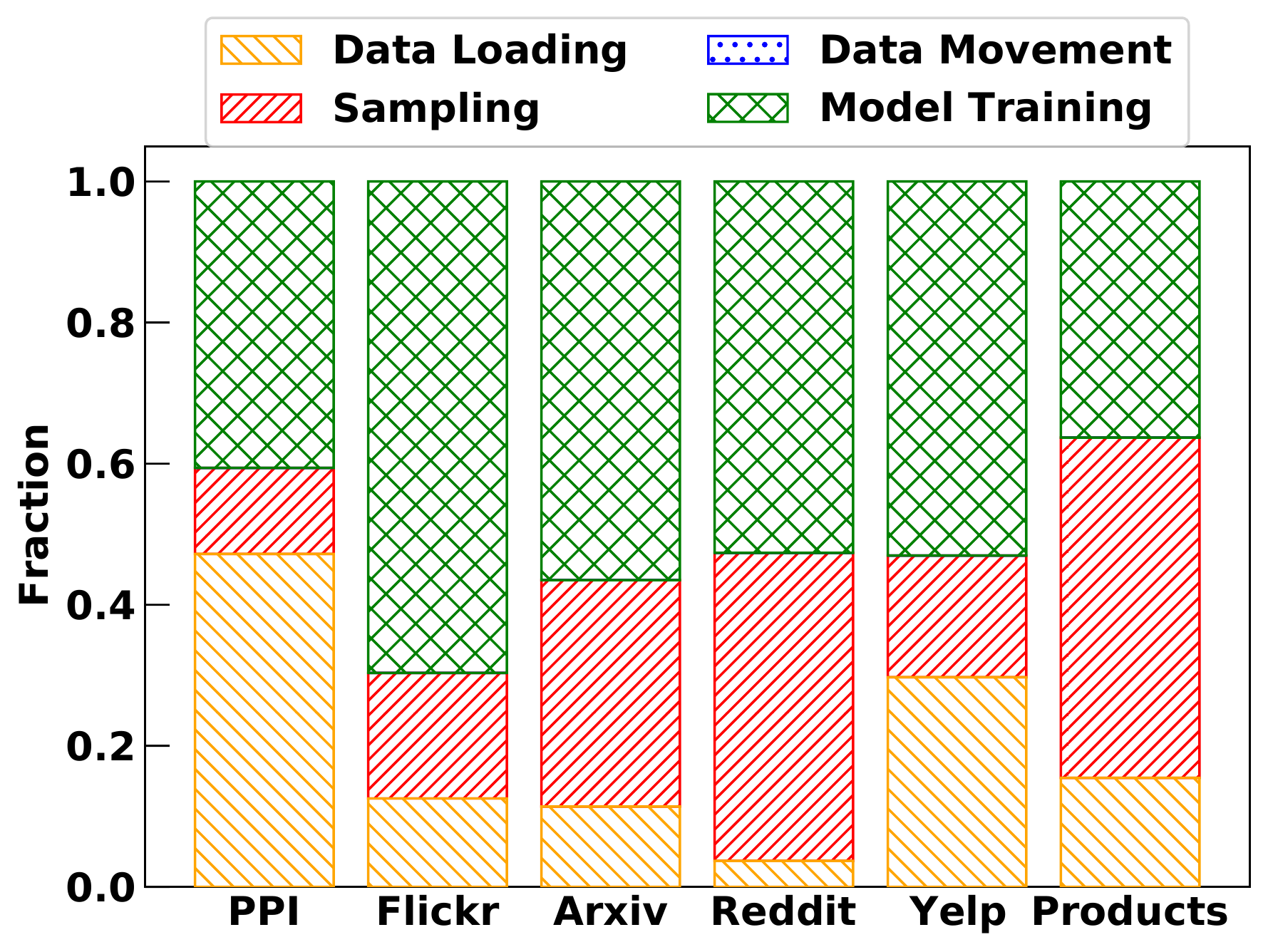}
	}
	\hspace{0mm}
	\subfloat[DGL-CPUGPU]{%
		\includegraphics[width=0.47\linewidth, trim=0cm 0cm 0cm 0cm, clip]{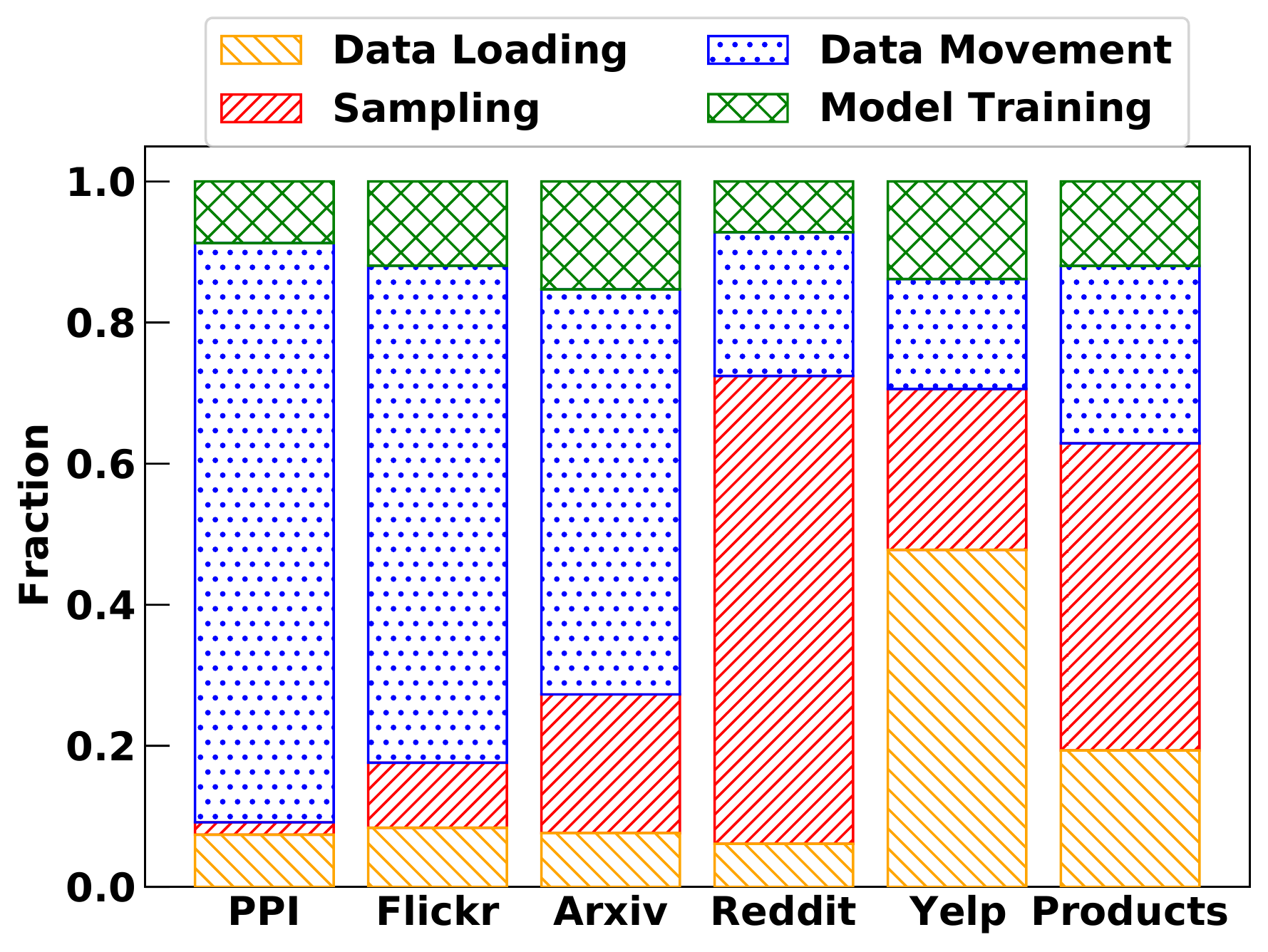}
	}
	\vspace{-2mm}
	\subfloat[PyG-CPU]{%
		\includegraphics[width=0.47\linewidth, trim=0cm 0cm 0cm 0cm, clip]{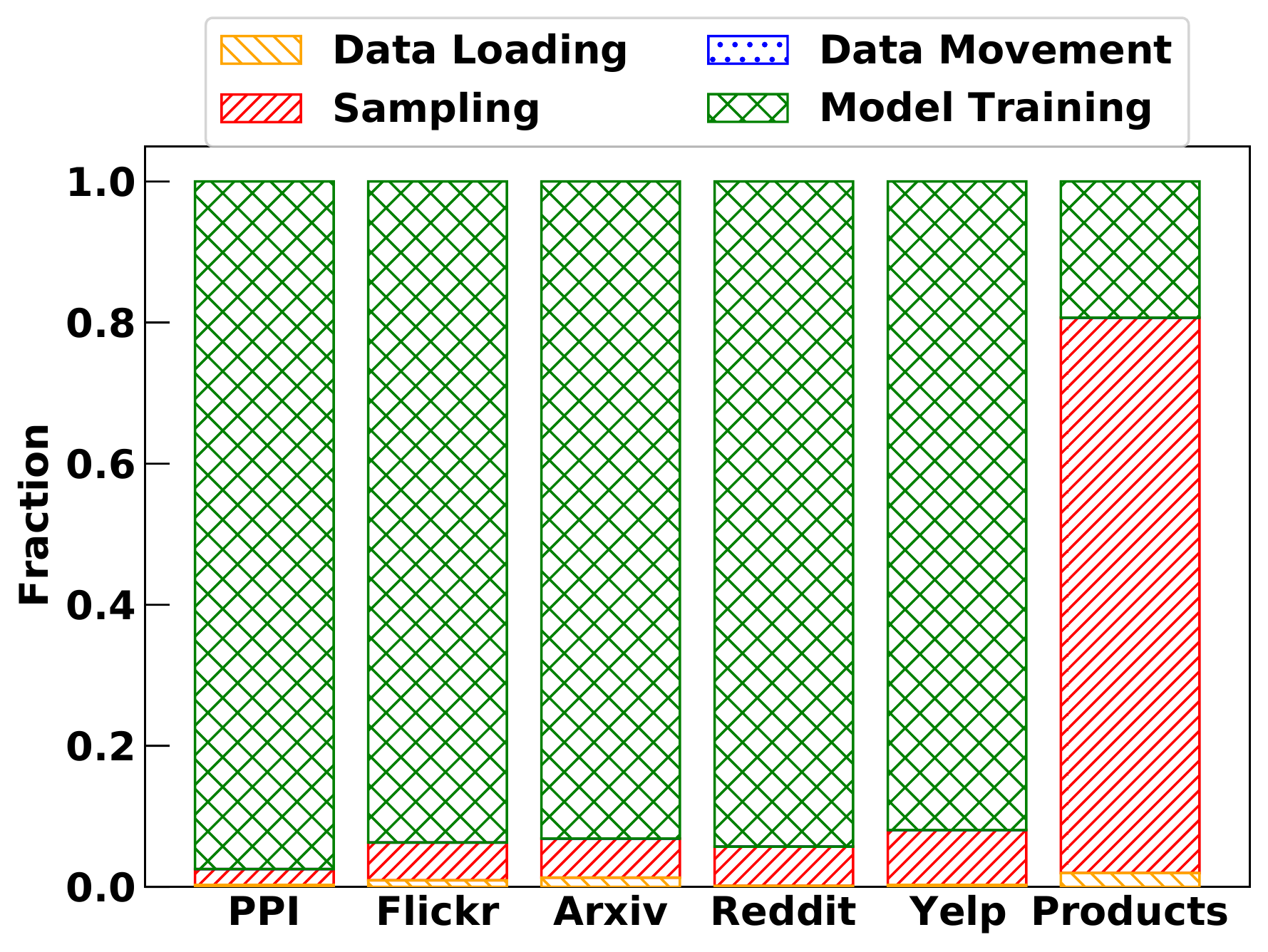}
	}
	\vspace{0mm}
	\subfloat[PyG-CPUGPU]{%
		\includegraphics[width=0.47\linewidth, trim=0cm 0cm 0cm 0cm, clip]{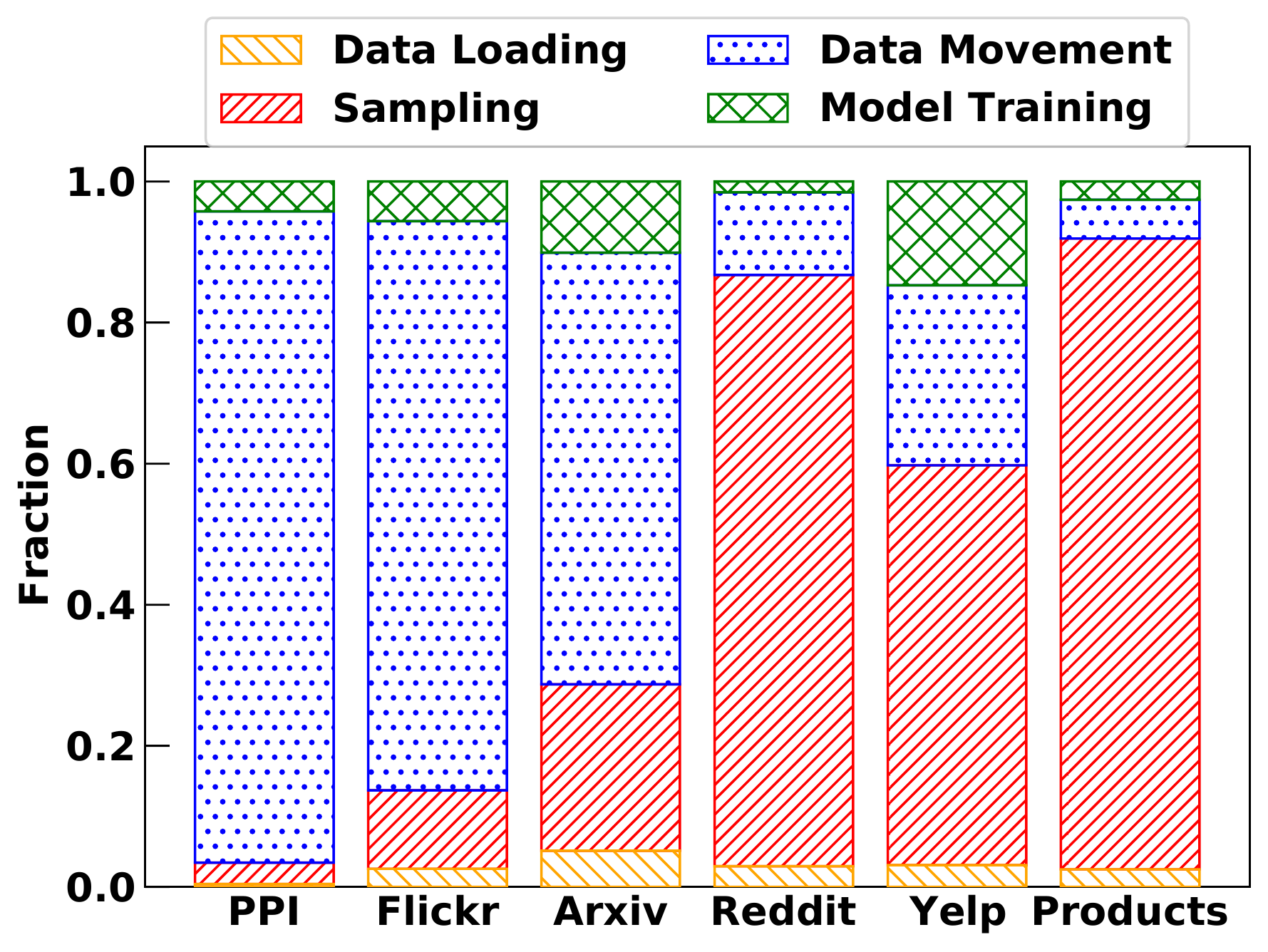}
	}
	\vspace{0mm}
	\caption{Runtime breakdown of GraphSAINT.}
	\label{fig:saint-breakdown}
	\vspace{-3.5mm}
\end{figure}

\begin{figure}[t!]
	\captionsetup[subfloat]{captionskip=1pt}
	\centering
	\subfloat{%
		\includegraphics[width=0.47\linewidth, trim=0cm 0cm 0cm 0cm, clip]{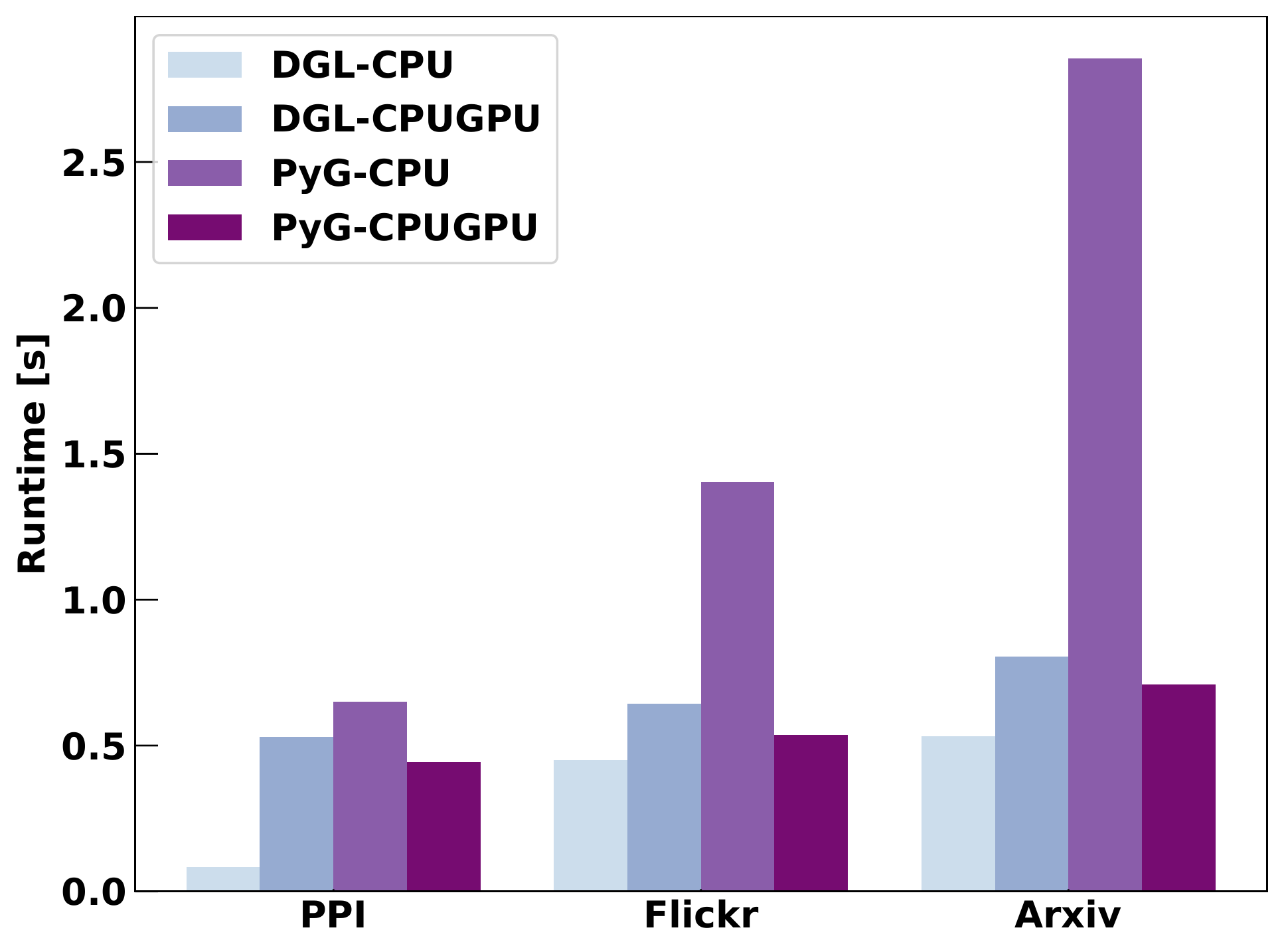}
	}
	\hspace{0mm}
	\subfloat{%
		\includegraphics[width=0.47\linewidth, trim=0cm 0cm 0cm 0cm, clip]{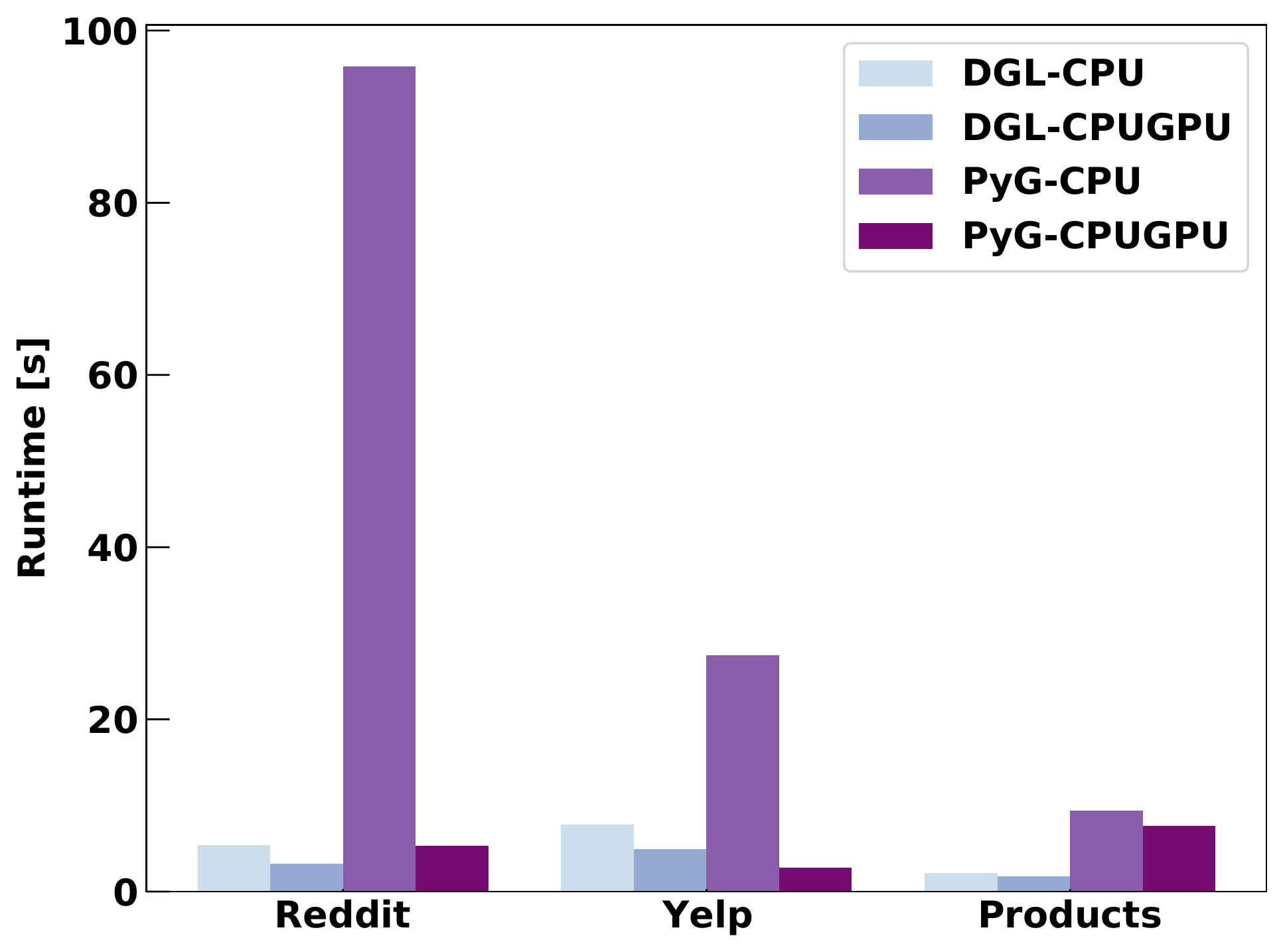}
	}
	\vspace{0mm}
	\caption{Total runtime of GraphSAINT.}
	\label{fig:saint-runtime}
	\vspace{-4mm}
\end{figure}

\begin{figure}[t!]
	\captionsetup[subfloat]{captionskip=1pt}
	\centering
	\subfloat{%
		\includegraphics[width=0.47\linewidth, trim=0cm 0cm 0cm 0cm, clip]{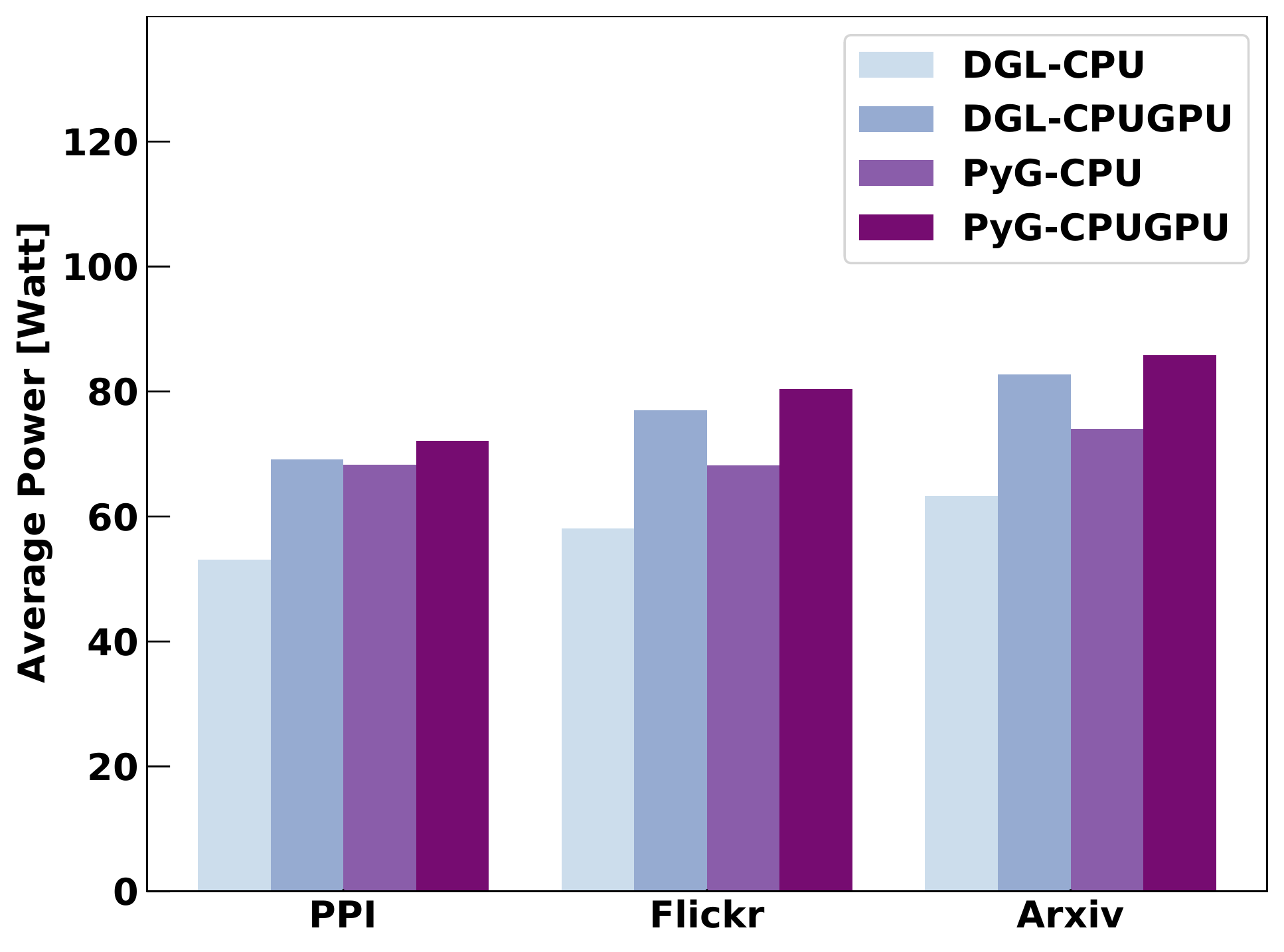}
	}
	\vspace{0mm}
	\subfloat{%
		\includegraphics[width=0.47\linewidth, trim=0cm 0cm 0cm 0cm, clip]{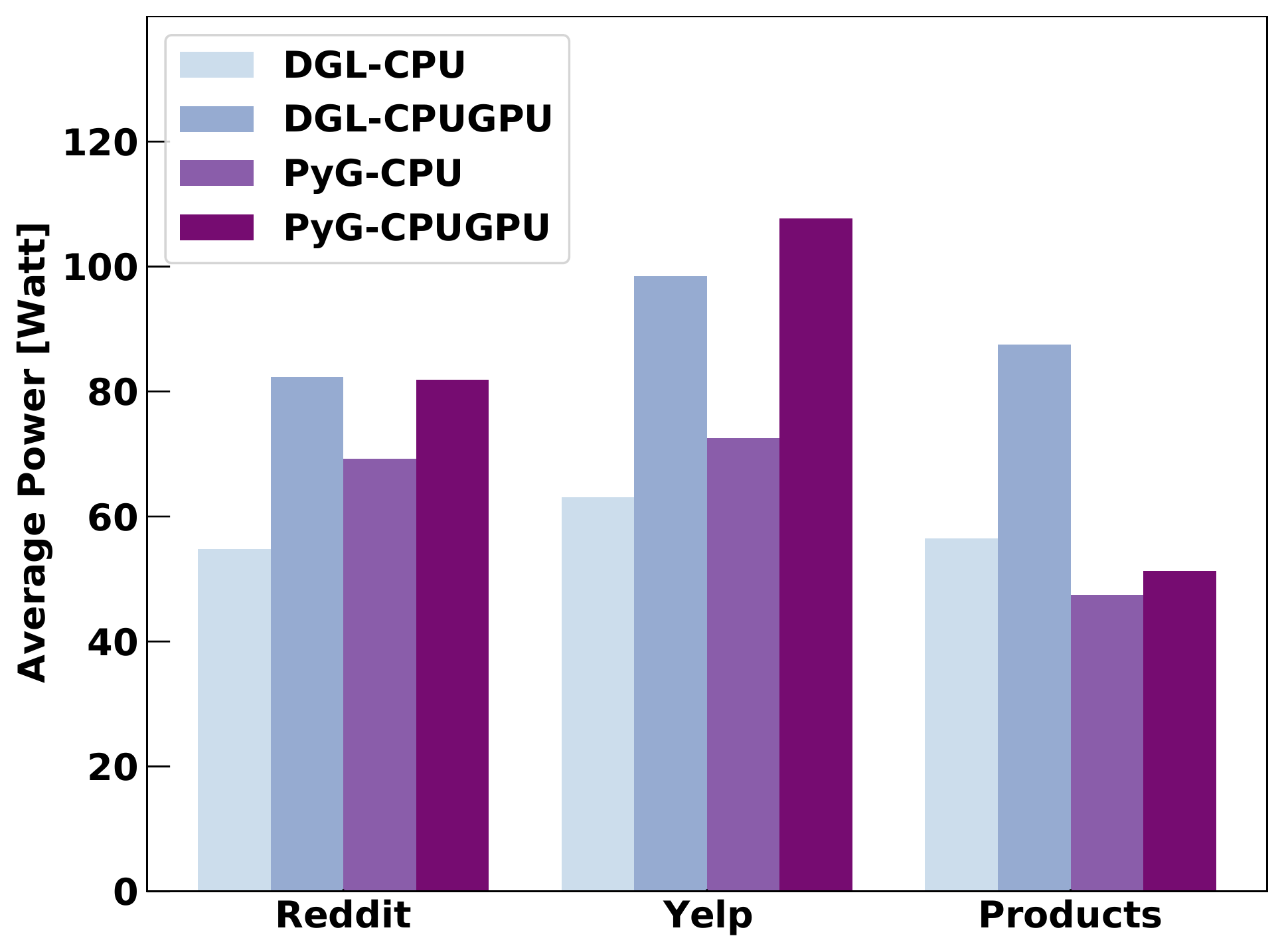}
	}
	\vspace{0mm}
	\caption{Average power consumption of GraphSAINT.}
	\label{fig:saint-power}
	\vspace{-4mm}
\end{figure}

\begin{figure}[t!]
	\captionsetup[subfloat]{captionskip=1pt}
	\centering
	\subfloat{%
		\includegraphics[width=0.47\linewidth, trim=0cm 0cm 0cm 0cm, clip]{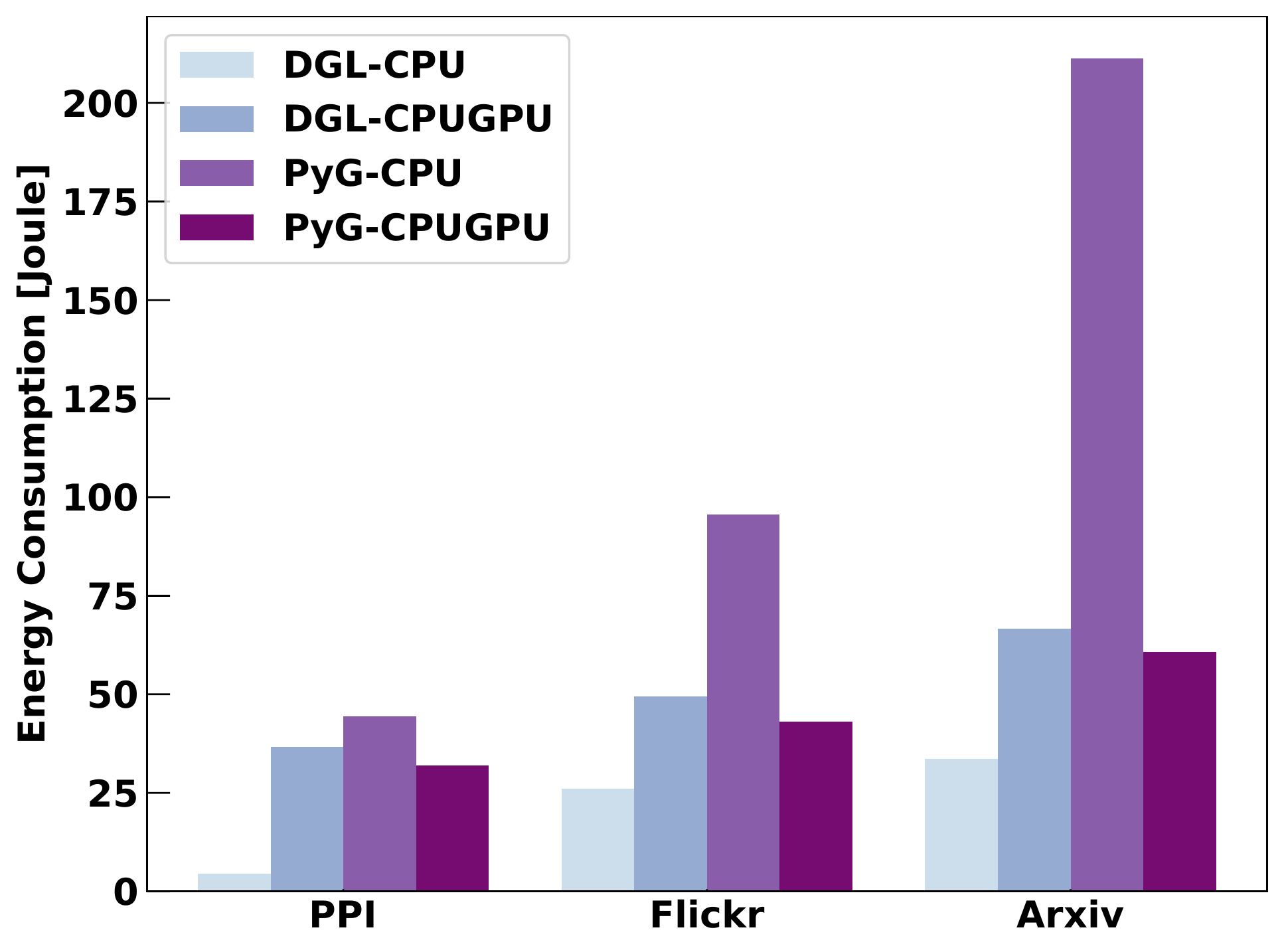}
	}
	\vspace{0mm}
	\subfloat{%
		\includegraphics[width=0.47\linewidth, trim=0cm 0cm 0cm 0cm, clip]{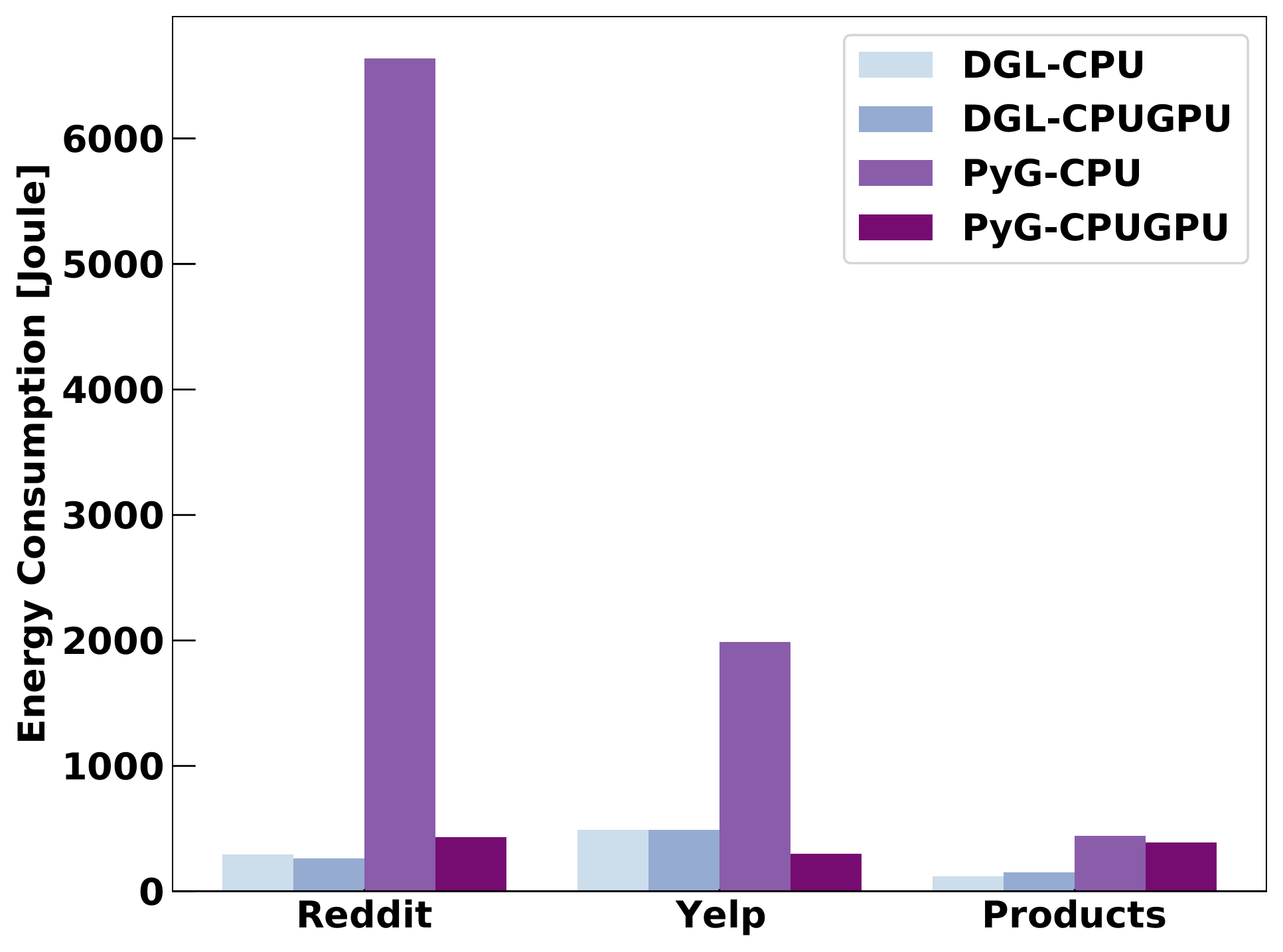}
	}
	\vspace{0mm}
	\caption{Energy consumption of GraphSAINT.}
	\label{fig:saint-energy}
	\vspace{-4mm}
\end{figure}

We evaluate three representative sampling-based GNNs, namely GraphSAGE, ClusterGCN, and GraphSAINT on CPU and GPU separately. We use `DGL-CPU' and `PyG-CPU' to indicate when both sampling and training are done on CPU and use `DGL-CPUGPU' and `PyG-CPUGPU' to indicate when sampling is done on CPU while training is done on GPU. We present their runtime breakdown, total runtime, average power consumption, and energy consumption in Figures~\ref{fig:sage-breakdown}--\ref{fig:saint-energy}. Note that, for all three GNNs, we use the same hyperparameters of their samplers as used in the above functional testing. We use two convolutional layers for all three models and the hyperparameters of each GNN model are set to be the same across DGL and PyG for a fair comparison. The reported results are based on the models trained by 10 epochs. We repeated the same experiments multiple times and observed more or less the same results.  

As shown in Figure~\ref{fig:sage-breakdown}, Figure~\ref{fig:cluster-breakdown}, and Figure~\ref{fig:saint-breakdown}, we break the runtime of each GNN into four parts, which are data loading, sampling, data movement, and model training. Data loading is done by `data loader' to load the input graph and its associated node features from storage to CPU memory. Sampling is done by `sampler' to extract subgraphs and fetch the node features of the sampled subgraphs from the entire feature matrix for mini-batch training. Data movement is to copy the initial weight matrices of a GNN model, each subgraph matrix, and its corresponding node features from CPU to GPU. Note that there is no data movement (from CPU to GPU) for DGL-CPU and PyG-CPU. Model training includes forward propagation, backward propagation, and update of model weights. Note that as the number of training epochs increases, the fraction of data loading in total runtime will decrease since it is a one-time operation. However, sampling, data movement, and model training are performed repeatedly for different mini-batches.

\vspace{1mm}

\noindent \textbf{Observation 4:} \textit{Sampling is slow for all three GNNs and can take up to 90\% of total runtime.}

\vspace{1mm}

This observation indicates that there is a need to optimize sampling and its associated operations. In particular, for PyG, its CPU kernel could be improved for not only sampling but also model training on CPU. In addition, we observe that data movement can also take a large portion of total runtime in both frameworks. As shall be shown in Section~\ref{sec:case-study}, data pre-loading in the frameworks can be used to mitigate this issue.

\vspace{1mm}

\noindent \textbf{Observation 5:} \textit{DGL is generally more efficient than PyG on both CPU and GPU in terms of runtime and energy consumption, especially for large graphs.}

\vspace{1mm}

We observe that PyG is more efficient than DGL for small graphs when CPU is used for sampling and GPU is used for training, while DGL is generally more efficient for the other cases. In particular, PyG-CPUGPU is generally more efficient than DGL-CPUGPU for GraphSAINT. This behavior can be explained as follows. With mini-batch training, a GNN model is trained based on sampled subgraphs, which are much smaller than the input graph. We observe that each sampled subgraph (corresponding to a mini-batch) by GraphSAINT sampler is relatively smaller than the ones by GraphSAGE's neighborhood sampler and ClusterGCN sampler. Here the one with GraphSAGE's neighborhood sampler has multiple subgraphs for a mini-batch. Also, recall that the performance gap of the GraphSAINT sampler between DGL and PyG is insignificant, as shown in Figure~\ref{fig:sampler}. Since PyG is more efficient in model training with small graphs, PyG becomes more efficient even for medium-size graphs with GraphSAINT, as shown in Figure~\ref{fig:saint-runtime}.

In addition, we find that there is no clear winner between DGL and PyG regarding average power consumption, which indicates that energy consumption mainly depends on overall runtime. We observe that GraphSAINT is more efficient in both runtime and energy consumption compared with the GraphSAGE and ClusterGCN, thanks to its light-weight sampling and GNN operations. Note that they are trained for the same number of epochs in our experiments. Nonetheless, we emphasize that different choices of the hyperparameters for each GNN in optimizing its accuracy would affect the efficiency in runtime and energy consumption differently.

\subsection{Case Studies}\label{sec:case-study}

We next turn out attention to three case studies to further evaluate the performance of two GNN frameworks, regarding data pre-loading, GPU-based sampler, and full-batch model training. We focus on GraphSAGE for the case studies.

\begin{figure}[t!]
	\vspace{-2mm}
	\captionsetup[subfloat]{captionskip=1pt}
	\centering
	\subfloat[Speedup of data movement]{%
		\includegraphics[width=0.47\linewidth, trim=0cm 0cm 0cm 0cm, clip]{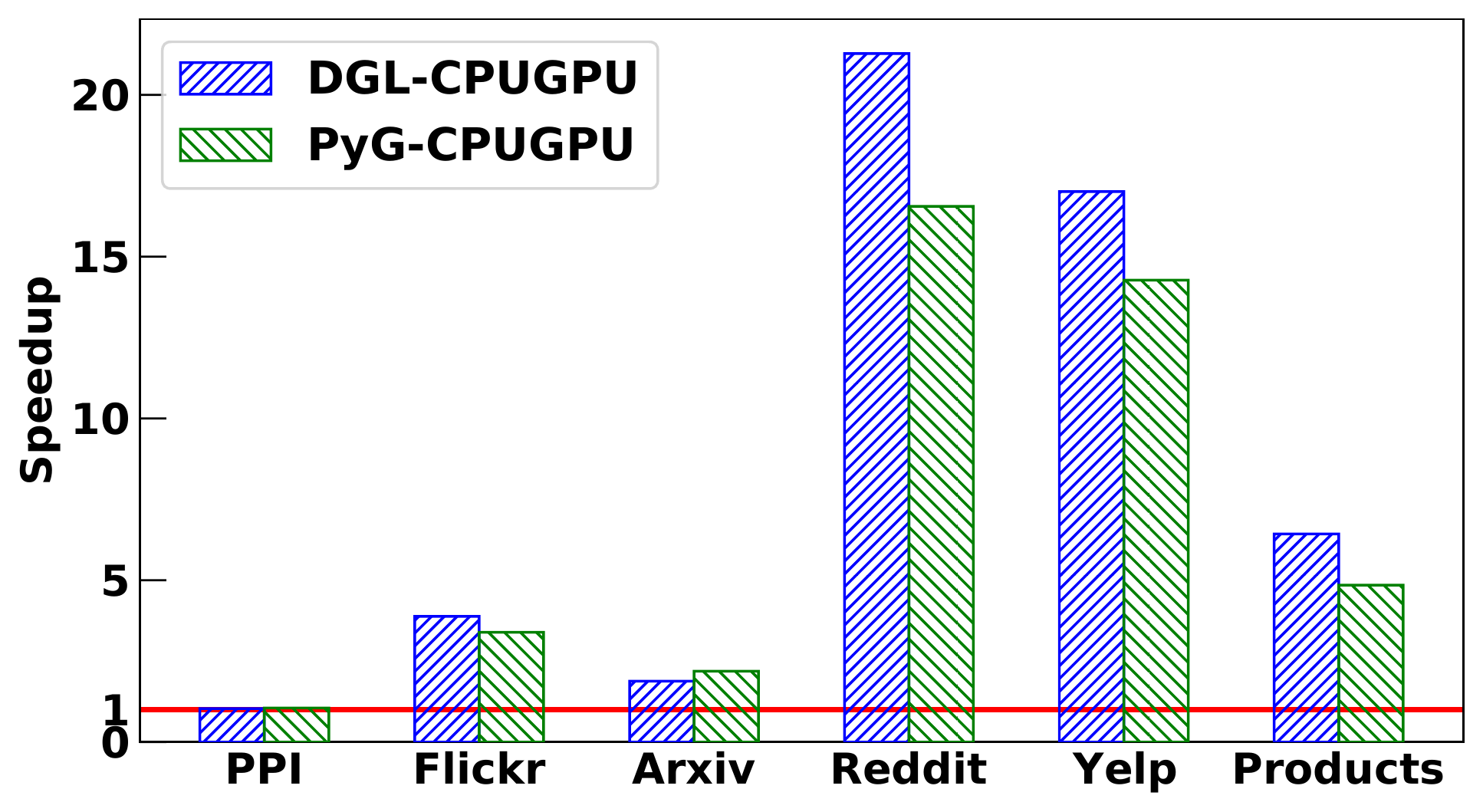}
	}
	\vspace{0mm}
	\subfloat[Speedup of total runtime]{%
		\includegraphics[width=0.47\linewidth, trim=0cm 0cm 0cm 0cm, clip]{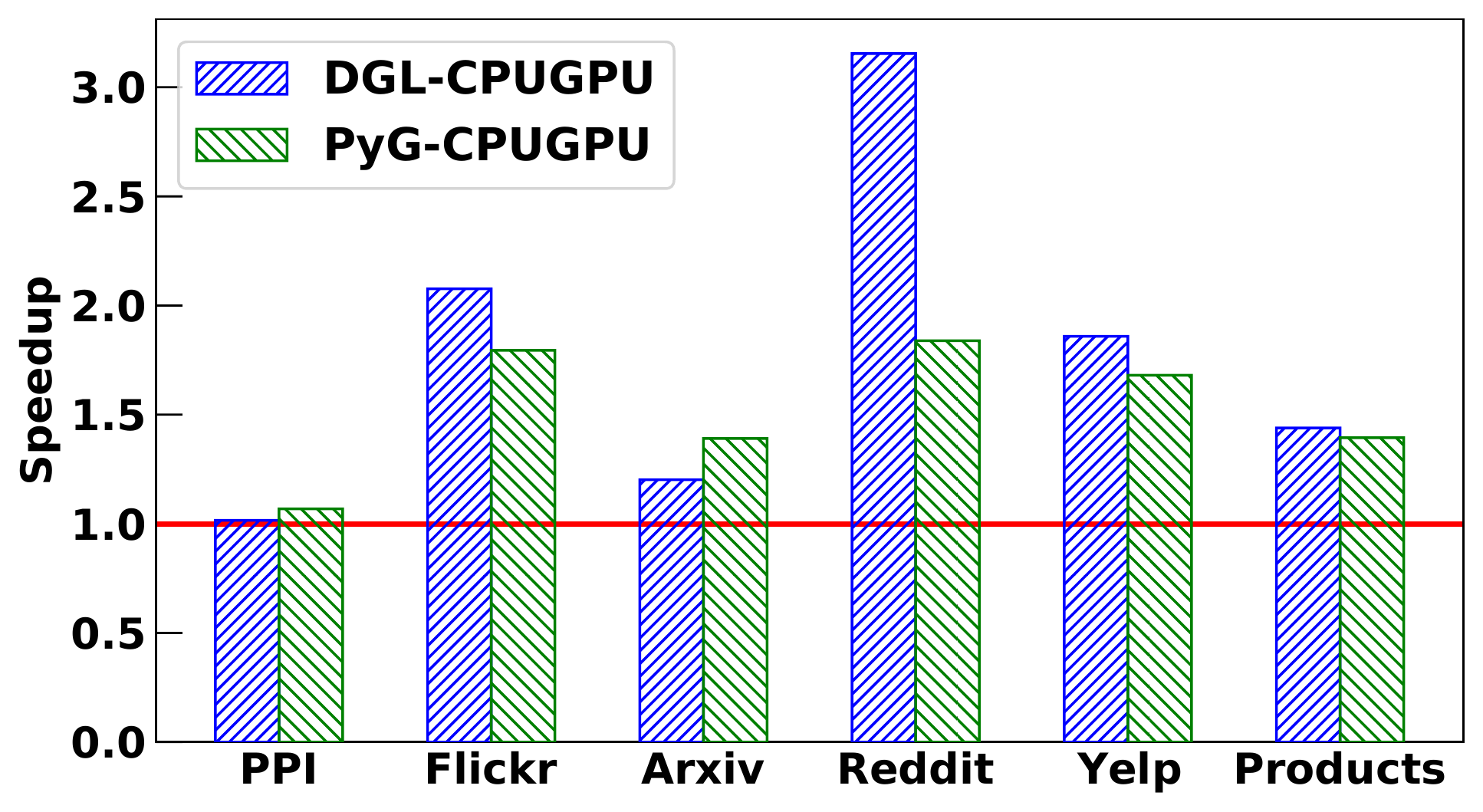}
	}
	\hspace{0mm}
	\caption{Speedup of GraphSAGE when pre-loading the input graph and node features into GPU.}
	\label{fig:speedup-preloading}
	\vspace{-2mm}
\end{figure}

\begin{figure}[t!]
	\captionsetup[subfloat]{captionskip=1pt}
	\centering
	\subfloat[DGL-CPUGPU]{%
		\includegraphics[width=0.47\linewidth, trim=0cm 0cm 0cm 0cm, clip]{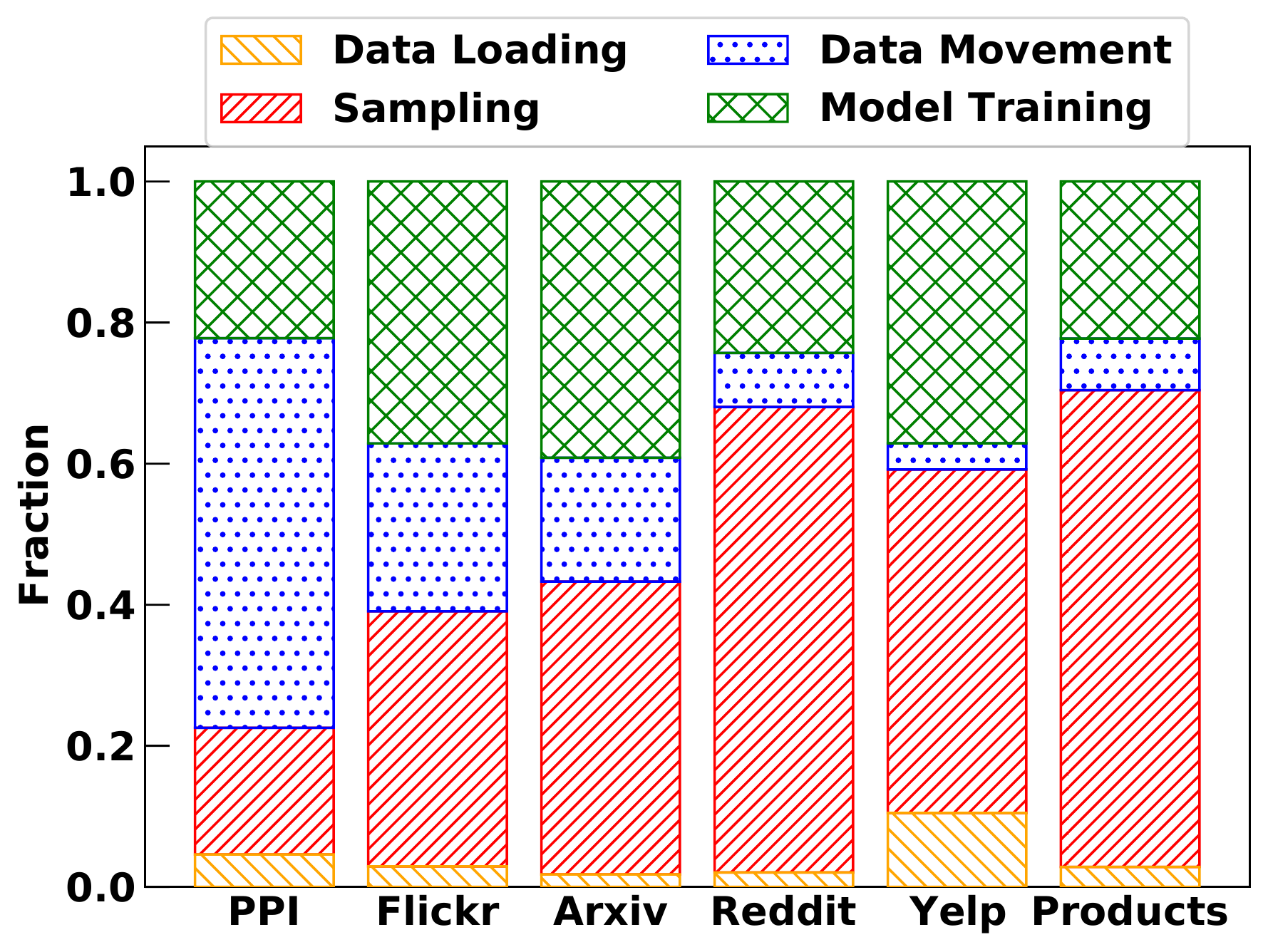}
	}
	\vspace{0mm}
	\subfloat[PyG-CPUGPU]{%
		\includegraphics[width=0.47\linewidth, trim=0cm 0cm 0cm 0cm, clip]{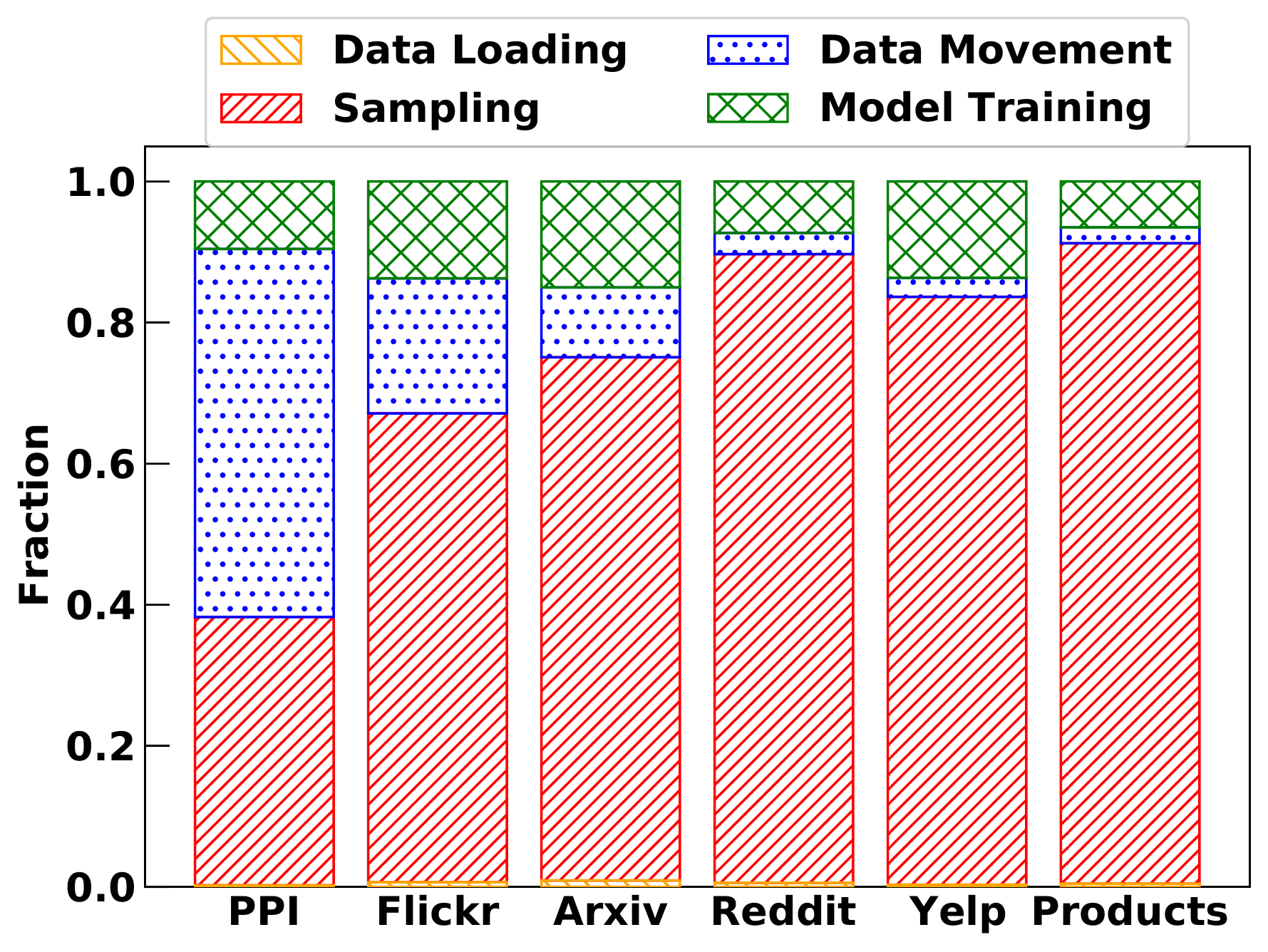}
	}
	\vspace{0mm}
	\caption{Runtime breakdown of GraphSAGE with data pre-loading.}
	\label{fig:breakdown-preloading}
	\vspace{-2mm}
\end{figure}

\vspace{1mm}
\noindent \textbf{Pre-loading entire graph and node features into GPU.} As shown in Section~\ref{sec:performance}, data movement can be a problem when we use CPU for sampling and GPU for model training. We here change the implementation strategy so as to pre-load the entire graph and its associated node features into the GPU \emph{upfront}, which can avoid the overhead of repeated data movement, i.e., the movement of the features of nodes chosen in each mini-batch. Both frameworks provide such a pre-loading option. With this option, the adjacency matrices of sampled subgraphs only need to be copied from CPU to GPU for each mini-batch periodically. Note that a mini-batch is composed of a number of sampled subgraphs in GraphSAGE, where the number of sampled subgraphs is the mini-batch size. We present the resulting performance of DGL and PyG with GraphSAGE in Figure~\ref{fig:breakdown-preloading} for runtime breakdown and in Figure~\ref{fig:speedup-preloading} for speedup results.

\vspace{1mm}

\noindent \textbf{Observation 6:} \textit{The data pre-loading can significantly reduce overall data movement time in both frameworks.}

\vspace{1mm}

As expected, the pre-loading strategy saves up to 20x data movement time, thereby leading to about 2x overall speedup. Nonetheless, \emph{it is only feasible when the GPU memory is large enough to hold the entire graph and its associated node features as well as the weight matrix of a GNN model.} It is often not the case in practice, especially for large graphs. An alternative yet effective strategy would be to cache the features of nodes that are most frequently used for model training, i.e., partial information of the graph, into GPU memory upfront, to reduce overall data movement time~\cite{dong2021global}.

It is worth noting that DGL further provides an advanced feature called `pre-fetching' for asynchronous data movement and model computation. We observed that the performance of DGL can be further improved, albeit a little bit, with this feature. We omit the results here for brevity.

\begin{figure*}[t!]
	\captionsetup[subfloat]{captionskip=1pt}
	\centering
	\subfloat[Speedup over DGL-CPUGPU]{%
		\includegraphics[width=0.3\linewidth, trim=0cm 0cm 0cm 0cm, clip]{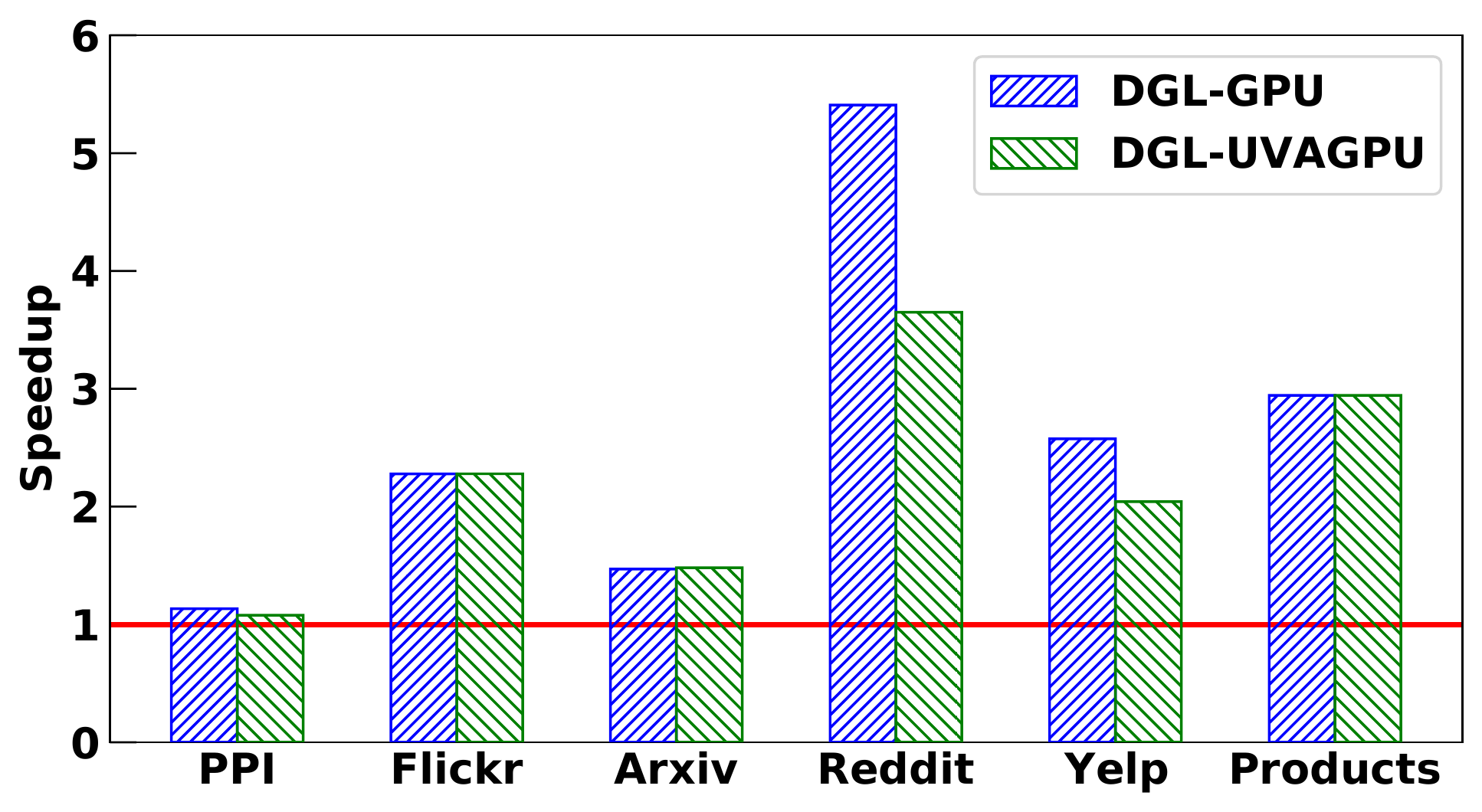}
	}
	\hspace{0mm}
	\subfloat[Powerup over DGL-CPUGPU]{%
		\includegraphics[width=0.3\linewidth, trim=0cm 0cm 0cm 0cm, clip]{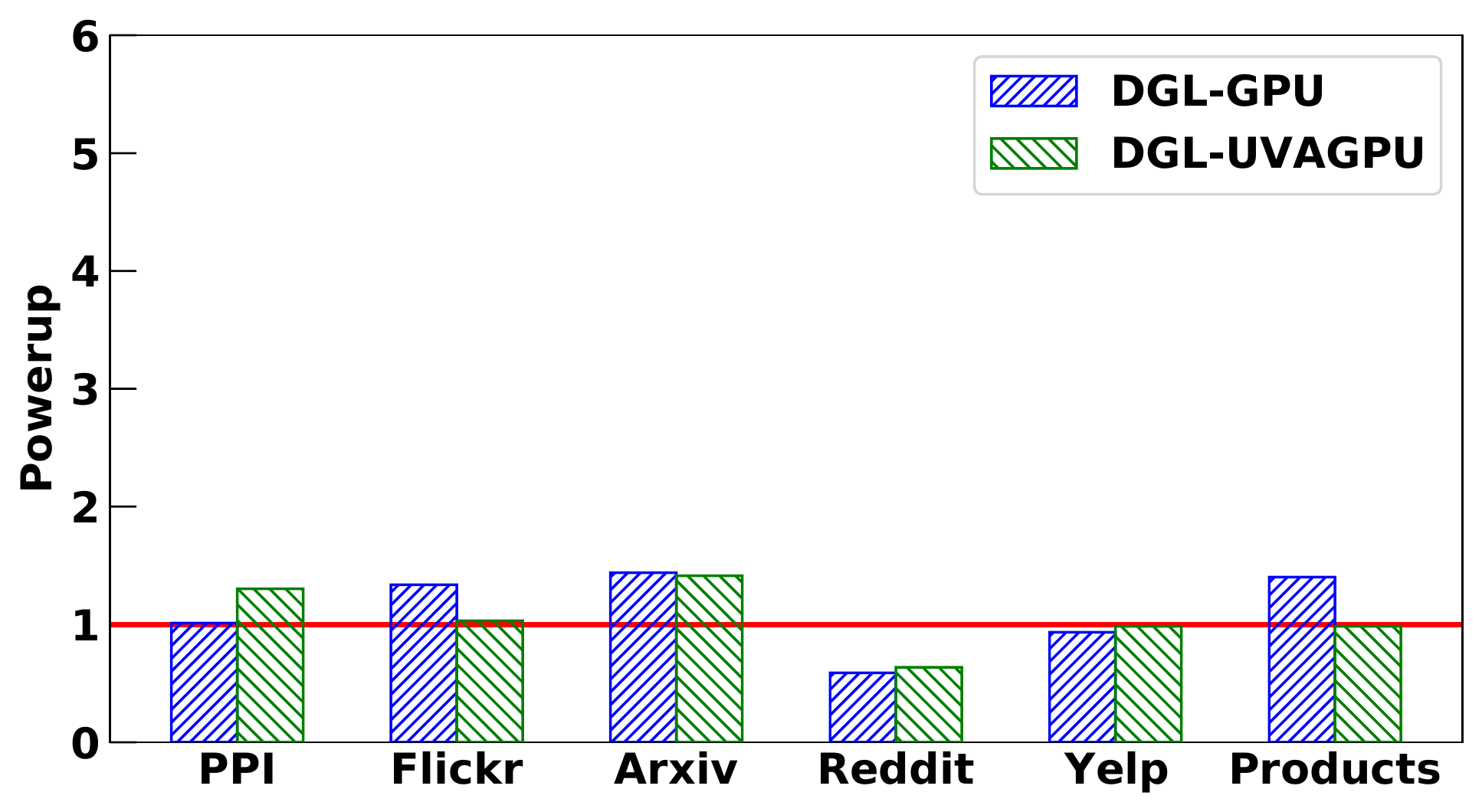}
	}
	\hspace{0mm}
	\subfloat[Greenup over DGL-CPUGPU]{%
		\includegraphics[width=0.3\linewidth, trim=0cm 0cm 0cm 0cm, clip]{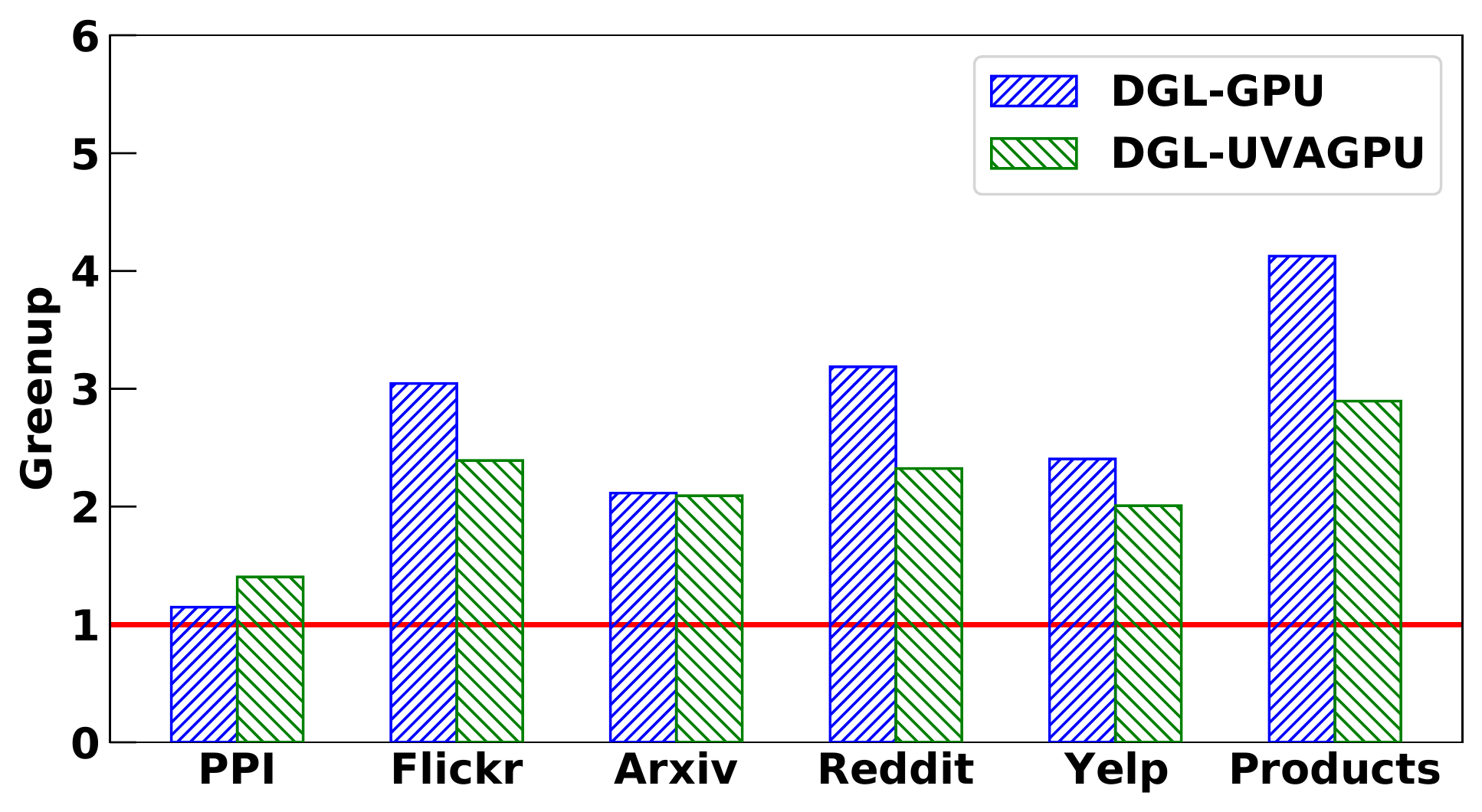}
	}
	\vspace{0mm}
	\caption{GPS-UP metrics of GraphSAGE with DGL's GPU-based sampler and UVA-based sampler.}
	\label{fig:uvagpu-speedup}
\end{figure*}

\begin{figure}[t!]
	\vspace{-2mm}
	\captionsetup[subfloat]{captionskip=1pt}
	\centering
	\subfloat[DGL-GPU]{%
		\includegraphics[width=0.47\linewidth, trim=0cm 0cm 0cm 0cm, clip]{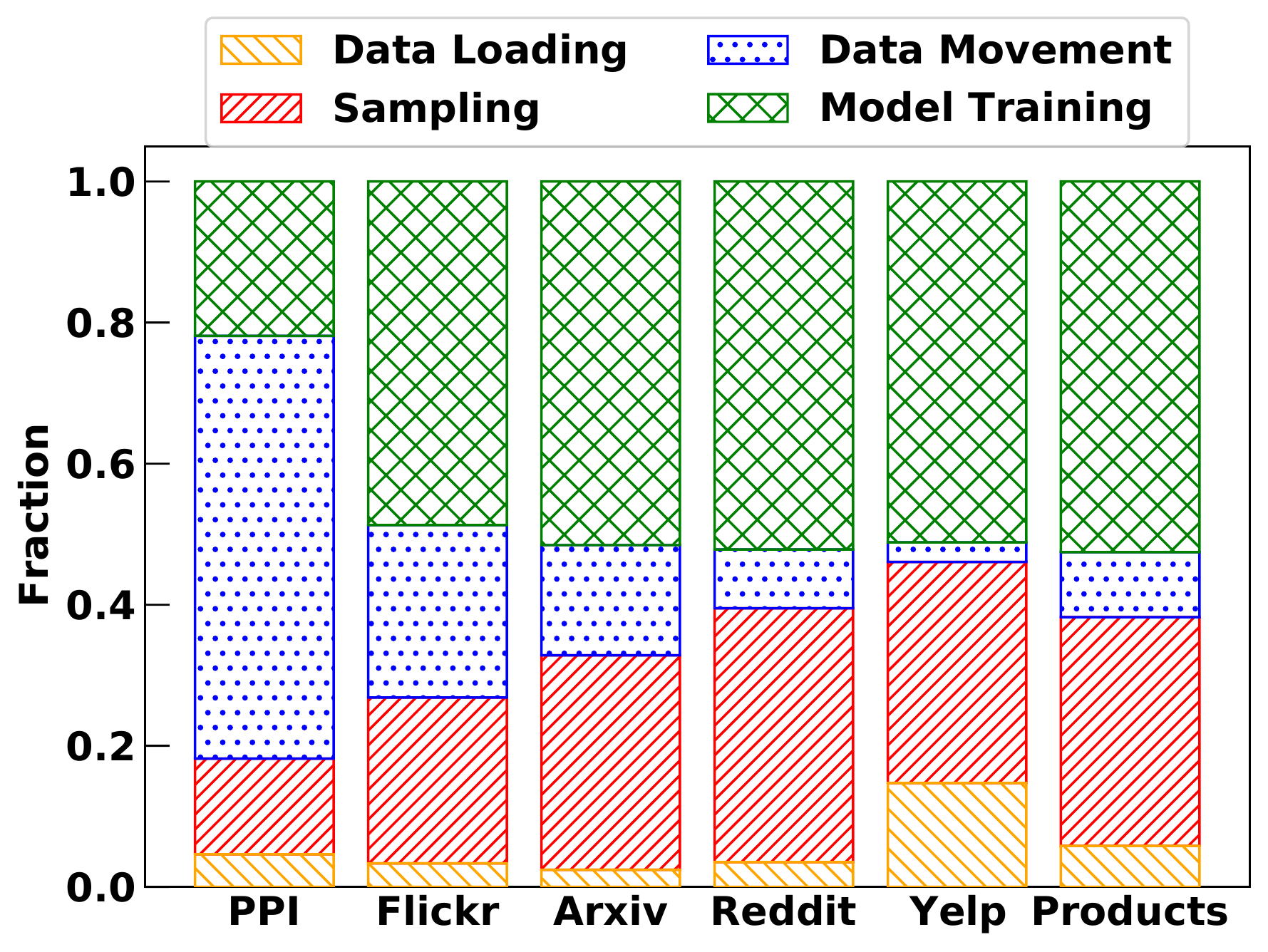}
	}
	\vspace{0mm}
	\subfloat[DGL-UVAGPU]{%
		\includegraphics[width=0.47\linewidth, trim=0cm 0cm 0cm 0cm, clip]{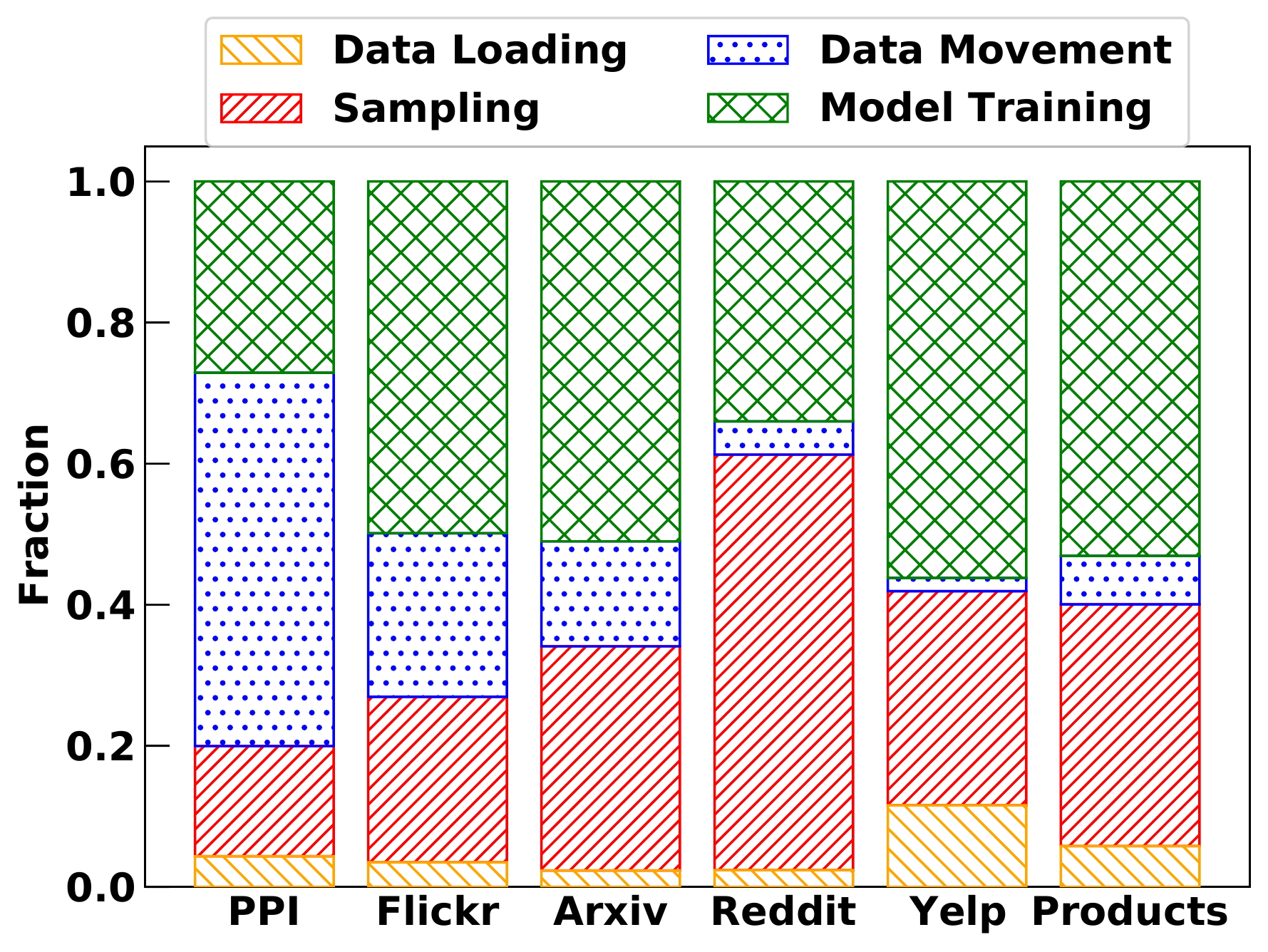}
	}
	\vspace{0mm}
	\caption{Runtime breakdown of GraphSAGE with DGL's GPU-based sampler and UVA-based sampler.}
	\label{fig:uvagpu}
\end{figure}

\vspace{1mm}
\noindent \textbf{GPU-based sampler.} As mentioned before, the GPU-based neighborhood sampler, i.e., the sampler in GraphSAGE, is available in DGL to accelerate its sampling operation and eliminate the need of moving sampled subgraphs from CPU to GPU for each mini-batch. If the GPU-based sampler is used together with the pre-loading option, it can also eliminate the repeated data transfer of node features, corresponding to sampled subgraphs for each mini-batch. This combination is, however, infeasible for the cases with large graphs and/or high-dimensional feature data that do not fit into GPU memory.

In addition, DGL supports another sampler for GraphSAGE, which is the CUDA-Unified Virtual Addressing (UVA)-based sampler. It uses GPU to perform the sampling operation on the input graph and node features pinned on CPU memory via zero-copy access. This UVA support allows DGL to deal with much larger graphs with the benefits of using GPU for sampling and model training. Note that both UVA-based sampler and GPU-based sampler are currently only available for GraphSAGE in DGL.

We evaluate the performance of the GPU-based sampler (`DGL-GPU') and UVA-based sampler (`DGL-UVAGPU') to see how much improvement they can achieve. For the former, we also use the data pre-loading option. Their runtime-breakdown results are reported in Figure~\ref{fig:uvagpu}. Here the data movement for DGL-GPU contains two parts, which are (1) copying the input graph and node features to GPU for sampling and (2) moving the initial GNN model from CPU to GPU for training. For DGL-UVAGPU, the data movement is only for the initial model.

\vspace{1mm}

\noindent \textbf{Observation 7:} \textit{The portion of the sampling operation in total runtime becomes smaller compared with the one with DGL-CPUGPU. However, even with GPU for sampling, it can still take up to 40\% of total runtime for DGL-GPU and 60\% for DGL-UVAGPU. This indicates the non-trivial overhead of the sampling operation and the potential benefit of further accelerating the sampler.}

\vspace{1mm}

We next use GPS-UP (Speedup, Greenup, and Powerup) metrics introduced in~\cite{abdulsalam2015using} for further efficiency analysis. The metrics are designed for comparing two different implementations. One of them is an non-optimized version (i.e. baseline) and the other is an optimized version for better performance. Specifically, they are defined as
\begin{equation*}
\setlength{\abovedisplayskip}{5pt}
\setlength{\belowdisplayskip}{5pt}
\text{Speedup} = \frac{T_{\phi}}{T_o}, \quad \text{Greenup} = \frac{E_{\phi}}{E_o},
\end{equation*}
\begin{equation*}
\setlength{\abovedisplayskip}{5pt}
\setlength{\belowdisplayskip}{5pt}
\text{Powerup} = \frac{P_o}{P_{\phi}} = \frac{E_o/T_o}{E_{\phi}/T_{\phi}} = \frac{\text{Speedup}}{\text{Greenup}},
\end{equation*}
where $T_{\phi}$, $E_{\phi}$, and $P_{\phi}$ are the runtime, energy consumption, and average power of the non-optimized version, respectively, and $T_o$, $E_o$, and $P_o$ are the corresponding values of the optimized one, respectively. We here use DGL-CPUGPU as baseline and report Speedup, Greenup, and Powerup results achieved by DGL-GPU and DGL-UVAGPU over DGL-CPUGPU in Figure~\ref{fig:uvagpu-speedup}.

\vspace{1mm}

\noindent \textbf{Observation 8:} \textit{The use of GPU for sampling saves both time and power in most cases, leading to significant energy saving.}

\vspace{1mm}

As can be seen from Figure~\ref{fig:uvagpu-speedup}(a), DGL-GPU achieves up to 5.5x speedup over DGL-CPUGPU. DGL-UVAGPU is sightly slower than DGL-GPU, because the former uses zero-copy access to CPU memory, which is generally slower than having access to GPU onboard memory. From Figure~\ref{fig:uvagpu-speedup}(b), we also observe that Powerup is not always above one, which implies that the power consumption of using GPU for the sampler can be higher than the CPU counterpart. It happens, especially when there are a large number of edges for each node, e.g., the case of Reddit, making the sampling computation on GPU heavier. Nonetheless, as shown in Figure~\ref{fig:uvagpu-speedup}(c), we observe that Greenup is always above one. In other words, it is \emph{more energy-efficient} using GPU for the sampler. While GPU can consume more power than CPU for the sampling operation, it significantly reduces the total runtime, which translates into smaller overall energy consumption.

Our observations indicate the benefits of using GPU for the sampler of GNNs. Nonetheless, this GPU support is currently only limited to GraphSAGE in DGL, and there is no such support in PyG. Note that there is a recent study~\cite{jangda2021accelerating} that leverages GPUs to accelerate graph sampling for GNNs.

\begin{figure}[t]
	\vspace{-2mm}
	\captionsetup[subfloat]{captionskip=1pt}
	\centering
	\subfloat{%
		\includegraphics[width=0.47\linewidth, trim=0cm 0cm 0cm 0cm, clip]{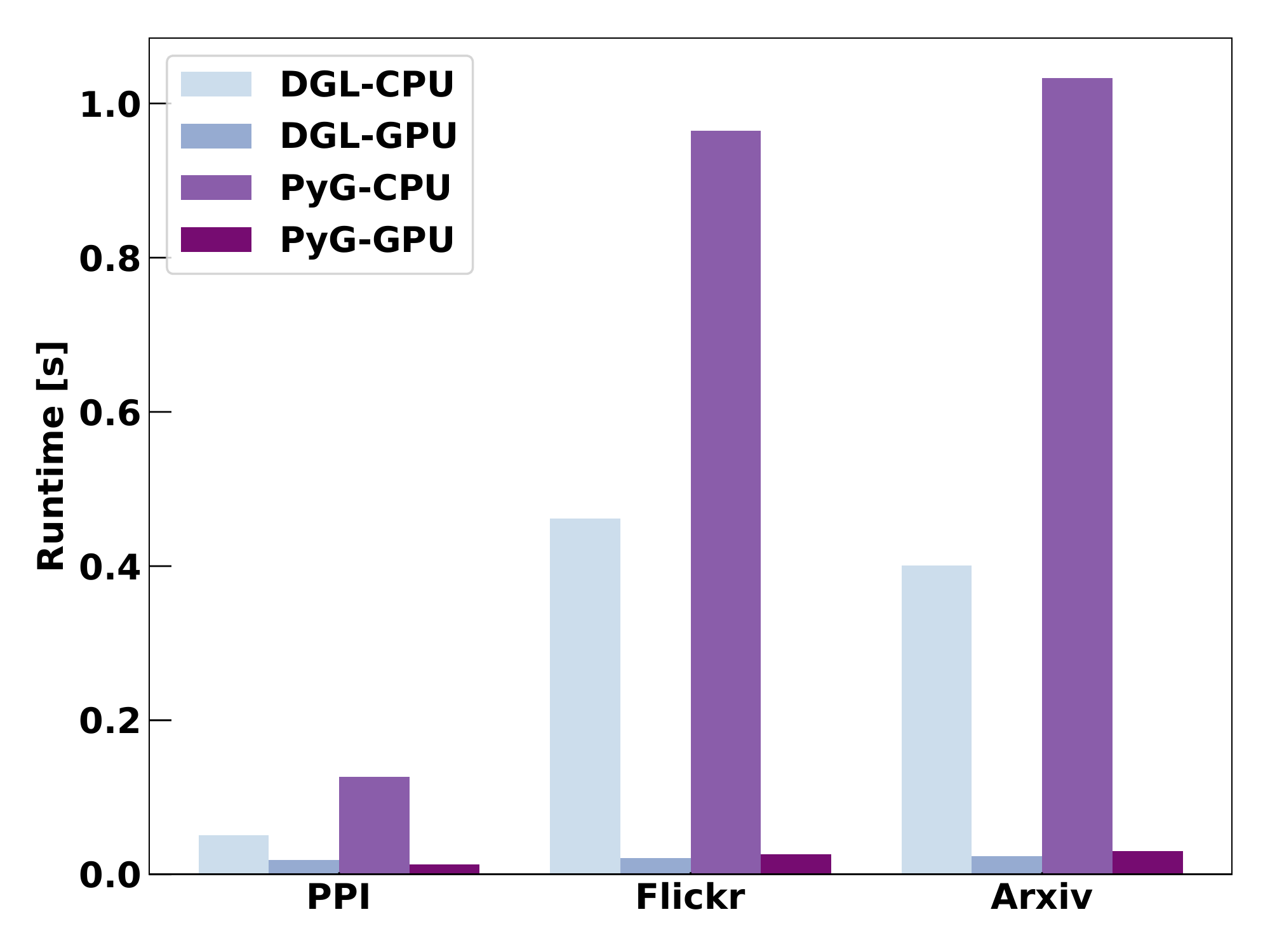}
	}
	\hspace{0mm}
	\subfloat{%
		\includegraphics[width=0.47\linewidth, trim=0cm 0cm 0cm 0cm, clip]{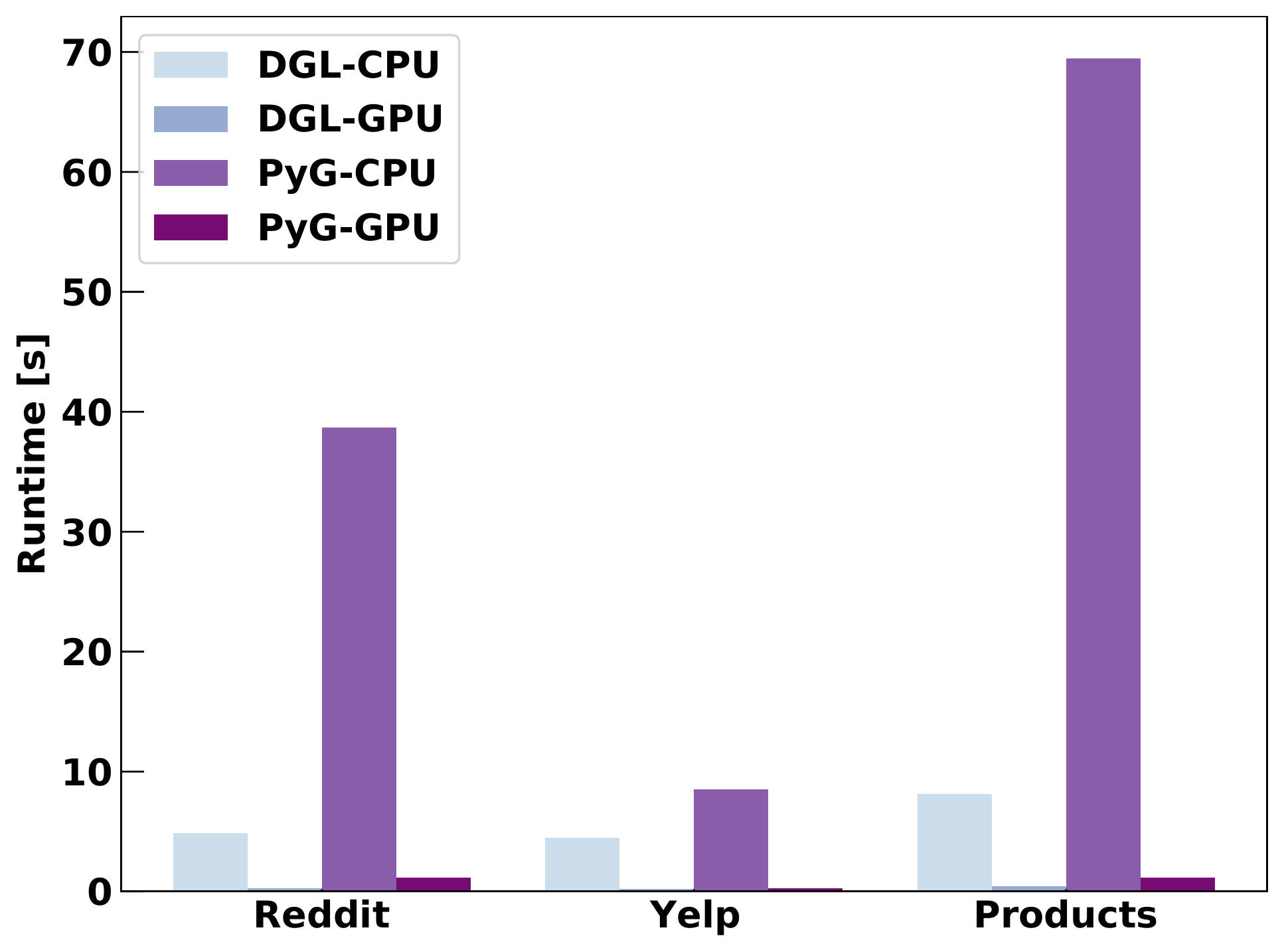}
	}
	\vspace{-1mm}
	\caption{One epoch training time of full-batch GraphSAGE.}
	\label{fig:fullbatch}
	\vspace{-3mm}
\end{figure}

\begin{figure}[t!]
	\captionsetup[subfloat]{captionskip=1pt}
	\centering
	\subfloat{%
		\includegraphics[width=0.47\linewidth, trim=0cm 0cm 0cm 0cm, clip]{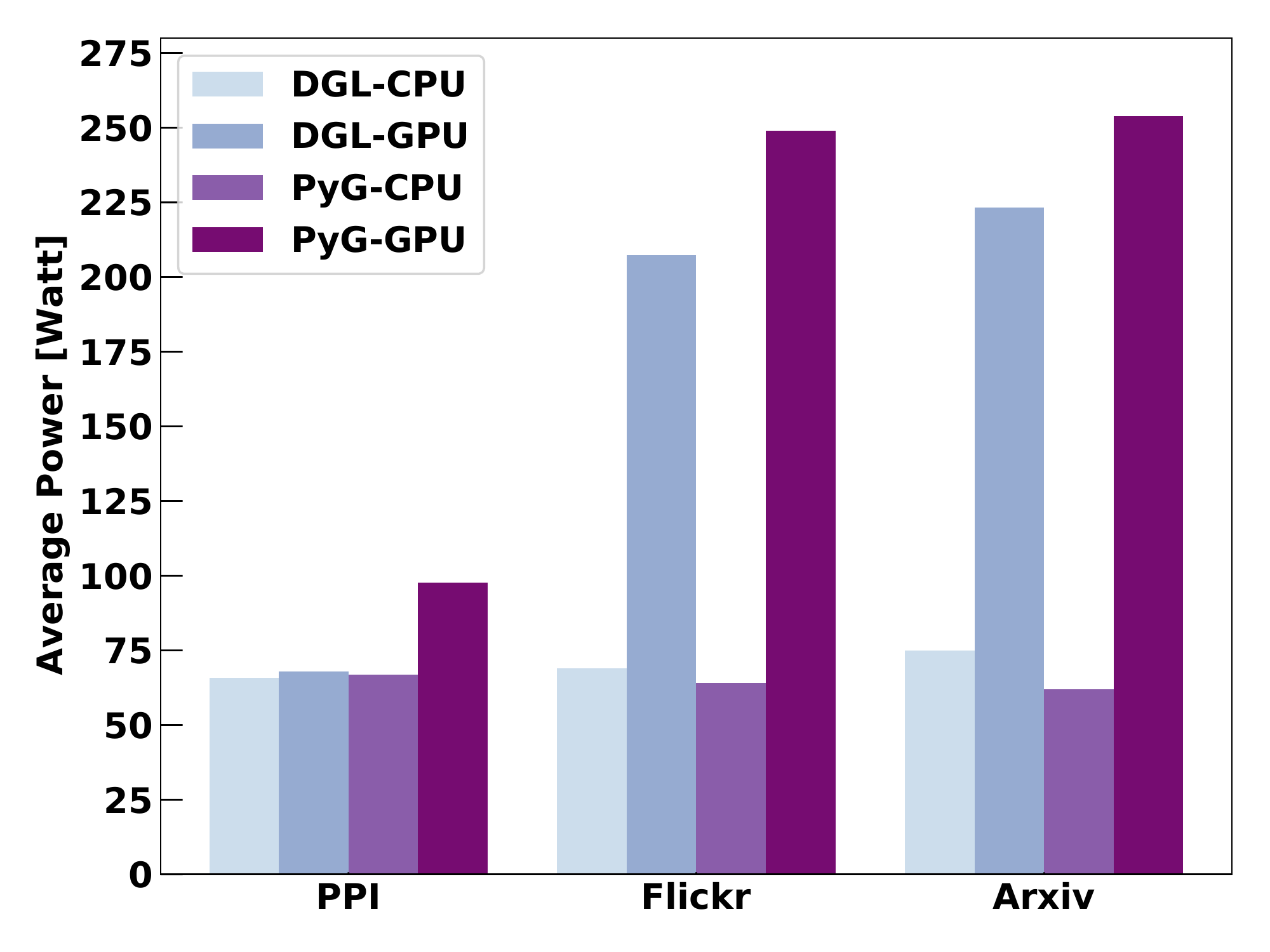}
	}
	\hspace{0mm}
	\subfloat{%
		\includegraphics[width=0.47\linewidth, trim=0cm 0cm 0cm 0cm, clip]{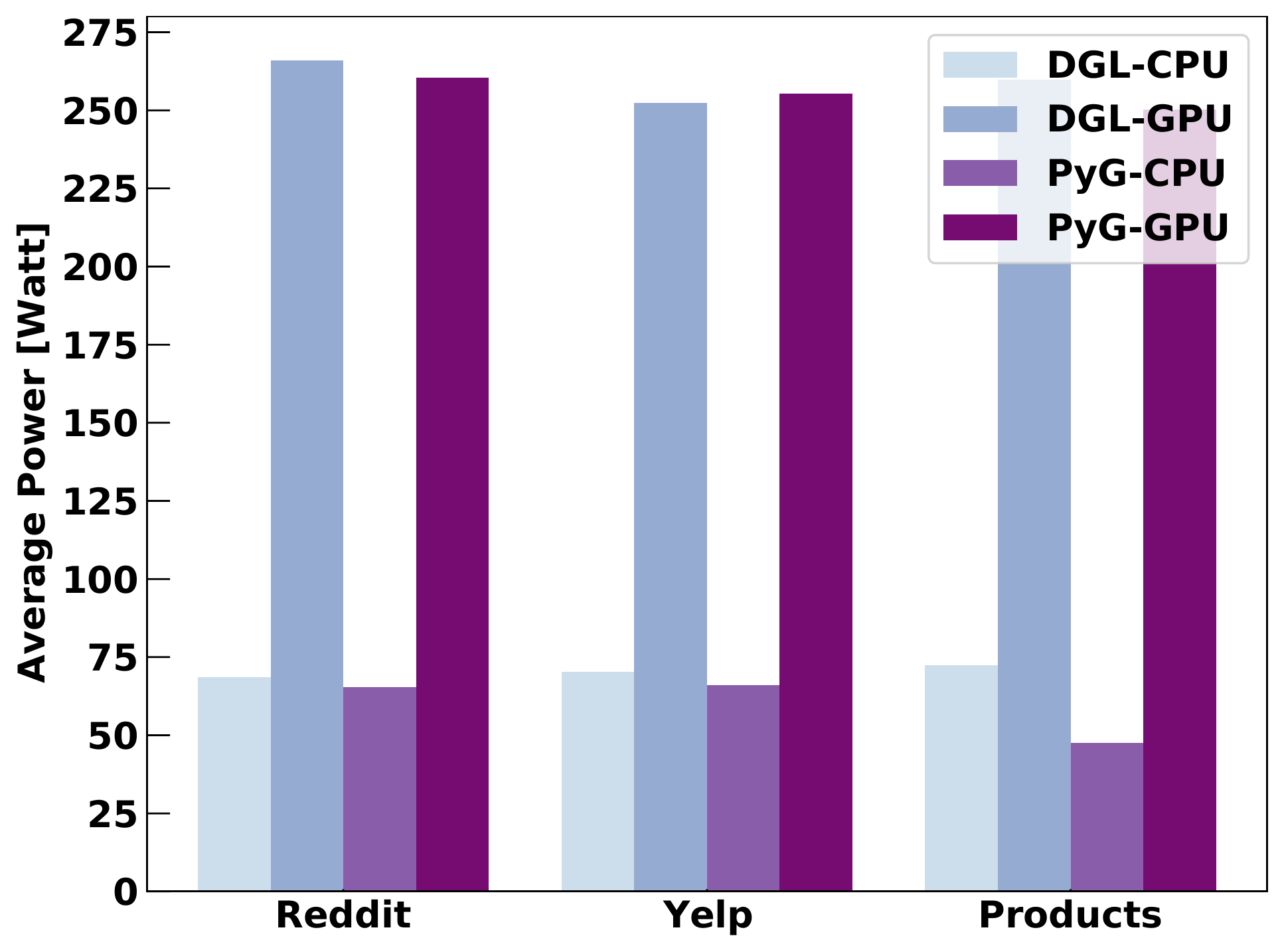}
	}
	\vspace{-1mm}
	\caption{Average power consumption of full-batch GraphSAGE while training.}
	\label{fig:fullbatch-power}
	\vspace{-3mm}
\end{figure}

\begin{figure}[t!]
	\captionsetup[subfloat]{captionskip=1pt}
	\centering
	\subfloat{%
		\includegraphics[width=0.47\linewidth, trim=0cm 0cm 0cm 0cm, clip]{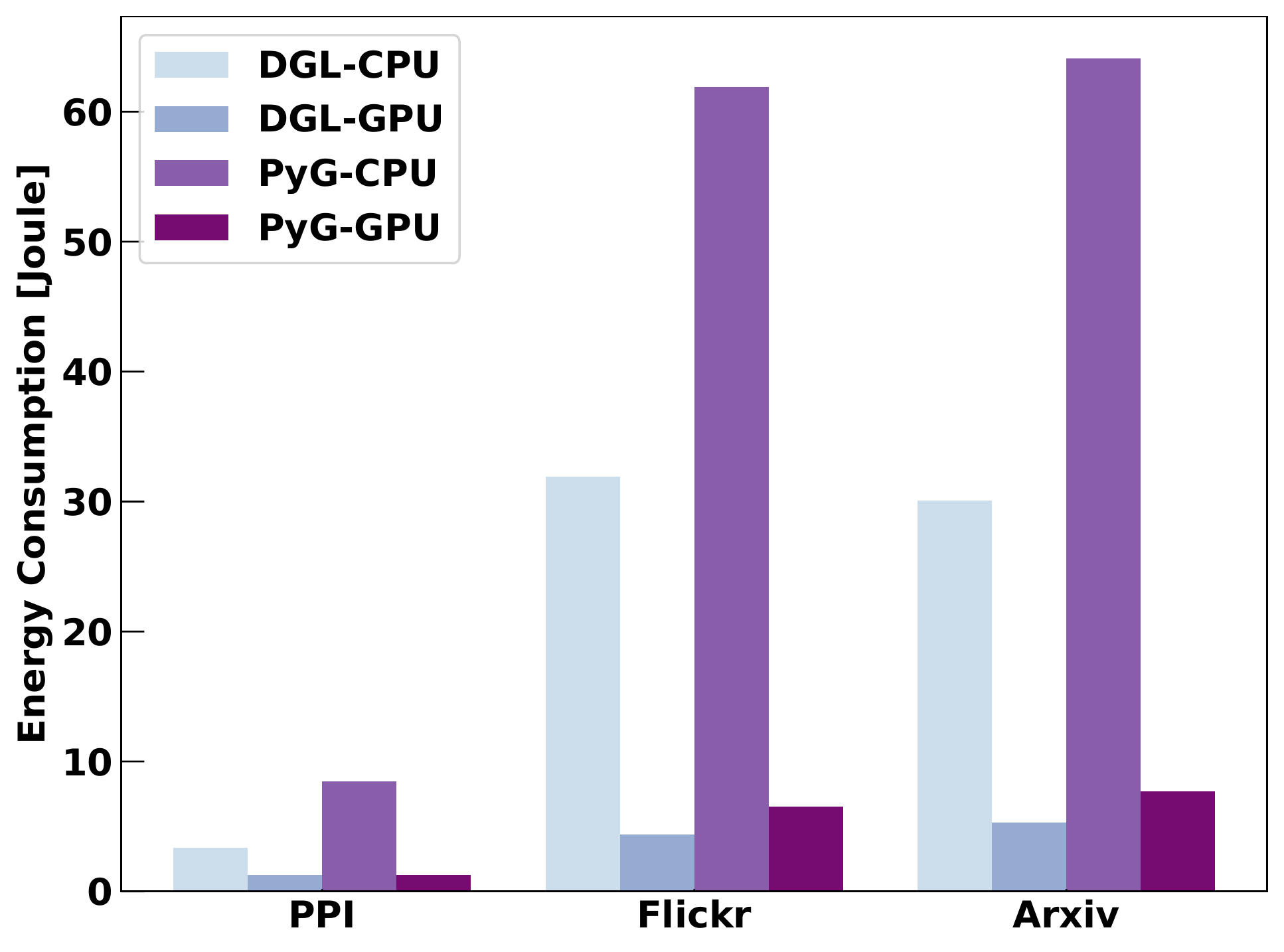}
	}
	\hspace{0mm}
	\subfloat{%
		\includegraphics[width=0.47\linewidth, trim=0cm 0cm 0cm 0cm, clip]{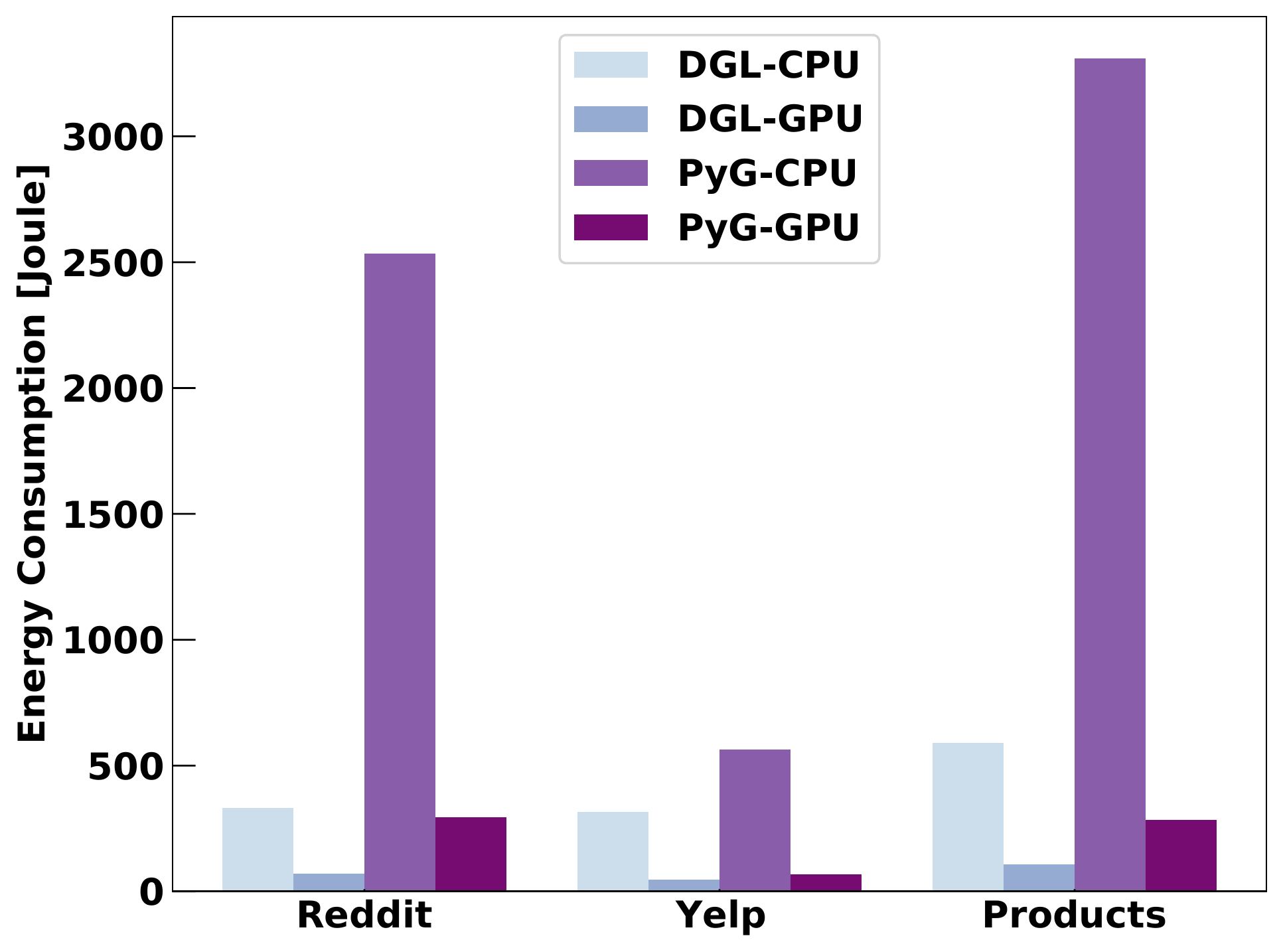}
	}
	\vspace{-1mm}
	\caption{One epoch energy consumption of full-batch GraphSAGE.}
	\label{fig:fullbatch-energy}
	\vspace{-3mm}
\end{figure}

\vspace{1mm}
\noindent \textbf{Full-batch training.} We have focused on three sampling-based GNNs with mini-batch training to evaluate the performance of the frameworks. For a comprehensive evaluation, we here consider \emph{full-batch} training to train a GraphSAGE model, which is done based on the entire graph \emph{without} neighborhood sampling. Specifically, we use a GraphSAGE model with two layers having mean-aggregator and train the model on CPU and GPU using DGL and PyG separately. We present the experiment results, which are the average results of 100 runs for one training epoch, in runtime, power consumption, and energy consumption in Figures~\ref{fig:fullbatch}--\ref{fig:fullbatch-energy}, respectively.

We observe that DGL-CPU is much faster than PyG-CPU for full-bath model training. DGL-GPU training is slower than its PyG counterpart on the smallest graph PPI, while it is faster for the other five datasets. The results are consistent with our functional test results as reported above. We also observe that there is no clear difference in the average power consumption between the frameworks for model training. That is, the differences in energy consumption between the frameworks mainly come from their differences in training time.

\section{Conclusion}

We have characterized the efficiency of two mainstream GNN frameworks with three state-of-the-art sampling-based GNNs and six real-world graph datasets, which cover a wide range of graph sizes. We have conducted extensive experiments to evaluate the performance of each key component of GNN models and frameworks as well as each GNN's model performance from the efficiency perspective in runtime and power/energy consumption. We expect that our observations at many different levels would be useful for further improvement and optimization of GNN models and frameworks.

\section*{Acknowledgments}
This work was supported in part by a grant from SK hynix America and an equipment gift from NVIDIA. This work was also supported in part by the National Science Foundation under Grant IIS-2209921.

\bibliographystyle{IEEEtranS}
\bibliography{ref}

\end{document}